\documentclass{article} 
\usepackage[final]{colm2026_conference}

\usepackage{extra_style}

\title{The Collaboration Gap: Exploration and Benchmarking of Open-World Agentic Cooperation}

\author{Tim R. Davidson\textsuperscript{\dag}\thanks{Research conducted during an internship at Microsoft Research.}
\quad \quad
Adam Fourney \textsuperscript{\ddag}
\quad \quad
Saleema Amershi\textsuperscript{\ddag}
\quad \quad
Robert West\textsuperscript{\dag} 
\vspace{0.5em}
\\
 \textbf{Eric Horvitz\textsuperscript{\ddag}}
\quad \quad
\textbf{Ece Kamar\textsuperscript{\ddag}}
\\\\
\textsuperscript{\dag}EPFL \;
\textsuperscript{\ddag}Microsoft Research
}

\begin{document}

\ifcolmsubmission
\linenumbers
\fi

\maketitle

\begin{abstract}
The trajectory of AI development suggests that we will increasingly rely on agent-based systems powered by language models, composed of independently developed agents with different information, privileges, and tools.
The success of these systems will depend critically on effective collaboration among these heterogeneous agents, even under partial observability.
Despite intense interest, the literature lacks empirical studies evaluating agentic collaboration without relying on fixed communication protocols, limiting insights for open-world deployments.
We propose an illustrative collaborative maze-solving benchmark that (i) isolates collaborative capabilities, (ii) modulates problem complexity, (iii) enables scalable automated grading, and (iv) imposes no output-format constraints, benchmarking unguided, natural communication. 
Using this benchmark, we evaluate 32 leading open- and closed-source models in solo, homogeneous, and heterogeneous pairings.
Our results reveal a surprising \textit{collaboration gap}: models that perform well solo often degrade substantially when required to collaborate.
We identify mitigations that show remarkable influence.
For example, a small nudge via a \textit{relay inference} approach, where a stronger agent leads before handing off to a weaker one, closes much of the gap.
Our findings argue for (1) collaboration‑aware evaluation, (2) training strategies to enhance collaborative capabilities, and (3) deliberate interaction design to elicit agents' collaboration skills, principles relevant to AI--AI and human--AI settings where agents must establish common ground.
\kern-0.4em\renewcommand{\thefootnote}{$\S$}
\footnote{Code for this project is available at {\hypersetup{urlcolor=black}{
\texttt{https://github.com/trdavidson/collaboration\_gap}.
}}}
\renewcommand{\thefootnote}{\arabic{footnote}}
\end{abstract}

\section{Introduction} \label{sec:introduction}

Although collaborative interaction is remarkably efficient for humans \citep{garrod2004conversation}, it is unclear to what extent such collaborative competency holds for AI agents powered by language models (LMs).
Collaborative competency among agents is a rising concern with the deployment of agentic systems. The quest for faster innovation and economic progress through AI promises a steep increase in systems composed of multiple, independently developed AI agents. Due to privacy and organizational constraints, agents are provided with varying capabilities, contextual information, privileges, tools, and allowable actions. 
 
Current attempts at integrating multi-agent solutions strongly rely on predetermined interoperability protocols, e.g., MCP \citep{mcp2024anthropic}, A2A \citep{a2a2025google}, and ACP \citep{acp2025ibm},
or centrally orchestrated architectures \citep{GuoCWCPCW024}.
Such reliance poses two clear obstacles for open-world integration of AI agents:
First, given the fast-paced rise of heterogeneous agentic actors and systems, defining a fixed communication protocol in advance for every possible real-world situation is not feasible. 
Second, even if developing an expressive, standardized protocol \textit{was} feasible, there is no way to enforce such protocols on agents controlled by multiple parties.
Consequently, successful open-world deployment of AI agents will rely on their ability to engage in flexible, on-the-fly communication.

The emergence of sophisticated capabilities from pre-training data is a defining feature of modern \lm{}s \citep{brown2020language, wei2022emergent}. 
Yet, most benchmarks focus on static reasoning capabilities \citep{lm-eval-survery2024}, instead of capabilities unique to agentic use cases such as dynamic, long-horizon, multi-agent collaboration in partially observed environments \citep{davidson2024evaluating}.
Without targeted evaluations, this leaves a critical question regarding the limits of current training recipes:
How does solo performance on tasks relate to \textit{ad hoc} collaborative competency of AI agents \citep{stone2010ad}?

So far, research on collaboration involving LM-powered agents has predominantly focused on human--AI collaboration, optimizing AI's role of a ``helpful assistant'' \citep{bai2022training}.
Still, it is unclear how much progress in this area transfers to AI--AI collaboration.
Since collaboration underpins many societal functions, understanding how AI agents work together is vital for their successful integration.

AI's impressive performance in areas like creative writing, coding, and STEM subjects makes it tempting to also expect proficiency in collaboration, a capability so natural to humans it is easily taken for granted.
However, effective collaborative interaction among humans has been shown to rely on multiple levels of analysis, drawing upon numerous subtle cues and assumptions \citep{clark1996using, garrod2004conversation}.
Research on grounding --- or achieving mutual understanding --- between AI systems and people, building on results from studies in psychology, further emphasizes the need for multiple perceptual and inferential skills to achieve fluid collaboration in the open world \citep{pejsa2014collab}.
Our limited understanding of AI agents' collaborative capabilities can be partially explained by their recency.
More fundamentally, evaluating
advanced
AI capabilities is simply very challenging; benchmarks quickly become outdated 
\citep{
maslej2025artificial, 
kwa2025measuringaiabilitycomplete},
confounded by other factors \citep{henderson2018}, or fail ecological plausibility \citep{de2020towards}.
To study autonomous AI--AI collaboration at scale, we need tasks with automated outcome metrics that allow us to modulate complexity, isolate collaborative capabilities, and put minimal constraints on model outputs.

The arrival of strong \lm{}s enabled the creation of general-purpose agents using rich natural language.
This shifted focus from learning low-level control problems through, e.g., reinforcement learning (RL) \citep{zhu2024survey}, to orchestrating high-level reasoning and task execution among capable, pre-trained agents \citep{zhuge2025mindstorms}.
Although such orchestrated solutions and carefully crafted agentic teams have shown great promise \citep{
GuoCWCPCW024, 
fourney2024magentic,
hong2024metagpt, 
chen2024agentverse,
tran2025multi}, a recent study of popular multi-agent systems collaborating on math and coding tasks identified ``\textit{failures arising from ineffective communication, poor collaboration, [and] conflicting behaviors among agents}'' as an important failure mode \citep{pan2025multiagent}.
These results highlight that collaborative capabilities can form a bottleneck even in tightly orchestrated systems.

Studies of emergent interactions in role-based societal simulations offer interesting insights into model behavior \citep{park2023generative}, but lack control and measurable outcomes relevant to collaboration. 
Other recent works by \citet{wu2025collabllm} and \citet{zhou2025sweet} optimize \lm{}s' collaborative capabilities for human--AI collaboration through RL.
While these studies use \lm{}s to simulate the human actions, the core focus is not AI--AI collaboration.
In general, most studies explore \textit{homogeneous} interactions, where all agents are powered by the same underlying model, or a limited selection of \textit{heterogeneous} models, likely limiting their generalization \citep{wynn2025talk}.
\citet{davidson2024evaluating} use negotiations to evaluate interactions between heterogeneous agents with minimal output constraints.
While negotiations have important collaborative elements, they often contain incentives to withhold information and deceive.

We introduce a scalable maze-solving benchmark designed to isolate dynamic collaborative capabilities.
Using this benchmark across 32 leading open- and closed-source \lm{}s, we identify a critical and counterintuitive phenomenon we term the ``collaboration gap:'' models that are highly capable solo performers exhibit a statistically reliable performance drop when collaborating with an identical copy of themselves.
Our contributions are threefold:
(1) We formally define and provide empirical evidence for the collaboration gap, showing it appears especially severe in distilled models.
(2) We analyze the dynamics of heterogeneous collaboration, revealing that task performance is heavily influenced by which agent initiates the dialogue.
(3) Based on this insight, we propose ``relay inference,'' a new collaborative strategy where a more capable model ``seeds'' the initial steps of a task for a weaker one, substantially boosting collaborative performance of weaker models.
We interpret our results as an existence proof that effective collaboration represents a distinct axis of capability that current training strategies insufficiently capture.

\section{Measuring agents' collaborative capabilities} \label{sec:method}

\begin{figure}
    \vspace{-2em}
    \centerfloat
    \captionsetup{font=small}
\includegraphics[width=0.85\linewidth]{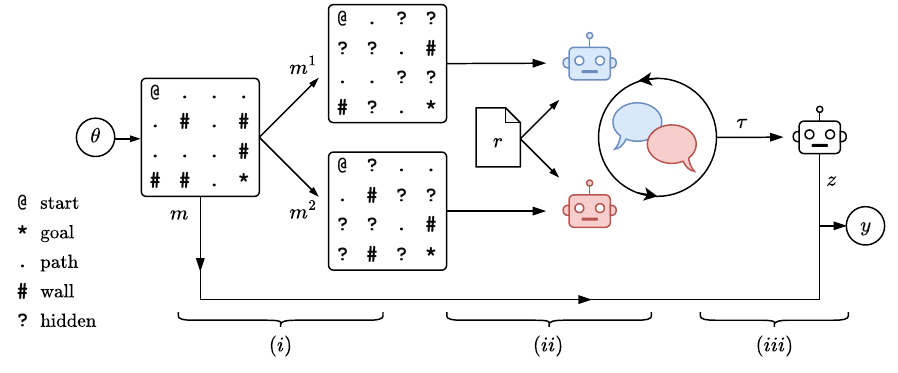}
    \vspace{-1em}
    \caption{\textbf{Core System Diagram.} In sequence, we (i) first generate a random maze $m \sim \mathcal{M}(\theta)$ and split it into two copies, $m^1, m^2$, with roughly half the cells obfuscated using ``?'' symbols, (ii) each agent is then given a maze copy along with the rules, $r$, after which they engage in a dialogue until either the maximum number of turns is reached or the task is completed, finally (iii) the raw transcript, $\tau$, is passed to a ``grader'' agent that extracts the agreed upon route, $z$, which is checked against the ground-truth maze to decide outcome $y$.}
    \label{fig:system-diagram}
\end{figure}

\subsection{Collaborative maze solving}
\label{sec:method:maze-solving}
Following seminal studies on human collaboration \citep{garrod1987saying, garrod1994conversation}, we employ ``maze solving'' as our target task.
Mazes contain desirable properties that make them especially suited to study collaboration.
First, they represent ``fixed'' constructs that nevertheless can be described in a variety of ways; none more correct than another.
For example, the same maze may be described using row-column notation or using visual descriptions, creating the need to ``align'' representations between parties. 
Secondly, while there are many ways to describe a solution, its correctness can be readily verified.
Thirdly, a maze can be solved both ``solo'' and as a team.
This allows us to disentangle agents' ``raw'' maze-solving capabilities from those needed for collaboration.
Finally, mazes can be made arbitrarily complex by simply scaling up their sizes and wall density.

In our study, mazes consist of \maze{N}{N} grids with a start and goal state, separated by path and wall cells.
The dimensions of a maze, distance between start and goal state, and wall density are parameterized by $\theta$.
Following \hyperref[fig:system-diagram]{Figure \ref{fig:system-diagram} (i)}, we first sample a maze instance, $m_i \sim \mathcal{M}(\theta)$.
Next, we introduce a ``twist,'' designed to incentivize agents to collaborate:
Instead of simply providing each agent a copy of the complete maze, we distribute the information by randomly obfuscating half the cells of each copy resulting in $m_i^1$ and $m_i^2$.
Combined, the two copies recover the complete map ($m_i = m_i^1 \cup m_i^2$).
To fairly compare solo to collaborative performance on the decomposed task, we can present a single agent with both of the distributed maps to measure its capability to handle distributed information.

To generate trajectories, we provide two agents (parameterized by \lm{}s) each an incomplete maze copy as a visual, text-based representation of \maze{N}{N} symbols, along with the following rules, $r$: (1) ``\textit{both agents must agree on a move before it is executed},'' and (2) ``\textit{only one move can be executed at a time}.''
The first rule prevents a single agent from unilaterally jeopardizing the rollout, while the second ensures a minimal number of interactions.
Except for a predetermined completion phrase, we do not enforce or provide \textit{any} other guidance on the communication protocol the agents must use or output format they must follow.
Given the visual format of the maze copies, this leaves considerable room for interpretation.
The agents then engage in a multi-turn dialogue attempting to solve the task.
At each turn, agents have access only to their copy of the maze, the rules, and the dialogue thus far.
This process continues until either the maximum number of turns is reached or the predefined completion phrase is uttered, producing final transcript $\tau_i$ (\hyperref[fig:system-diagram]{Figure \ref{fig:system-diagram} (ii)}).
Because no output structure is enforced, there is no straightforward way to deterministically extract the proposed solution.
Instead, we use a third agent as a ``grader,'' tasked with extracting the agreed upon moves at each turn from the raw transcript, $a_{\text{grader}}: \tau_i \to z_i$.
Moves can follow different ``maze schemes.''
For example, they can consist of directions like ``up'' and ``down,'' or coordinates like ``$(0,1) \to (0,2)$.'' They can also use different origin locations, e.g., top-left = 1, or bottom-left = 0.
To ensure the most favorable interpretation we thus transform $z_i$ under a large set of potential schemes and evaluate each against the original maze $m_i$ to obtain the final outcome $y$ (\hyperref[fig:system-diagram]{Figure \ref{fig:system-diagram} (iii)}).
See App. \ref{app:prompts-models} for prompts used.

In contrast to human collaboration studies to date, the ability to use auto-grading allows us to massively scale our experiments.
We perform extensive sensitivity analyses to measure the stability of repeated and different grader models, finding limited variance across average outcomes and virtually no evidence of model-specific biases (see Appendix \ref{app:grading}).

\subsection{Communication challenges}
\label{sec:method:communication}

\begin{figure}[!t]
    \vspace{-2em}    
    \centerfloat
    \captionsetup{font=\captionFontSize}
    \begin{minipage}{0.33\linewidth}
        \begin{messagebox}[BLUE_BOX, verbatim]{grok-3: t1}
            \tiny
\vspace{1em}
[...]\\
From my map, I can see the following:\\
\\
- \textbf{My current position is at (1,1) marked as '@'.}\\
- \textbf{The goal is at (5,5) marked as '*'.}\\
- There are walls ('\#') at positions (1,6), (2,3), (2,5), and (6,6).\\
\textbf{There are clear paths ('.') at positions (1,2)}, (1,3), (1,4), (1,5), \textbf{(2,1)}, (2,2), (2,6),
(3,1), (3,3), (3,5), (4,1), (4,2), (5,1), (5,2), and (6,1).\\
...\\
\textbf{For now, based on what I can see, a possible first move for me would be to move right from
(1,1) to (1,2) since it's a clear path}.\\
...
        \end{messagebox}
    \end{minipage}
    \hfill
    \begin{minipage}{0.33\linewidth}
        \begin{messagebox}[ORANGE_BOX,
            halign title=center
        ]{grok-3: t2}
            \tiny
Thanks for sharing your map details. Let's collaborate to navigate this maze efficiently.
Here's the information from my map:\\
- \textbf{My current position is at (1,1) marked as '@'.}\\
- \textbf{The goal is at (5,5) marked as '*'.}\\
- \textbf{There are walls ('\#') at positions (1,2)}, (3,5), (4,1), and (6,2).\\
...\\
Given this, \textbf{moving right to (1,2) is not possible due to the wall on my map}. Instead, I
suggest we consider moving down to (2,1). \textbf{From your map, (2,1) is a clear path, and while it's
hidden on my map, I’ll trust your information}.
\end{messagebox}
    \end{minipage}
        \hfill 
        \begin{minipage}{0.32\linewidth}
        \begin{messagebox}[BLUE_BOX,
            halign title=center
        ]{grok-3: t3}
            \tiny
...\\
After reviewing the combined information:\\
\\
\textbf{I agree that (1,2) is a wall based on your map, even though it appears as a path on mine.}\\
Since a wall on one map confirms it's impassable, we cannot move right to (1,2). I apologize
for suggesting that move earlier.\\
Moving down to (2,1) seems like a viable option. On my map, (2,1) is a clear path, and
you’ve confirmed it’s also a path on your map. This makes it a safe move.\\
...
        \end{messagebox}
    \end{minipage}
        \hfill 
    \caption{\textbf{Coordinate Grounding Failure.} In the above exchange between two grok-3 agents, the row-col coordinates $(1,2)$ and $(2,1)$ are a path/hidden for agent 1, and hidden/wall for agent 2.
    The first agent uses (row, col) coordinates to describe its map, but does not explicitly specify this (It also wrongly communicates \textit{both} $(1,2)$ and $(2,1)$ as paths).
    Because the start and goal coordinates are symmetric, i.e., $(1,1)$ and $(5,5)$, this is not immediately clear to the second agent, which opts to use (col, row) coordinates.
    }
    \label{fig:coordinate-grounding-failure}
\end{figure}

While straightforward, our maze-solving task offers a rich evaluation environment to test diverse collaborative challenges.
Perhaps the most important among those is the concept of ``grounding,'' the process by which participants attempt to establish mutual understanding \citep{clark1991grounding}.
The goal is to ensure that exchanged information is understood by all parties the same way.
The first challenge is to build a shared mental model of the maze itself.
In order to plan a route, agents must exchange information about their partial views to \textit{ground} their understanding of the full maze, e.g., they must establish a shared understanding of the start and goal position.

Absent a fixed communication protocol, agents must further agree on how to \textit{refer} to locations and actions \citep{garrod1987saying, harnad1990symbol}.
For instance, two agents are in trouble if one uses (row, col) coordinates and the other interprets these as (col, row) (see Figure \ref{fig:coordinate-grounding-failure}).
Similarly, the condition that both agents need to agree on each move necessitates \textit{action grounding}, i.e., to ensure that the intended move is clear to the other agent.

As the trajectory progresses, the agents have to resolve different types of conflicts.
For example, they might encounter \textit{perceptual conflicts} where they disagree on the contents of a specific cell or their current location.
This requires the ability to resolve the inconsistency and update their shared understanding of the maze.
The unstructured nature of the task also introduces a procedural challenge: who gets to propose the next move? Should one agent take the lead, or should they alternate?
This can be especially useful in the case of different capabilities: can agents recognize which is more capable and dynamically defer?
To that end, \textit{Theory of Mind} \citep{premack1978does} --- the capacity to reason about the other agent's mental state --- can be beneficial both to effectively convince another agent to take an action and to resolve potential conflicts, as we observe in Section \ref{sec:results:heterogenous}.

\section{Experimental setup} \label{sec:exp-setup}

\xhdr{Maze details}
Our main experiments are conducted using \maze{6}{6} mazes with a wall density of 30\% and an average solution path of 7 to 9 steps.
These parameters ensure most models can solve mazes solo, providing a meaningful collaborative baseline; ablations are in App.~\ref{app:maze-params}.

\xhdr{Solo baseline} 
We establish a baseline for solo agents' ``raw'' maze-solving capabilities under two conditions:
a single, fully visible map, and a ``distributed map'' where information is split across two maps as described in Section \ref{sec:method:maze-solving}.
The gap between these conditions reflects the challenge of using distributed information.
In each setting an agent is permitted a ``critic'' step, in which it can review its proposed solution.
We collect 100 samples per setting.

\xhdr{Homogeneous collaboration} In a natural extension of the distributed solo setting, e.g., distributed information and iterative discussion, an agent now has to collaborate with an independent copy of itself, effectively isolating the collaboration capability.
As each agent copy only has access to half the map, they need to engage the other agent to fill in the missing pieces.
We perform 100 rollouts per agent, with a maximum of 50 turns.

\xhdr{Heterogeneous collaboration} This setting follows the same mechanics as the homogeneous one, but focuses on collaborations between agents of different model families and/or different strengths.
This allows us to examine possible ``ordering'' effects on collaborative outcomes.
For each pairing, we perform at least 50 rollouts with a maximum of 50 turns.

\xhdr{Relay inference} 
We further explore the opportunities of efficient heterogeneous agent deployment using a ``relay inference'' approach.
Given two available agents, $A$ and $B$, where $A$ is stronger (costlier) than $B$, assume we have to deploy an agent to collaborate with another party's agent, and that we cannot control the choice of agent used by the other party.
In the event the other party chooses the weaker $B$ to save costs, we would like to know if a strong model can be used to (i) ``prime/ground'' a rollout before switching to a weaker model, and (ii) ``recover'' after starting with a weaker model.
To that end, we first generate a set of rollouts using $AB$ and $BB$.
We then ``freeze'' the first $K \in \{2, 4, 6, 8\}$ turns, after which respectively the weaker or stronger model is swapped in to complete the rollout.
We perform at least 100 rollouts per model pairing and relay point $K$.

\xhdr{Models used}
For the solo and homogeneous collaborative experiments we include most commercially available frontier models (full list in App.~\ref{app:prompts-models}).
To avoid combinatorial explosions in heterogeneous settings, we focus on selected models of the most popular providers: Anthropic, Cohere, Google, OpenAI, and xAI.
For each, we investigate ``in-family'' pairings and selected ``cross-family'' pairings.
We use gpt-4.1 as the grading model (see App.~\ref{app:grading}).

\xhdr{Outcome metrics} We distinguish between a binary success rate, i.e., the provided solution is a correct path from start to goal state, and a ``weighted outcome,'' which is defined as the ratio between ($a$) the optimal path length from start to goal and ($b$) the distance from the last valid position on a path to the goal state, $\frac{a - b}{a}$.
Note that this ratio can be negative.
To provide a continuous evaluation signal, we report weighted outcomes in the results section.

\section{Results} \label{sec:results}

\begin{figure}[t]
    \vspace{-3em}
    \centerfloat
    \captionsetup{font=\captionFontSize}
    \includegraphics[width=0.99\linewidth]{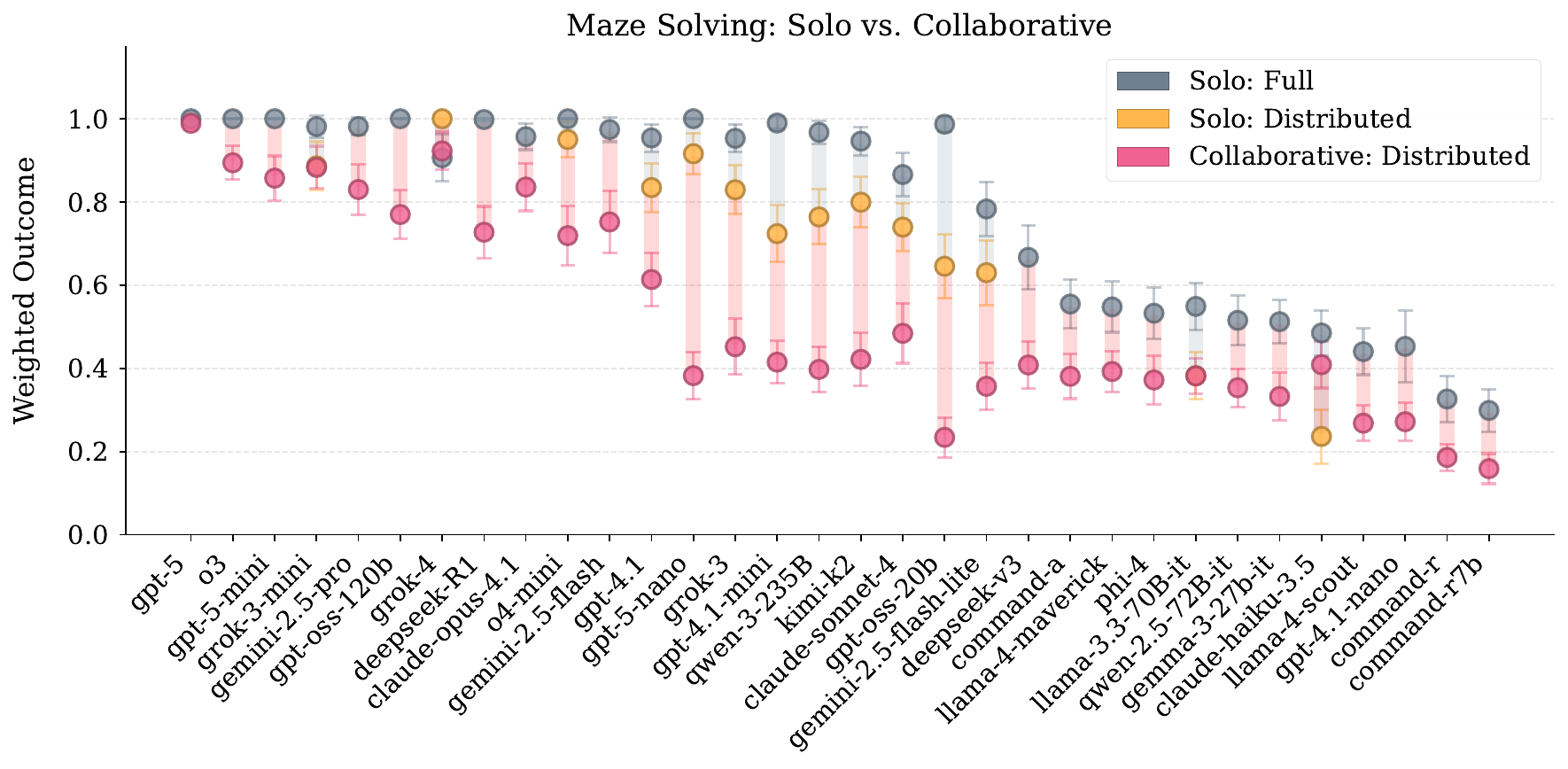}
    \vspace{-0.5em}
    \caption{
    \textbf{The Collaboration Gap.}
    Mean weighted outcomes of \maze{6}{6} maze-solving experiments in solo and homogeneous collaboration mode with 95\% CI.
    We set a ``maze solving'' capability baseline by presenting a model with the full maze (gray marks), showing weighted outcomes surpass 0.5 for most models.
    Next, we present models the same mazes but ``distribute'' the information over two map copies (yellow marks), reporting a performance degradation only if values pass a 95\% CI between the solo settings.
    Finally, we evaluate models' collaborative capabilities by pairing them with an independent copy of themselves, providing each a distributed half of the maze (red marks).
    Most models display a significant performance ``gap'' between solving mazes solo and collaboratively.
    }
    \label{fig:collab-gap-weighted}
\end{figure}

The results section will focus on weighted outcome results in the various settings.
Additional dialogue samples, ablations, and results are available in the Appendix.

\subsection{Solo baselines versus homogeneous collaboration}
\label{sec:results:solo}
Weighted outcomes for solo experiments are shown in Figure \ref{fig:collab-gap-weighted} (gray and yellow markers).
Most evaluated models are capable of at least partially solving mazes under full visibility, with half the models reaching near perfect solve-rates (gray).
Performance drops significantly for about a third of the models when changing to distributed information (yellow).
Overall, most models are capable of (partially) solving \maze{6}{6} mazes.
Virtually all studied models experience a statistically reliable performance drop when moving from a solo to a collaborative setting (red markers), a degradation unlikely to be explained by context or turn limits (see Appendix \ref{app:additional-results:efficiency}).
Larger models and ``thinking'' models tend to perform better than smaller, ``regular'' models, though exceptions exist.
For example, grok-3 and kimi-k2 both underperform grok-3-mini.\footnote{Crucially, grok-3-mini is reportedly \textit{not} a distilled model, but purpose-built for its size \citep{grok3minixai}.} 
Distilled models appear disproportionately affected, e.g., note the large drops from gpt-5 to gpt-5-nano and gemini-2.5-pro to gemini-2.5-flash-lite, suggesting that our current distillation strategies may lose more than just information from the tail-ends of a base model's knowledge distribution \citep{du2021compressed}.

\begin{figure}[!hb]
    \centering
    \vspace{-.5em}
    \captionsetup{font=\captionFontSize}
    \begin{minipage}{0.54\linewidth}
        \begin{messagebox}[GREEN_BOX, verbatim]{o3: turn 1}
            \tiny
        \vspace{1em}
        Here’s my view of the maze \textbf{(row 0/col 0 = top-left)}:\\
        Row 0: (0,0)=@, (0,1)=?, (0,2)=., (0,3)=?, (0,4)=?, (0,5)=?\\
        Row 1: (1,0)=?, (1,1)=?, (1,2)=., (1,3)=\#, (1,4)=\#, (1,5)=?\\
        Row 2: (2,0)=., (2,1)=?, (2,2)=., (2,3)=?, (2,4)=., (2,5)=?\\
        Row 3: (3,0)=?, (3,1)=?, (3,2)=., (3,3)=., (3,4)=?, (3,5)=.\\
        Row 4: (4,0)=\#, (4,1)=?, (4,2)=., (4,3)=?, (4,4)=., (4,5)=?\\
        Row 5: (5,0)=., (5,1)=?, (5,2)=?, (5,3)=\#, (5,4)=?, (5,5)=*  (goal)\\
        \\
        \textbf{1. Could you tell me the contents of any “?” cells you can see}, especially (0,1) and (1,0) so I know a safe first move?\\  
        \textbf{ 2. Where is your current position on your map?}
        \textbf{We’ll use coordinate format (row,col).}
        \end{messagebox}
    \end{minipage}
    \hfill 
    \begin{minipage}{0.45\linewidth}
        \begin{messagebox}[RED_BOX,
            halign title=center
        ]{gpt-4.1-mini: turn 1}
        \tiny
        \vspace{1em}
        Hello! I see that we are \textbf{starting at the position marked '@'} on my map, and \textbf{the goal is marked '*'}. There are some \textbf{hidden cells marked with '?'} that we need to figure out together.
        \\\\
        To begin, I suggest we \textbf{share what we see around our starting position}. \textbf{On my map, from '@', I can move right to '.', or down to '?'}. \textbf{What do you see around your starting position?}
        \end{messagebox}
    \end{minipage}
    \caption{\textbf{Primer Messages.} The difference in the first ``primer'' message to start a dialogue sent by a strong (o3) and weaker model (gpt-4.1-mini).
    Note the effort to unambiguously ``\textbf{ground}'' language on multiple levels by o3, and the lack of clear grounding by gpt-4.1-mini.
    }
    \label{fig:primer-examples}
\end{figure}

What explains these differences? As established above, both o3 and gpt-4.1-mini are capable of solving mazes solo, yet have drastically different outcomes in collaborative mode.
Qualitative analysis of their first messages is illuminating: In Figure \ref{fig:primer-examples}, we display the first message sent by o3 (left) and the first message sent by gpt-4.1-mini (right) to another copy of themselves.
The stronger o3 immediately seeks to align its maze representation by providing a fully determined schema, request for missing information, and immediate next steps.
It also makes sure to ``ground'' the starting position.
In contrast, gpt-4.1-mini only attempts to ground the meaning of different symbols, without proposing a communication schema, or clearly defining its starting position.

App. \ref{app:sec:mechanism:grounding} formalizes this grounding-outcome link.
We first annotate a large, balanced sample of rollouts for various grounding proxies (e.g., aligning on start and goal state) and strategy type (global vs. local planning).
We then compute a grounding score and fit both an additive and interactive mediation model, predicting weighted outcomes while controlling for several confounding factors (e.g., model family and optimal solution length).
Results of clustered bootstrapping on agent pairs are displayed in App. Table \ref{tab:mediation_results}, showing a statistically significant relation between grounding proxies and weighted outcomes.

\subsection{Heterogeneous collaboration}
\label{sec:results:heterogenous}

\begin{figure}[ht!]
    \vspace{-1.5em}
    \centerfloat
    \captionsetup{font=\captionFontSize}
    \begin{subfigure}[b]{0.31\textwidth}
        \centerfloat
        \includegraphics[width=\textwidth]{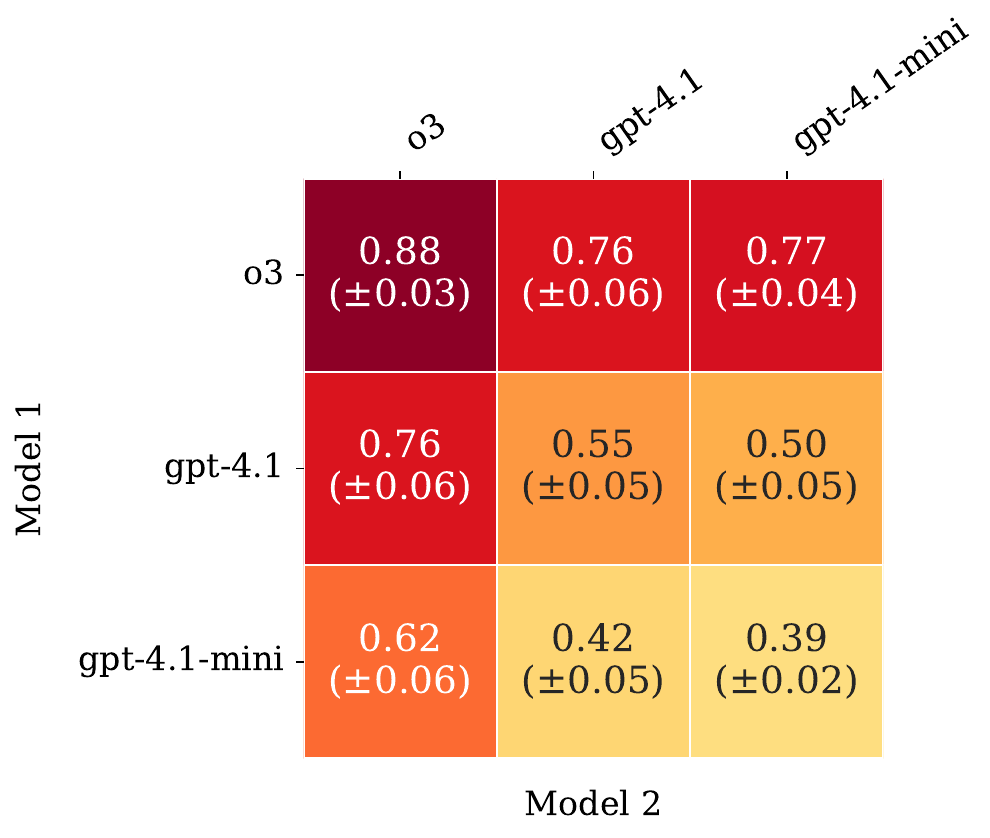} 
        \caption{OpenAI}
        \label{fig:collab-gpt41}
    \end{subfigure}
    \hspace{-1em}
    \begin{subfigure}[b]{0.40\textwidth}
        \centerfloat
    \includegraphics[width=\textwidth]{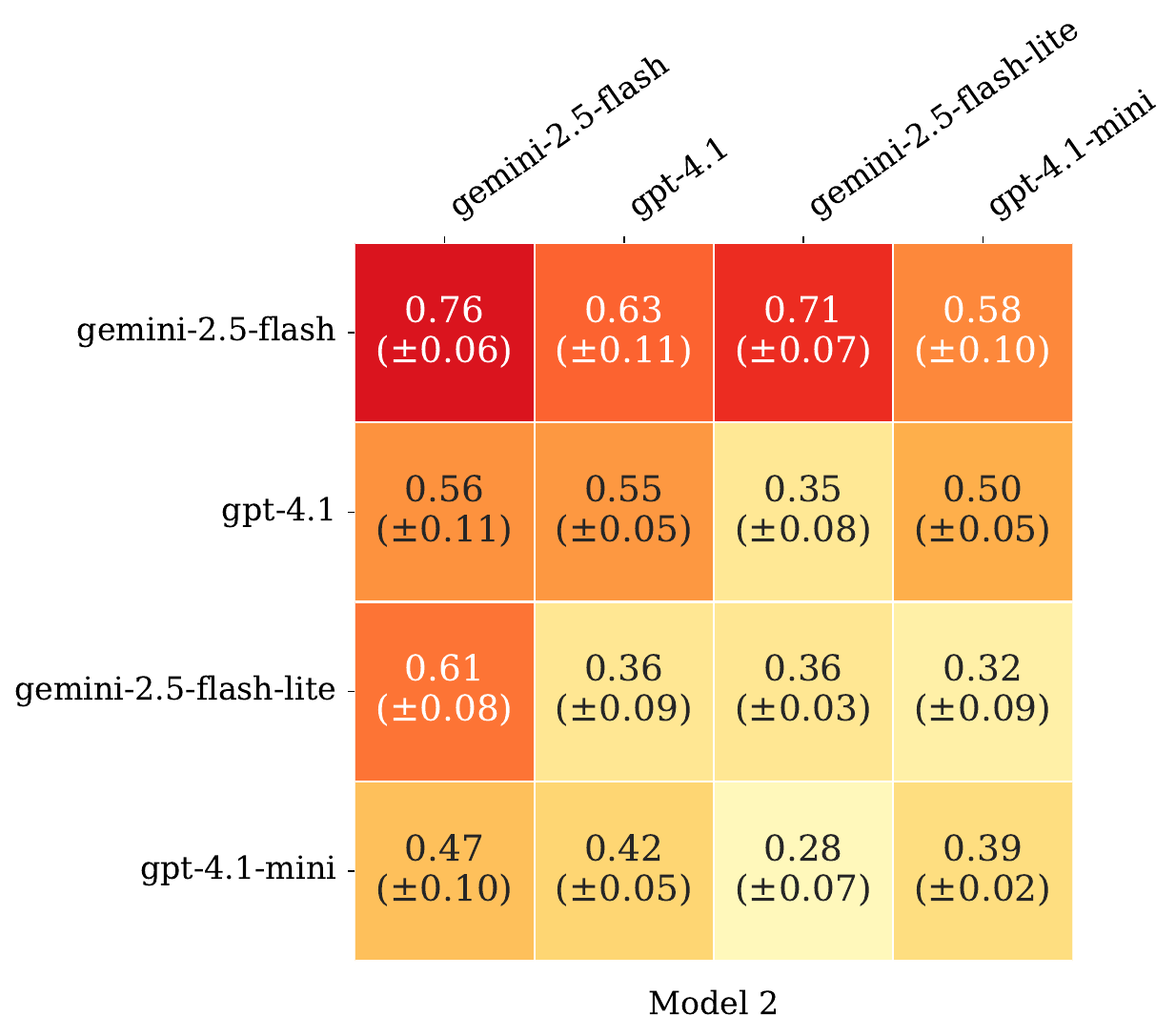} 
        \caption{OpenAI and Google}
        \label{fig:collab-gemini}
    \end{subfigure}
    \hspace{-1em}
    \begin{subfigure}[b]{0.395\textwidth}
        \centerfloat
    \includegraphics[width=\textwidth]{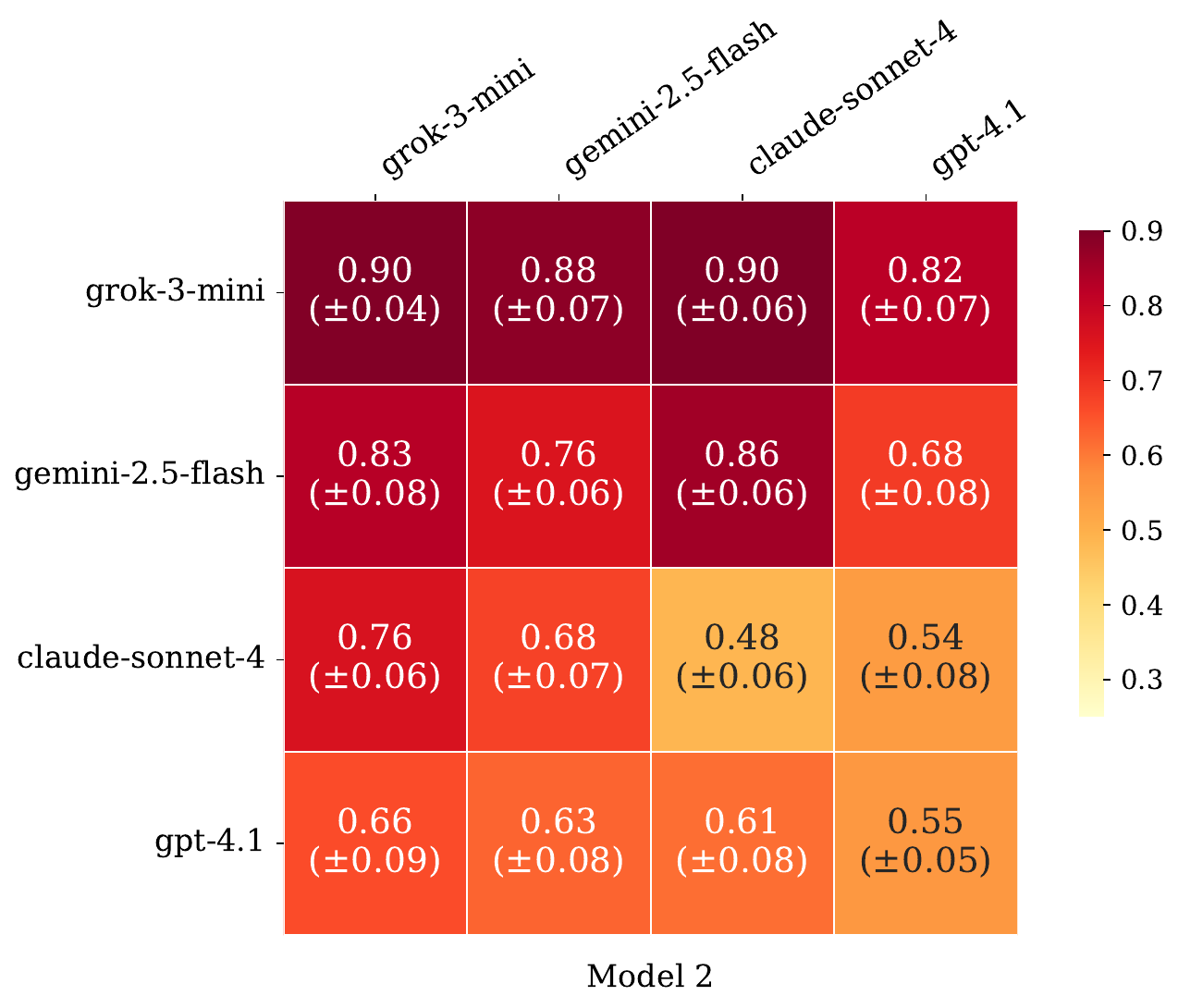} 
        \caption{Anthropic, Google, OpenAI, xAI}
        \label{fig:collab-mix}
    \end{subfigure}
    \caption{
    \textbf{Heterogeneous Collaboration.}
    Each cell represents mean weighted outcomes with 95\% CI below.
    We observe significant ordering and cross-family effects.
    }
    \label{fig:main:collab-heterogeneous}
\end{figure}

Results for selected heterogeneous collaborations are shown in Figure \ref{fig:main:collab-heterogeneous}, with additional results available in Appendix \ref{app:additional-results:wo}.
The diagonals represent homogeneous collaboration.

\subsubsection{Same model family: working with models of different strength}
In panel (a) we observe that pairing a strong model like o3 with a weaker model fails to replicate the performance of the stronger model.
This is surprising, as the stronger model retains the power to veto bad moves, even if it cannot unilaterally solve the maze.
Secondly, we note that pairings can display pronounced ordering effects.
For example, gpt-4.1-mini paired with o3 performs markedly better when o3 starts the collaboration (0.77 $\pm$ 0.04 vs. 0.62 $\pm$ 0.06).
A sample exchange started by gpt-4.1-mini with o3 moving second is shown in Figure \ref{fig:imitation-examples}.
In stark contrast to the grounded approach o3 displayed in Figure \ref{fig:primer-examples}, o3 partially ``imitates'' the communication format introduced by the weaker gpt-4.1-mini.
While o3 still proposes a coordinate system to use, it no longer attempts to ground the complete map, instead opting for a greedy ``local'' strategy.
Similar imitation tendencies can be seen the other way around, where gpt-4.1-mini adopts conventions introduced by o3 (see Figure \ref{fig:app:imitation-strong-weak}).

\begin{figure}[!ht]
    \centering
    \captionsetup{font=\captionFontSize}
    \begin{minipage}{0.40\linewidth}
        \begin{messagebox}[RED_BOX, verbatim, halign title=center]{gpt-4.1-mini: turn 1}
            \tiny 
\vspace{1em}
Hello! I see\textbf{ my current position is at the top-left corner marked '@'}, and \textbf{the goal '*' is somewhere in the bottom-right corner}, but many cells are hidden. Let's start by moving right to reveal more of the maze. Shall we move right?
        \end{messagebox}
    \end{minipage}
    \hfill
    \begin{minipage}{0.59\linewidth}
        \begin{messagebox}[GREEN_BOX, verbatim, halign title=center]{o3: turn 2}
            \tiny
            \vspace{1em}
Agreed—\textbf{moving right from the start puts us at (row 0, col 1)}. On my map that cell is clear, and the \textbf{next one to the right (row 0, col 2)} is also an open path ‘.’. The top row is clear until a wall at column 5, so we could keep heading right, or drop \textbf{down from here (row 1, col 1)} which is also open on my side.
I propose we move right again to (0, 2). Sound good?
        \end{messagebox}
    \end{minipage}
    \caption{\textbf{Style Imitation.} Example of a stronger model imitating the style of a weaker one.
    }
    \label{fig:imitation-examples}
\end{figure}

\subsubsection{Different model family: working with models of varying strength}
When crossing model families (b), we notice the same general trends as in (a): the stronger model generally serves as an upper bound on performance and ordering affects outcomes.
We additionally observe that the weaker model's performance does not necessarily represent a lower bound, e.g., gemini-2.5-flash-lite and gpt-4.1-mini each perform better with copies of themselves than when combined.
We also find evidence that models can display a certain affinity for models of their own family, e.g., note how gemini-2.5-flash-lite does not improve its performance when paired with the stronger gpt-4.1, but performs well when paired with gemini-2.5-flash (see also App. \ref{app:sec:mechanism:disagreements}).

In panel (c) we show pairings of flagship models meant for everyday tasks from leading model builders.
Across different model families, we find the same ordering effects.
Although within the confidence interval, claude-sonnet-4 is the sole exception to this trend: its collaborative performance with gemini-2.5-flash and gpt-4.1 surpasses both of the models' homogeneous performance.
Grok-3-mini shows itself an especially capable collaborator, managing to keep performance close to its high homogeneous baseline.
Notably, it actively challenges incorrect moves rather than deferring to weaker partners.
For instance, see the exchange between grok-3-mini and gemini-2.5-flash in Figure \ref{fig:active-correction}, in which grok-3-mini proposes to move from (1,2) to (2,2).
The gemini model (incorrectly) disagrees with the move and suggests moving back to re-evaluate.
Instead, the grok-3-mini model (i) recognizes the confusion, (ii) attempts to re-calibrate the shared map information, and finally (iii) queries for agreement.
The gemini model recognizes its mistake and the agents continue.

Appendix \ref{app:sec:mechanism:disagreements} quantifies such disagreement dynamics.
We first annotate a large, balanced sample of rollouts for the presence of two sorts of disagreements: ``\textit{factual}'' disagreements based on perception or move legality, and ``\textit{non-factual}'' disagreements based on grounding or strategy.
We then model the likelihood of disagreement events using a logistic regression accounting for factors such as model strength (weak, medium, strong) and interaction type (homogeneous or not).

Results of clustered bootstrapping on agent pairs are presented in Table \ref{tab:disagreement_drivers} of the Appendix, showing that on average, the odds of encountering any conflict are significantly lower for homogeneous collaborations (OR = 0.60).
Strong- and medium-strength models are more likely to encounter disagreements than weaker ones, while factual conflicts are the most likely in medium-strength models.
Non-factual conflicts appear to increase with capability, suggesting a ``sophistication'' shift in agent conflict, with strong models significantly more likely to trigger disputes over strategy or grounding than the medium-strength reference models.
Interestingly, homogeneous pairs disagree less often, but not about different things: once a conflict occurs, they are just as likely as heterogeneous pairs of comparable strength to be disputing strategy or grounding (OR = 0.97).

\begin{figure}[t]
    \centerfloat
    \captionsetup{font=\captionFontSize}
    \begin{minipage}{0.23\linewidth}
        \begin{messagebox}[BLUE_BOX, verbatim]{grok-3-mini: t4}
            \tiny
\vspace{1em}
Yes, I agree to move to (1,2). Now that we're at (1,2), from our combined maps, (2,2) is a path. \textbf{I suggest we move down to (2,2) next}. Do you agree?
        \end{messagebox}
    \end{minipage}
    \hfill
    \begin{minipage}{0.26\linewidth}
        \begin{messagebox}[ORANGE_BOX,
            halign title=center
        ]{gemini-2.5-flash: t5}
            \tiny
I agree we are at (1,2). However, my map shows (2,2) as a wall (\#). Also, combining our maps, (1,3) is a wall. This means \textbf{we are currently blocked from moving forward from (1,2). I suggest we move back to (1,1) to re-evaluate.} Do you agree?
        \end{messagebox}
    \end{minipage}
        \hfill
        \begin{minipage}{0.25\linewidth}
        \begin{messagebox}[BLUE_BOX,
            halign title=center
        ]{grok-3-mini: t6}
            \tiny
\textbf{There's a confusion. From the map you shared, (2,2) is a '.' (path), not a wall, but you said it's a wall. Based on that, I think we should move down to (2,2) since it appears open in your map. Do you agree?}
        \end{messagebox}
    \end{minipage}
        \hfill 
        \begin{minipage}{0.24\linewidth}
        \begin{messagebox}[ORANGE_BOX,
            halign title=center
        ]{gemini-2.5-flash: t7}
            \tiny
\textbf{You are absolutely right! My apologies, I misread my own map.} (2,2) is indeed a '.' (path) on my map.
\\\\
Given that, I agree with your suggestion. Let's move to (2,2). Do you agree?
        \end{messagebox}
    \end{minipage}
    \caption{\textbf{Perceptual Grounding.} Example of two agents resolving a disagreement.
    }
    \label{fig:active-correction}
\end{figure}

\subsection{Relay inference}

\begin{figure*}[ht!]
    \vspace{-0.5em}
    \centerfloat
    \captionsetup{font=\captionFontSize}
    \begin{subfigure}[b]{0.47\textwidth}
        \centerfloat
        \includegraphics[width=\textwidth]{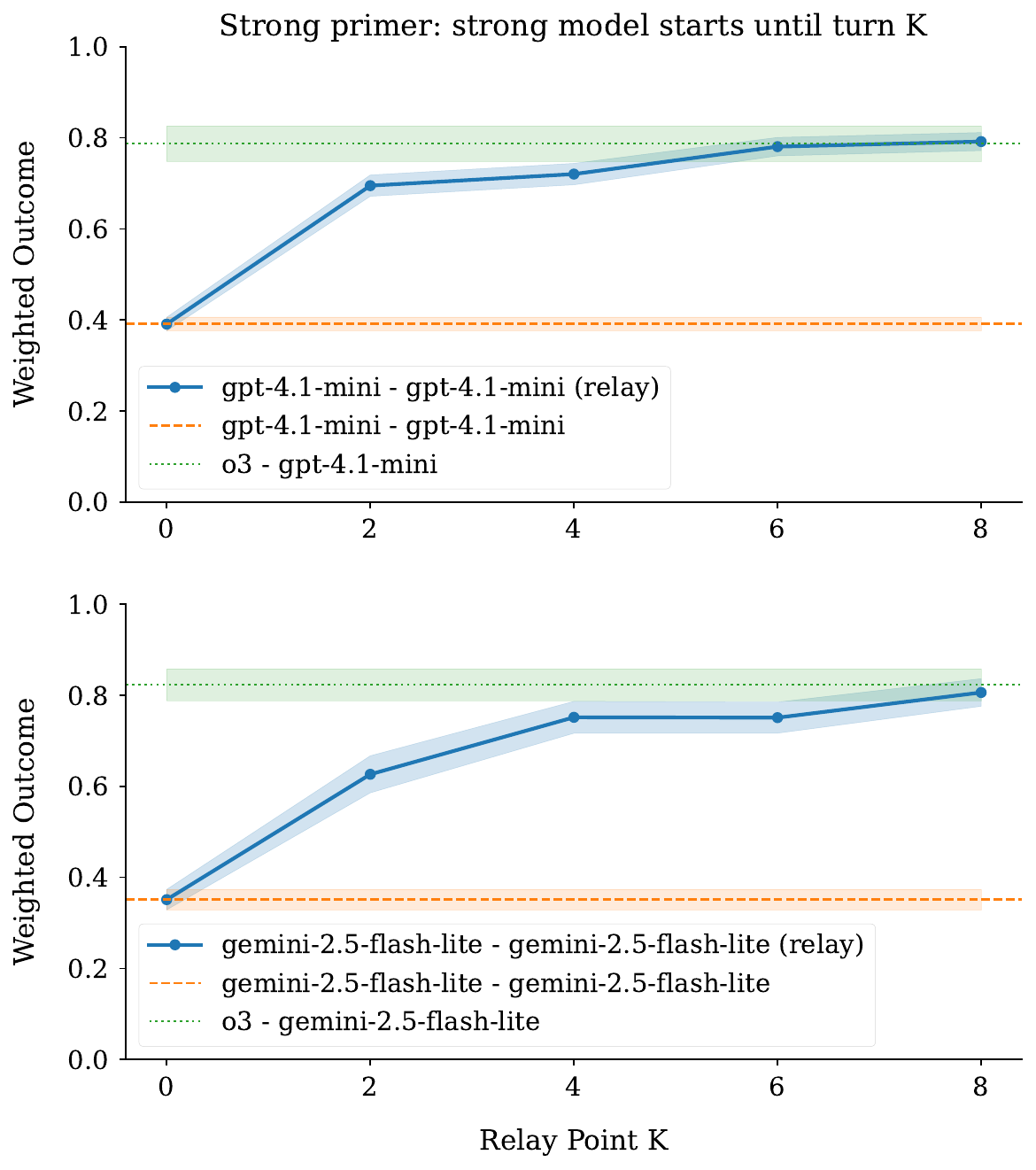} 
        \caption{Strong Primer}
        \label{fig:relay-strong-primer}
    \end{subfigure}
    \hspace{1.5em}
    \begin{subfigure}[b]{0.47\textwidth}
        \centerfloat        
        \includegraphics[width=\textwidth]{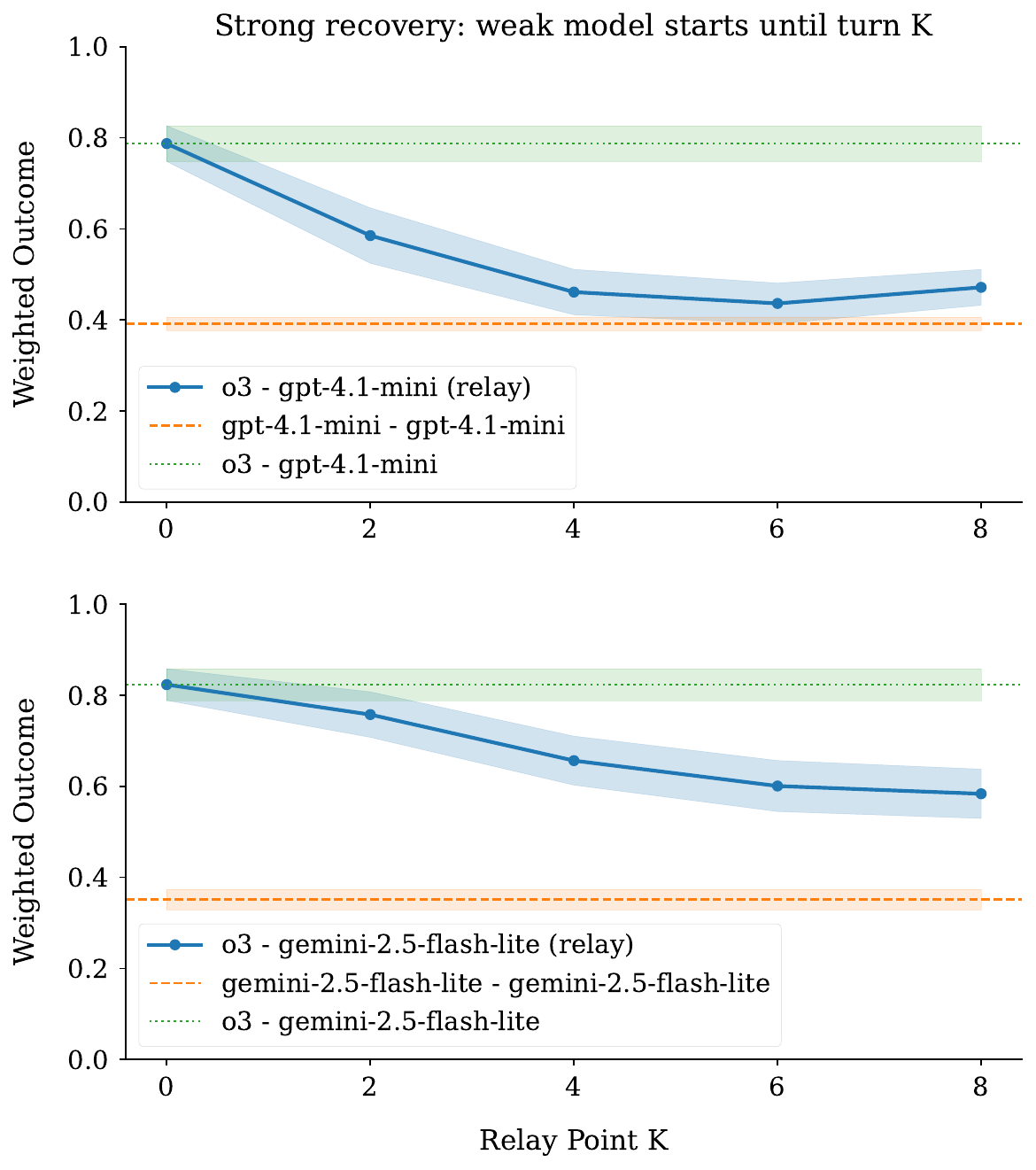} 
        \caption{Strong Recovery}
        \label{fig:relay-strong-recovery}
    \end{subfigure}
    \vspace{-0.4em}
    \caption{
    \textbf{Relay Inference.}
    We test the importance of the initial interactions on outcomes using a strong model (o3) and two weak models (gpt-4.1-mini, top and gemini-2.5-flash-lite, bottom): 
    (a) a strong and weak model interact for $K$ turns, after which another copy of the weak model is swapped in for the strong one; 
    (b) two weak models interact for $K$ turns, after which the strong model is swapped in.
    Using a strong model for priming appears more effective than using it for recovery.
    }
    \label{fig:main:relay-inference}
\end{figure*}

Relay inference results are displayed in 
Figure \ref{fig:main:relay-inference}.
Panel (a) shows that a single priming message from the stronger o3 can substantially boost performance for both the weaker models (gpt-4.1-mini, top, and gemini-2.5-flash-lite, bottom).
Crucially, o3 only sees its incomplete map copy at the first turn, preventing it from unilaterally proposing a full solution.
Instead, o3's opening message grounds the coordinate schema and reference frame, reducing coordination to instruction-following for the weaker model (see App. \ref{app:example-snippets:strong-weak}).
Conversely, panel (b) shows diminishing returns for late interventions: prolonged initial exchanges between weak models make strong recovery harder, likely due to the style imitation effects observed in Section \ref{sec:results:heterogenous}.
Taken together, these results suggest that using strong models to ``seed'' collaborations can be more effective than using them as backup ``experts,'' jumping in to course correct.

\section{Discussion} \label{sec:discussion}
Due to space constraints, we provide an extensive related work section in Appendix \ref{app:related-work}.

\xhdr{Collaborative models are the future of agentic AI} 
A compelling case for ``small'' language models to lead the agentic age was recently made by \citet{belcak2025small}, who argue that capable, small \textit{specialist} models would be more practical and economical to solve specialist tasks compared to large and expensive \textit{generalist} models.
However, there is an important caveat missing from this discussion: the more specialized an agent becomes, the higher the likelihood it encounters challenges outside its area of expertise.
Consequently, the need to effectively collaborate with ``other'' agents to fill capability gaps increases with specialization.
Our work suggests that for current models, ``naively'' breaking up problems to be solved by multiple agents could introduce collaborative slippage, emphasizing the necessity of explicitly training for collaborative capabilities.

\xhdr{Unlocking collaborative capabilities}
``In-context learning,'' or simply ``prompting'' can be a powerful tool to unlock latent capabilities in \lm{}s \citep{brown2020language}.
Subsequently, a large body of literature and guides has sprung up focused on ``prompt engineering'' to improve model outputs \citep{promtsurvey2024}.
Two important findings from our study are the potential of priming to improve collaboration in weaker models, and, conversely, the subdued performance when weaker models lead.
With human--AI collaboration typically led by humans, these observations hold lessons for a future where AI continues to improve.
First, it gives credence to national initiatives to increase people's competence in interacting with AI \citep{ukgovtraining2025, usgovtraining2024}.
Second, it suggests AI will increasingly have to solve the \textit{specialist librarian problem} \citep{taylor1968question} to improve human--AI and AI--AI collaboration.
In this view, information queries are not single events, but dynamic processes during which the librarian (AI) first has to decipher someone's ``true'' needs before being able to solve them.

\xhdr{Mazes as a lower bound} Clearly, maze environments do not capture the full complexity of real-world collaboration, nor do they serve as a proxy for \textit{all} types of collaboration.
Yet, as argued in Section \ref{sec:method:communication} they do simulate \textit{many} vital aspects of collaboration.
The fact that a large collaboration gap exists even in the stylized case of mazes leads us to conjecture that the gap might be wider --- quantitatively as well as qualitatively --- in more complex, real use cases.
Our ``distributed'' information methodology offers a scalable approach to design new collaboration tasks, the exploration of which we leave to future work.
Take for instance the partial observability that can arise in collaborative software engineering:
Just as two agents in a maze must reconcile different map views, coding agents can face a ``split context,'' where each agent can observe only a subset of the required files. 
This can happen if a code base does not fit into an agent’s context window or if agents have different role-based access controls.
Echoing the maze setup, agents must agree on the folder structure to organize new files or imports, and align on coding conventions, package versions, and shared dependencies, while also resolving the occasional warnings and errors.

\section{Conclusion} \label{sec:conclusion}

Human civilization is a story of continuous, rich collaborations.
While AI promises a new chapter, our work reveals a critical roadblock: a fundamental ``collaboration gap'' demonstrated across today's leading language models powering AI agents.
This is a counterintuitive phenomenon where agents with high solo capabilities exhibit a sharp performance collapse in simple teamwork.
Enabled by a novel benchmark that isolates collaborative capabilities, our work demonstrates interaction order is a critical factor.
Our findings offer a promising insight, showing that strategically priming interactions via ``relay inference'' can reliably improve joint outcomes.
The gap's appearance in our simple testbed is particularly concerning, as it reveals a blindspot in current training paradigms rather than an artifact of task complexity.
Collaboration is a critical capability and recognizing its centrality is an opportunity for investing in addressing gaps and boosting AI capabilities.
This calls for a paradigm shift, an idea concisely articulated by \citet{grosz1996collaborative}: ``\textit{capabilities needed for collaboration cannot be patched on, but must be designed
in from the start}.''
We challenge the research community to treat collaborative intelligence as a core objective to be designed for, not as an emergent property to be hoped for.

\subsubsection*{Acknowledgments}
The authors thank Eric Zhu and members of the Microsoft Research AI Frontiers Team for stimulating discussions and feedback that contributed to this work.

\newpage
\clearpage

\section*{Reproducibility statement}
\label{app:limitations}
We open-source the code used to conduct the experiments described in this work at:\\
\url{https://github.com/trdavidson/collaboration\_gap}.

\xhdr{System prompts and user instructions}
We share all prompts and models used to produce the results of this study in Appendix \ref{app:prompts-models}, also including a detailed discussion on our implementation considerations.
As with all works aimed at evaluating capabilities across \lm{}s, there is likely a set of system prompts and instructions more suitable for a particular model.
Due to the opaqueness of closed-source models, it is unfortunately impossible to determine if our prompts clashed with existing developer prompts, unfairly putting some models at a disadvantage.
We made best efforts to ensure that all studied models understood the instructions and respected the agent-to-agent message protocol described in Appendix \ref{app:multi-role-format}.

We did not always succeed: upon qualitative inspection it was found that the command-a model from Cohere would sometimes ``prematurely'' use the predefined completion phrase when hallucinating about a successful outcome.
Due to the large scale of our experiments, modifying the instructions was no longer feasible.

\xhdr{Non-determinism}
Most models accessible over APIs are not deterministic \citep{nondeterminism2025}.
As the majority of our study consists of repeated interactions between (different) models, this non-determinism explodes.
To mitigate this problem we made sure to collect at least 100 rollouts for homogeneous and 50 rollouts for heterogeneous collaborations.
Whenever we report metrics, we include the 95\% confidence intervals based on standard errors. 
While this strategy does not guarantee that the metrics we report represent \textit{every} possible rollout, they do capture average tendencies.
We also made sure that all mazes generated during this study used fixed random seeds for a fair comparison between models.

\xhdr{Auto-grading}
As described in Section \ref{sec:method:maze-solving}, the unstructured format of the generated rollouts makes it impossible to deterministically extract the proposed solutions.
Since using human annotators is not feasible given the scale of our experiments (many tens of thousands of transcripts), we instead opt to use \lm{}s as graders.
In our experiments, we used OpenAI's gpt-4.1.
Due to the non-determinism considerations mentioned previously, this can in theory introduce unwanted noise into the evaluation process.
Furthermore, there is the possibility that some grader models might unfairly disadvantage participating models, e.g., an OpenAI model might be better at extracting results from OpenAI rollouts compared to those coming from Google or xAI.

We therefore conducted a comprehensive ablation in Appendix \ref{app:grading} to measure both the consistency of performing multiple rounds of grading using gpt-4.1 as well as possible changes in results when using the stronger OpenAI o3 model or Google's gemini-2.5-flash.
We further stratified our analysis across a selection of the different models studied.
We did not find any evidence for statistically significant noise in the grading process or unfair biases across models.

We did encounter at least one instance of an unanticipated maze schema not captured by our solution normalization process (see Appendix \ref{fig:gpt5-false-failure}).
The large number of manual author checks does not suggest this is a widespread problem.

\xhdr{Costs of multi-turn rollouts}
The goal of our study was to capture generalizable insights relevant to \lm{}s' collaborative capabilities.
We therefore attempted to include as many commercially available open- and closed-source models as possible.
However, we recognize that replicating our study comes at a considerable cost, likely prohibitive for many labs and practitioners.
This can partially be attributed to the higher cost of using some frontier models, but also to the increasing cost of generating a subsequent turn in a rollout.
Ignoring the constant system and user instructions messages, for an average message length of $K$ tokens and maximum number of turns $T$, the average input context length becomes $K \cdot (T -1)/2$.

\newpage

\bibliography{main}
\bibliographystyle{colm2026_conference}

\newpage

\appendix

\addcontentsline{toc}{chapter}{Appendices}
\section*{Appendices}
\startcontents[chapters]
\printcontents[chapters]{}{1}

\newpage
\clearpage

\section{Related work}
\label{app:related-work}

Multi-agent collaboration spans decades of research across symbolic AI, reinforcement learning, cognitive science, and natural language processing.
The main text has predominantly focused on our work's relation to studies in human collaboration due to our focus on collaboration using unconstrained natural language.
Rather than attempting exhaustive coverage, this section traces four paradigms most relevant to situating our contribution:
explicitly programmed rational agents, co-optimized reinforcement learning agents, ad hoc teamwork, and modern language-model agents.
By tracking this progression, we highlight a historical reliance on rigid communication scaffolding and demonstrate the contribution of our unguided benchmark and extensive empirical results.
We acknowledge that this framing necessarily omits worthy adjacent work and invite readers to treat this as a selective rather than comprehensive survey.

\xhdr{Disclosure of language model usage} We used language models to help identify potentially relevant related work and polish the prose; all cited papers were independently located, read, and verified by the authors.

\subsection*{Rational agents and system-wide protocols}
Early works on agentic collaboration focused heavily on optimizing ``system-level communication protocols'' to deploy symbolic, rational agents \citep{smith1980contract, rosenschein1988deals}.
These foundational efforts assumed the necessity of universal, formalized, and pre-determined communication frameworks dictated top-down by system designers.
Building on the formal groundwork of how agents form joint intentions and commit to shared goals \citep{cohen1990intention}, researchers explicitly designed rigid communication protocols or ``agent communication languages'' (ACL) in an effort to enable collaboration between agents \citep{labrou1999current}.
Notable examples include KQML \citep{finin1994kqml}, which served as a primary blueprint for creating a universal syntax and semantics for agent communication, Arcol \citep{bretier1996rational}, and FIPA \citep{poslad2001standardizing}, which grounded semantics through an external ontology, relying on strong assumptions such as agents being uniformly cooperative and truthful.
Other approaches utilized UML \citep{odell2000representing}, domain-specific ontologies \citep{cranefield2005ontologies}, and detailed protocol specifications \citep{poslad2007specifying} to standardize interactions.

Notably, contemporary interoperability protocols such as MCP \citep{mcp2024anthropic}, A2A \citep{a2a2025google}, and ACP \citep{acp2025ibm} represent a renewed effort to standardize agent communication for LM-based systems, but share the same fundamental assumption that participating agents can be made to conform to a pre-specified interface.

Parallel to the engineering of communication languages, the field also explored how agents could explicitly reason about collaboration.
\citet{GroszSidner1990} introduced ``SharedPlans'' to formalize how agents construct, communicate, and execute cooperative strategies, while others simulated emergent cooperation using simple agents \citep{batali1998computational, cangelosi2002computer} or explicitly modeled strategic helpfulness to others \citep{kamar2009incorporating}.
However, these predefined, rigid protocols faced severe limitations; constrained action spaces and explicit, programmatic reasoning around strategies struggle to transfer to heterogeneous agents or new, unanticipated environments.
As noted by \citet{singh1998agent} when reviewing the state of ACLs at the time: ``\textit{Agents from different vendors -- or even different research projects -- cannot communicate with each other}.''

\subsection*{Reinforcement learning agents}
In a move beyond designer-specified rules, \citet{littman1994markov} introduced an important step toward more dynamic, sequential policies by framing multi-agent coordination as a reinforcement learning (RL) problem.
In this paradigm, state transitions are a function of the joint action of multiple agents, making an agent's optimal policy conditional on the behavior of others, e.g., by learning dynamically when and what to communicate \citep{roth2006communicate}.
Deep learning built on this foundation, allowing agents to explicitly learn to cooperate \citep{lowe2017multi} and model responses to other agents \citep{jaques2019social}.

This led to an extensive body of research on emergent communication, where agents learn task-specific protocols from scratch through interaction (\citet{ foerster2016learning, sukhbaatar2016learning, lazaridou2017multi, havrylov2017emergence} and many others -- see \citet{zhu2024survey} for a recent survey).
These approaches demonstrated that effective coordination signals can emerge without human design.
Despite these advances, such ``learned'' languages or protocols tend to suffer from several drawbacks: they are not necessarily interpretable by humans \citep{kottur2017natural}, they can be inefficient in an information-theoretic sense \citep{chaabouni2019anti, rita-etal-2020-lazimpa}, and they typically require centralized training architectures \citep{eccles2019biases}.

Furthermore, these emergent protocols are optimized for a specific task or set of tasks and co-optimized agents, inherently limiting their transferability.
Scaling up to more environments has proven practically challenging \citep{chaabouni2022emergent}.
Efforts to address this through meta-learning and training agents on multiple tasks \citep{team2021open} have increased generalization across benchmarks like Hanabi \citep{bard2020hanabi}, Overcooked \citep{carroll2019utility}, and Melting Pot \citep{ leibo2021scalable}.
\citet{hu2020other} specifically target ``zero-shot coordination'' -- collaborating with unseen partners without joint training -- demonstrating meaningful gains on Hanabi.
However, even these approaches operate within shared action spaces and reward structures, leaving the leap to truly heterogeneous partners communicating in open-ended natural language unaddressed.

\textit{Relevance to our work}: RL approaches demonstrate that agents can learn to collaborate by interacting with their environment.
However, because they co-optimize their communication channels for specific environments, they are not necessarily compatible with unseen partners or new environments.
Our evaluation specifically targets open-world integration, testing agents' zero-shot ability to collaborate with heterogeneous models using flexible natural language rather than co-optimized, task-specific vectors.

\subsection*{Ad hoc teamwork}
Recognizing the limits of jointly optimized agents, the field of ``\textit{ad hoc teamwork}'' (AHT) emerged to study autonomous agent collaboration without prior coordination, initially introduced by \citet{bowling2005coordination} and formally defined by \citet{stone2010ad}.
The core objective is to create an autonomous agent capable of efficiently and robustly collaborating with previously unknown teammates on tasks where all teammates can contribute.
As noted in a comprehensive survey of the field by \citet{mirsky2022survey}, a primary research focus concerns modeling ``other agents,'' e.g., by specifying ``types'' of other agents that might be encountered \citep{barrett2011empirical}, learning teammate representations through neural networks \citep{rabinowitz2018machine} or genetic algorithms \citep{canaan2022generating}, and utilizing online (Bayesian) belief algorithms to update models of teammates dynamically \citep{barrett2017making}.

Other works explored adapting to teammates without explicit coordination protocols.
This included modeling teammates based on task recognition \citep{wu2021too}, learning to model ``adaptive'' teammates that change their behavior \citep{agmon2014modeling, mirsky2018sequential, xie2021learning}, and exploring communication with unknown teammates using limited symbols \citep{barrett2014communicating}.

\textit{Relevance to our work}: AHT is conceptually closest to the setting motivating our work, as it demands dynamic adaptation to unknown partners.
However, past work in this space often assumed either no direct communication channels (relying on implicit communication via action choices) or utilized highly restrictive, known protocols -- in part because sufficiently capable language-model agents have only recently emerged.
Our benchmark extends the ad hoc teamwork philosophy into the era of LMs, evaluating how agents dynamically adapt to one another using rich, unstructured natural language, without any prior coordination, shared training, or agreed-upon communication protocol.
In this sense, our work can be read as an empirical test of whether modern LMs have acquired ad hoc teamwork as an emergent capability.

\subsection*{Language model agents}

Capable language models have opened the door for agents to interact using natural language, shifting the paradigm from specialized control systems to generalized semantic actors.
Early successful versions of this approach often ``glued'' a language model layer on top of an RL or specialized system to interpret and interact in the physical world \citep{ahn2022can, zitkovich2023rt}, or to play interactive natural language games like Diplomacy \citep{meta2022human}.
Other approaches used LMs alongside RL to learn communication protocols that remain interpretable by humans \citep{li2024language}.
As LMs evolved, the field moved away from systems  specifically optimized for a single use case with strong action constraints and toward light scaffolding powered by LMs with frozen weights (see \citet{GuoCWCPCW024} and \citet{luo2025large} for recent surveys).

Despite this inherent flexibility, there are no guarantees that two LM agents share the same reference frame.
To compensate and facilitate convenient output parsing, the vast majority of current literature relies heavily on mutually known, structured communication protocols, e.g., using JSON or YAML \citep{ehtesham2025survey}.
For example, \citet{shi2023cooperation} explored ad hoc teamwork in a limited role-playing game using a fixed, structured communication protocol, as well as a memory module and critic steps.
Similarly, recent collaboration benchmarks enforce rigid interaction pipelines, such as multi-step critic-to-action protocols \citep{agashe2025llm}, structured LM-robot hybrid pipelines \citep{mandi2024roco}, and frameworks that pre-specify roles, communication protocol, and workflows \citep{hong2024metagpt}.
Even in classic benchmark extensions emphasizing forced collaboration and partial completion metrics \citep{sun2025collab}, or when participants optimize single base models for collaborative tasks \citep{smith2025evaluating}, a mutually known, structured protocol is generally assumed.

This reliance on structured protocols is understandable; fixed formats simplify output parsing and reduce failure modes in deployed systems. However, it comes at a scientific cost: it conflates task difficulty with communication overhead, making it challenging to isolate collaborative capability.

\textit{Relevance to our work}: Existing collaboration benchmarks for agents powered by language models rely on (significant) structural scaffolding, which is often not ecologically plausible for open-world integration.
Tasks further consist of problems that can only be completed in a team setting, making it impossible to separate an agent's static \textit{task skill} from its dynamic \textit{collaboration skill}.
By removing format constraints and fixed protocols, our maze-solving benchmark successfully isolates and evaluates true, unguided collaborative capabilities.
Our benchmark further allows for trivial complexity modulation --- by increasing the maze dimensions or changing the wall density --- making it broadly applicable for LMs of all strengths.
Finally, our extensive evaluations covering 32 leading closed- and open-source models provide a valuable snapshot of the current state of the field.

\newpage
\clearpage

\section{Prompts and models used}
\label{app:prompts-models}

\subsection{Models}
Table \ref{tab:model-overview} below shows the complete list of models evaluated in this study:

\begin{table}[h!]
\centerfloat
\begin{tabular}{@{}lrccc@{}}
\toprule
\textbf{Builder} & \textbf{Model} & \textbf{Open/Closed} & \textbf{Distilled} & \textbf{System Prompt Flexible} \\
\midrule
\textbf{OpenAI} & gpt-5 & Closed & No & Yes \\
& gpt-5-mini & Closed & Yes & Yes \\
& gpt-5-nano & Closed & Yes & Yes \\
& gpt-4.1 & Closed & No & Yes \\
& gpt-4.1-mini & Closed & Yes & Yes \\
& gpt-4.1-nano & Closed & Yes & Yes \\
& o3 & Closed & No & Yes \\
& o4-mini & Closed & Yes & Yes \\
& gpt-oss-120b & Open & No & Yes* \\
& gpt-oss-20b & Open & Yes & Yes* \\
\midrule
\textbf{Google} & gemini 2.5 pro & Closed & No & No \\
& gemini 2.5 flash & Closed & Yes & No \\
& gemini 2.5 flash-lite & Closed & Yes & No \\
& gemma-3-27b & Open & Yes & No \\
\midrule
\textbf{Anthropic} & claude opus 4.1 & Closed & No & No \\
& claude sonnet 4.0 & Closed & No & No \\
& claude haiku 3.5 & Closed & No* & No \\
\midrule
\textbf{xAI} & grok-4 & Closed & No & Yes \\
& grok-3 & Closed & No & Yes \\
& grok-3-mini & Closed & No* & Yes \\
\midrule
\textbf{DeepSeek} & deepseek-V3 & Open & No* & Yes \\
& deepseek-R1 & Open & No & Yes \\
\midrule
\textbf{Meta} & llama-4-maverick & Open & No & Yes \\
& llama-4-scout & Open & Yes & Yes \\
& llama-3.3 70B it & Open & Yes & Yes \\
\midrule
\textbf{Cohere} & command-a & Open & No & Yes \\
& command-r & Open & No & Yes \\
& command-r7b & Open & Yes & Yes* \\
\midrule
\textbf{Microsoft} & phi-4 & Open & No & Yes \\
\midrule
\textbf{Moonshot AI} & kimi-k2 & Open & No & Yes \\
\midrule
\textbf{Alibaba} & qwen-2.5-72B & Open & No & Yes \\
 & qwen-3-235B & Open & No & Yes \\
\bottomrule
\end{tabular}
\caption{\textbf{Comparison of AI Models Used to Power Agents by Model Builder.} We classify models as ``Open'' if their weights are publicly available allowing for private deployment. We use publicly available information to annotate if a model is distilled from a larger model or not. When in doubt we add an asterisk (*). The \textit{System Prompt Flexible} describes if (i) the model allows for a system prompt role, and (ii) if such a system prompt can be inserted at an arbitrary position.
An asterisk here indicates that, although the API/model did not throw an error on (i-ii), it also did not seem to \textit{use} more than one system prompt.
}
\label{tab:model-overview}
\end{table}

\newpage
\clearpage

\subsection{Facilitating agent-to-agent communication}
\label{app:multi-role-format}
\begin{figure}[!ht]
    \centering
    \begin{minipage}{0.95\linewidth}
        \begin{messagebox}[GRAY_BOX,
        verbatim,
            halign title=center
        ]{System Message}
            \scriptsize
            \vspace{1em}
The user will act as an intermediary between you and another agent.\\
The user will directly forward your messages to the other agent and vice versa.\\
The user will not see or modify the messages, but will relay them as is.\\
Messages coming from the other agent will be prefixed with "[other agent]:".\\
Messages coming from the user will be prefixed with "[user]:".\\
Do not add any additional prefixes or suffixes to your own messages.
        \end{messagebox}
    \end{minipage}
    \label{prompt:system}
    \caption{\textbf{System Prompt Used to Facilitate Agent-to-Agent Communication.}
    }
\end{figure}

Interactions with \lm{}s are defined by a history of ``message'' objects.
At the minimum, each message consists of its content, content type, and a ``role'' describing the source and/or function of the message.
Most modern \lm{}s are designed to recognize three types of messages:
\begin{itemize}
    \item \textbf{System Message}: High-level instruction(s) that define the AI's persona, rules, and context.
    \item \textbf{User Message}: Content submitted by the user for consideration by the AI.
    \item \textbf{Assistant Message}: Content generated by the AI in response to User Message(s).
\end{itemize}
Some of the more popular APIs only accept a single System message, placed either at the starting position of the message history or outside of the message history entirely (See Table \ref{tab:model-overview}).
It thus often becomes impossible to use the System role as an independent ``third party'' to a conversation.
Many APIs further do not support multiple Assistant messages in sequence.
Nor do they allow the first message in a history to have the Assistant role.

In order to facilitate dialogue between two AI agents in a way that is supported across all API providers, we thus have three options:

\xhdr{1. User as agent}
The most straightforward implementation is to simply use the User role when relaying messages between two agents.
That is, each agent receives messages from the other agent ``as if'' the other agent was a user.
Unfortunately, this comes with two complications: First, \lm{}s are aligned to behave in a particular way when interacting with users.
This can lead, among other things, to a well-reported ``sycophancy'' tendency in some models \citep{sharma2023understandingsycophancylanguagemodels}.
For example, they can become too agreeable.
This can hamper simulations meant to evaluate AI--AI interactions instead of human--AI ones.
Secondly, because some providers do not allow the first message in a history to have the Assistant role, one has to resort to a ``filler'' User message to generate rollouts.
As the model is made to believe it generated these, it is unclear what their influence is on the subsequent rollout.

\xhdr{2. Transcript style}
Given a history of messages, we can transform it into a single User message ``summarizing'' the dialogue as a transcript between the two agents \citep{davidson2024evaluating}.
While this circumvents the issues around having a starting User message and mitigates potential sycophancy tendencies, it likely pushes a model out of its ``regular'' distribution.
That is, models are optimized for sequential messages being sent back-and-forth -- not contributing lines to a script.

\xhdr{3. Explicit prefixes}
In this work we combined insights from both (1) and (2) above as follows: First, we use a System prompt to describe that the model is about to engage with another model and that the user will act as an intermediary.
It goes on to define ``prefixes'' to distinguish between messages coming from the user, ``\texttt{[user]:}'', and those coming from the other agent ``\texttt{other agent}:''.
Next, we use a User message with the user prefix to provide instructions for the task (see \ref{app:prompts:collab}).
This ensures models are always aware that they are conversing with another model and that the first message in the history is always of type User.
At each turn, we can simply convert all messages except the first one from type User to Assistant and vice versa, making sure User messages always contain the relevant prefix.

\newpage
\clearpage

\subsection{Solo prompts}

\begin{figure}[!ht]
    \centering
    \begin{minipage}{0.95\linewidth}
        \begin{messagebox}[GRAY_BOX,
        halign title=center
        ]{User Message}
            \scriptsize
            \vspace{1em}
\# Task\\
You are going to play a maze game.
The maze consists of a grid with walls and a goal.
Your task is to navigate through the maze, avoiding walls, to reach the goal.
\\\\
\#\# Rules\\
- You can only move to adjacent cells (up, down, left, right). \\
- You can not move diagonally or through walls.\\
- The game ends when you have reached the goal or if you step into a wall.\\
- **Your game is timed for speed and put on a leader board so hurry up!**\\
- You get one shot to complete the maze, so make sure to plan your moves carefully.\\
- Provide both your proposed complete solution and any considerations you have made in your reasoning.\\
- Provide your final solution as a list of coordinates.\\
\\\\
\#\# Map\\
Two maps of the maze are provided below, each showing a different portion of the same maze. Here are the two maps of the same maze with a legend of the symbols:
\\\\
Map view 1:\\
\\
@.???\\
.?..?\\
\#???.\\
?...?\\
.??\#*\\
\\
Map view 2:\\
\\
@?...\\
?.??\#\\
?\#..?\\
\#???.\\
?\#\#?*\\

Legend:\\

@ - Current Position \\
* - Goal Position \\
. - Path \\
\# - Wall \\ 
? - Hidden Cell\\

Time to get started! Please begin.
  
        \end{messagebox}
    \end{minipage}
    \caption{\textbf{Prompt Used for Solo Experiments.} Example of a ``distributed'' information prompt, containing two incomplete map copies.}
    \label{prompt:solo}
\end{figure}

\begin{figure}[!ht]
    \centering
    \begin{minipage}{0.95\linewidth}
        \begin{messagebox}[GRAY_BOX,
        halign title=center
        ]{User Message}
            \scriptsize
            \vspace{1em}
Carefully review the final solution you have provided above.\\
Ensure that it is a complete and valid path through the maze.\\
If needed, you can make adjustments to the path to ensure it meets the requirements of the maze game.\\
Remember to consider the maze's rules and constraints while reviewing your path.\\
After you have finished reviewing, please submit your final solution.
        \end{messagebox}
    \end{minipage}
    \caption{\textbf{Critic Prompt.} Models are given a ``critic'' step to improve on their initial solo solution.}
    \label{prompt:critic}
\end{figure}

\newpage
\clearpage

\subsection{Collaboration prompt}
\label{app:prompts:collab}

\begin{figure}[!ht]
    \centering
    \begin{minipage}{0.95\linewidth}
        \begin{messagebox}[GRAY_BOX,
        halign title=center
        ]{User Message}
            \scriptsize
            \vspace{1em}
\# Task\\
You are going to play the collaborative maze game "ACTI" together with another agent.
The maze consists of a grid with walls and a goal.
Your task is to navigate through the maze, avoiding walls, to reach the goal.
You will each get your own map of the **same** maze, with some coordinates hidden.
Because of the hidden coordinates, you will need to communicate with the other agent to share information about the maze and coordinate your movements.
\\\\
\#\# Rules\\
- You can only move to adjacent cells (up, down, left, right).\\
- You can not move diagonally or through walls.\\
- The game ends when you have reached the goal or if you step into a wall.\\
- **Your game is timed for speed and put on a leader board so hurry up!**\\
- You can only take one action per turn.\\
- An action is **immediately** considered executed once both agents agree.\\
- Be sure to say: ``ACTI!", after you reached the goal.
\\\\
\#\# Map\\
Here is **your** map of the maze with a legend of the symbols:\\
\\
@.??? \\
.?..? \\
\#???. \\
?...? \\
.??\#* \\

Legend:\\

@ - Current Position \\
* - Goal Position \\
. - Path \\
\# - Wall \\ 
? - Hidden Cell\\

Time to get started! Please begin.
        \end{messagebox}
    \end{minipage}
    \caption{\textbf{Prompt Used for Collaboration Experiments.} Note the rules agents must follow and the predefined completion phrase ``\texttt{ACTI!}.''}
    \label{prompt:collab}
\end{figure}

\newpage
\clearpage
\subsection{Verification prompts}
\label{app:prompts:verification}

\begin{figure}[!ht]
    \centering
    \begin{minipage}{0.95\linewidth}
        \begin{messagebox}[GRAY_BOX,
            halign title=center
        ]{User Message}
            \scriptsize
            \vspace{1em}
\# Task\\
You will be given the proposed solution to a maze game.
Your task is to extract the proposed route by the agent for the maze.
The agent might consider multiple coordinates at a time, but only a single coordinate in its final move or proposed solution.
Record each coordinate of the **final proposed route** in a separate entry.
\\\\
Further, you should also record the route schema, which describes the format of the coordinates used in the route.
To the best of your ability, you should determine the schema based on the dialogue.
\\\\
The schema should include:\\
- maze\_origin: The origin of the maze, i.e., is the agent treating 0 or 1 as the origin for the maze\\

- maze\_orientation: The orientation of the maze:\\
  "top\_left", i.e., if the top left corner is (0, 0)\\
  "bottom\_left", i.e., if the bottom left corner is (0, 0)\\
  "top\_right", i.e., if the top right corner is (0, 0)\\
  "bottom\_right", i.e., if the bottom right corner is (0, 0)\\
\\
- coordinates\_orientation: The orientation of the coordinates:\\
  "row\_col", if the first coordinate is the row and the second is the column,\\
  "col\_row", if the first coordinate is the column and the second is the row.\\

- coordinates\_symbols: The symbols used to represent coordinates:\\
  "number\_number", e.g., (1, 2)\\
  "letter\_letter", e.g., (A, B)\\
  "letter\_number", e.g., (A, 1)\\
  "number\_letter", e.g., (1, A)\\
  "directions", e.g., "up", "down", "left", "right"
\\\\
Respond in the following format:\\
\\
```yaml\\
route\_schema:\\
maze\_origin: "<0|1>" \\
  maze\_orientation: "<top\_left | bottom\_left | top\_right | bottom\_right>"\\
  coordinates\_orientation: "<row\_col | col\_row>"\\
  coordinates\_symbols: "<number\_number | letter\_letter | letter\_number | number\_letter | directions>"\\
route:\\
  - turn: <turn\_number>\\
    coordinates: [[<coordinate\_1>, <coordinate\_2>], ..]\\
    turn\_type: "move"  \# only consider final moves, not intermediate considerations\\
    agent: "agent\_1"\\
  - turn: <turn\_number>\\
    ...\\
```\\
\\
Do not include any other information, just the YAML object.\\

\# Dialogue\\
<insert dialogue>
        \end{messagebox}
    \end{minipage}
    \caption{\textbf{Verification Prompt for Solo Experiments.}}
    \label{prompt:solo-grader}
\end{figure}

\begin{figure}[!ht]
    \centering
    \begin{minipage}{0.95\linewidth}
        \begin{messagebox}[GRAY_BOX,
            halign title=center
        ]{User Message}
            \scriptsize
            \vspace{1em}
\# Task\\
You will be given a dialogue between two agents trying to solve a maze.
Your task is to extract the route taken by the agents through the maze at each turn.
Some turns, the agents will not move, and some turns they will.
The agents may communicate moves using directions, e.g., "up", "down", "left", "right".
Or the agents may communicate coordinates directly, e.g., "Let's move to (3, 4)".
For turns where both agents agreed to move, record the coordinates or direction of the cell they agreed to move to.
For turns where they did not move, record the coordinates of the cells or directions they considered, if any.
Make sure to record what happened at **each turn**.
You must be consistent in the move format you record, e.g., stick with directions or coordinates, do not mix them unless absolutely necessary.
\\\\
Further, you should also record the route schema, which describes the format of the coordinates used in the route.
To the best of your ability, you should determine the schema based on the dialogue.
\\\\
The schema should include:\\
- maze\_origin: The origin of the maze, i.e., are the agents treating 0 or 1 as the origin for the maze\\

- maze\_orientation: The orientation of the maze:\\
  "top\_left", i.e., if the top left corner is (0, 0)\\
  "bottom\_left", i.e., if the bottom left corner is (0, 0)\\
  "top\_right", i.e., if the top right corner is (0, 0)\\
  "bottom\_right", i.e., if the bottom right corner is (0, 0)\\
\\
- coordinate\_orientation: The orientation of the coordinates:\\
  "row\_col", if the first coordinate is the row and the second is the column,\\
  "col\_row", if the first coordinate is the column and the second is the row.\\

- coordinate\_symbols: The symbols used to represent coordinates:\\
  "number\_number", e.g., (1, 2)\\
  "letter\_letter", e.g., (A, B)\\
  "letter\_number", e.g., (A, 1)\\
  "number\_letter", e.g., (1, A)\\
  "directions", e.g., "up", "down", "left", "right"
\\\\
Respond in the following format:\\
\\
```yaml\\
route\_schema:\\
  maze\_origin: "<0|1>" \\
  maze\_orientation: "<top\_left | bottom\_left | top\_right | bottom\_right>"\\
  coordinates\_orientation: "<row\_col | col\_row>"\\
  coordinates\_symbols: "<number\_number | letter\_letter | letter\_number | number\_letter | directions>"\\
route:\\
  - turn: <turn\_number>\\
    coordinates: [[<coordinate\_1>, <coordinate\_2>], ..]\\
    turn\_type: "<move|consider>"\\
    agent: "<agent\_1|agent\_2>|both"\\
  - turn: <turn\_number>\\
    ...\\
```\\
\\
Do not include any other information, just the YAML object.\\

\# Dialogue\\
<insert dialogue>
        \end{messagebox}
    \end{minipage}
    \caption{\textbf{Verification Prompt for Collaboration Experiments.}}
    \label{prompt:collab-grader}
\end{figure}

\newpage
\clearpage

\section{Maze hyperparameter ablations}
\label{app:maze-params}

\subsection{Varying maze size}

\begin{figure*}[ht!]
    \centerfloat
    \begin{subfigure}[b]{0.95\textwidth}
        \centerfloat
        \includegraphics[width=\textwidth]{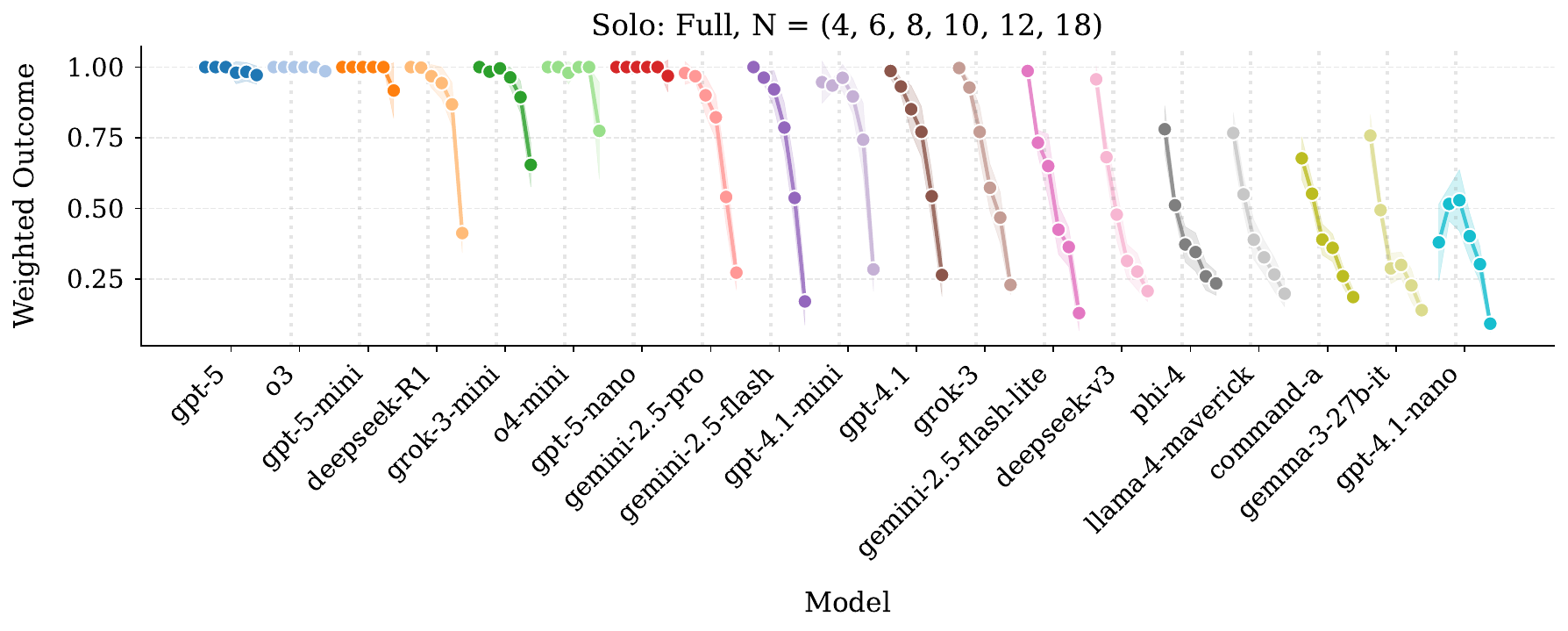} 
        \label{fig:app:solo-full}
    \end{subfigure}
    \hspace{1em}
    \begin{subfigure}[b]{0.95\textwidth}
        \centerfloat
        \includegraphics[width=\textwidth]{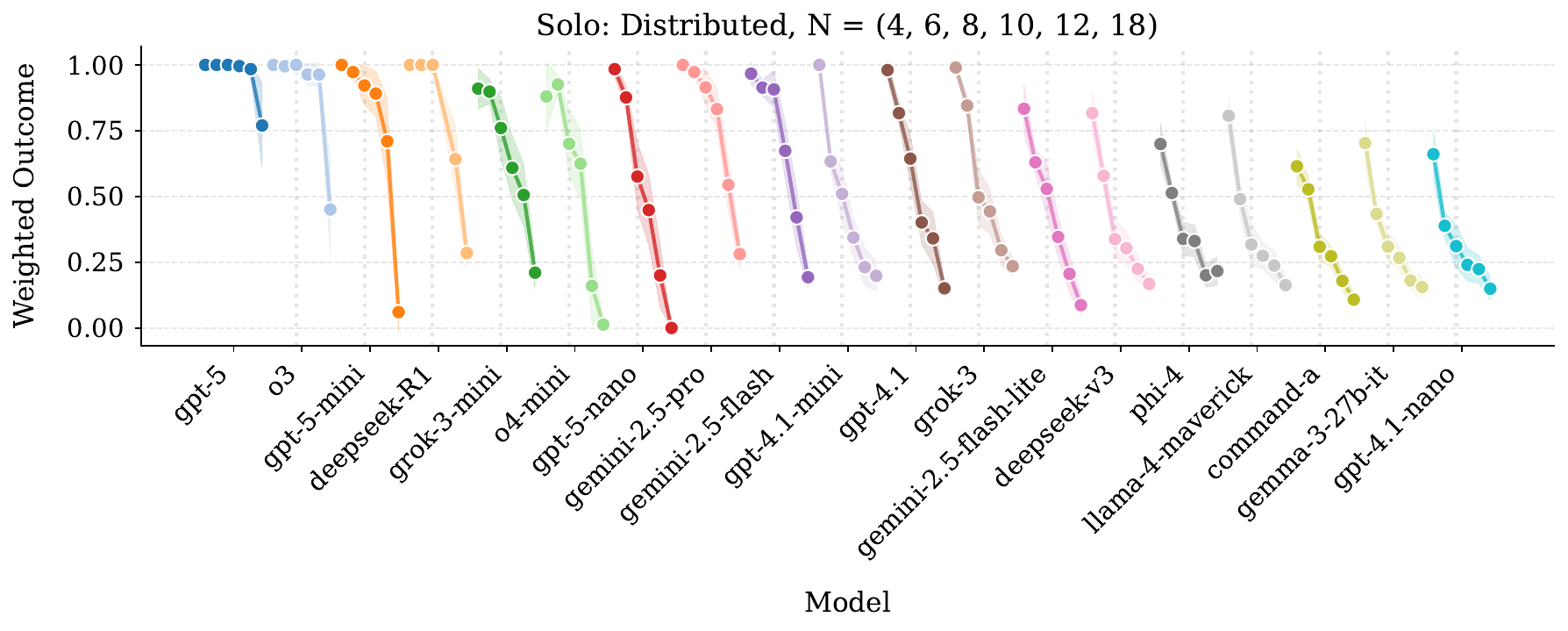} 
        \caption{\textbf{Solo: Full and Distributed.}
        For each model, we report solo mean performance with 95 \% CI on $N \times N$ mazes where $N \in \{4, 6, 8, 10, 12, 18\}$.
        We note that for most models performance drastically drops in the distributed setting for $N > 6$. Both gpt-5 and o3 perform remarkably well up to $N=12$.
        }
        \label{fig:app:solo-dist}
    \end{subfigure}
    \hspace{2em}
    \begin{subfigure}[b]{0.95\textwidth}
    \vspace{2em}
        \centerfloat
        \includegraphics[width=\textwidth]{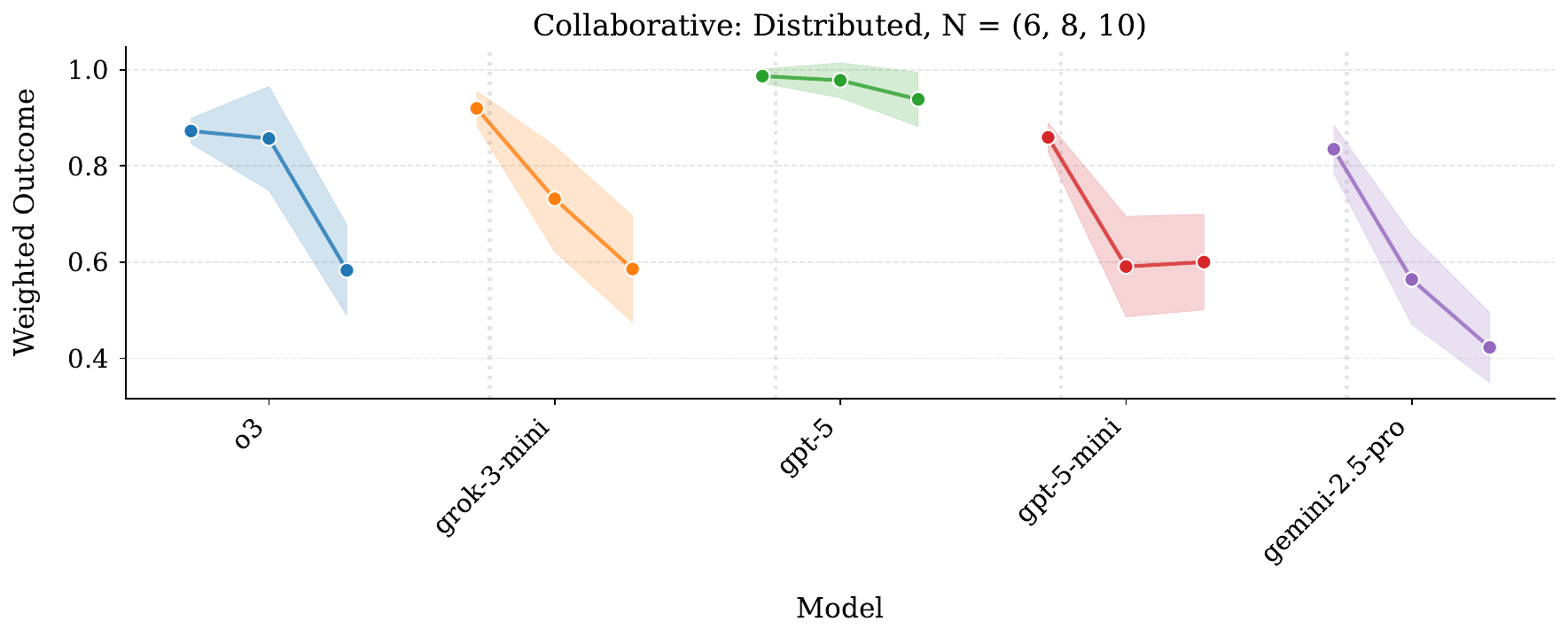} 
        \caption{\textbf{Homogeneous Collaboration.}
        We ablate collaborative performance for selected strong performers on $N \times N$ mazes where $N \in \{6, 8, 10\}$. Note that performance drops considerably for all models except gpt-5.
        }
        \label{fig:app:collab-dist}
    \end{subfigure}
    \caption{
    \textbf{Weighted Outcomes for Maze Size Ablations.}
    }
    \label{fig:app:maze-size-ablations}
\end{figure*}

\newpage
\clearpage
\subsection{Varying wall density}

\begin{figure*}[h!]
    \centerfloat
    \begin{subfigure}[b]{0.95\textwidth}
        \centerfloat
        \includegraphics[width=\textwidth]{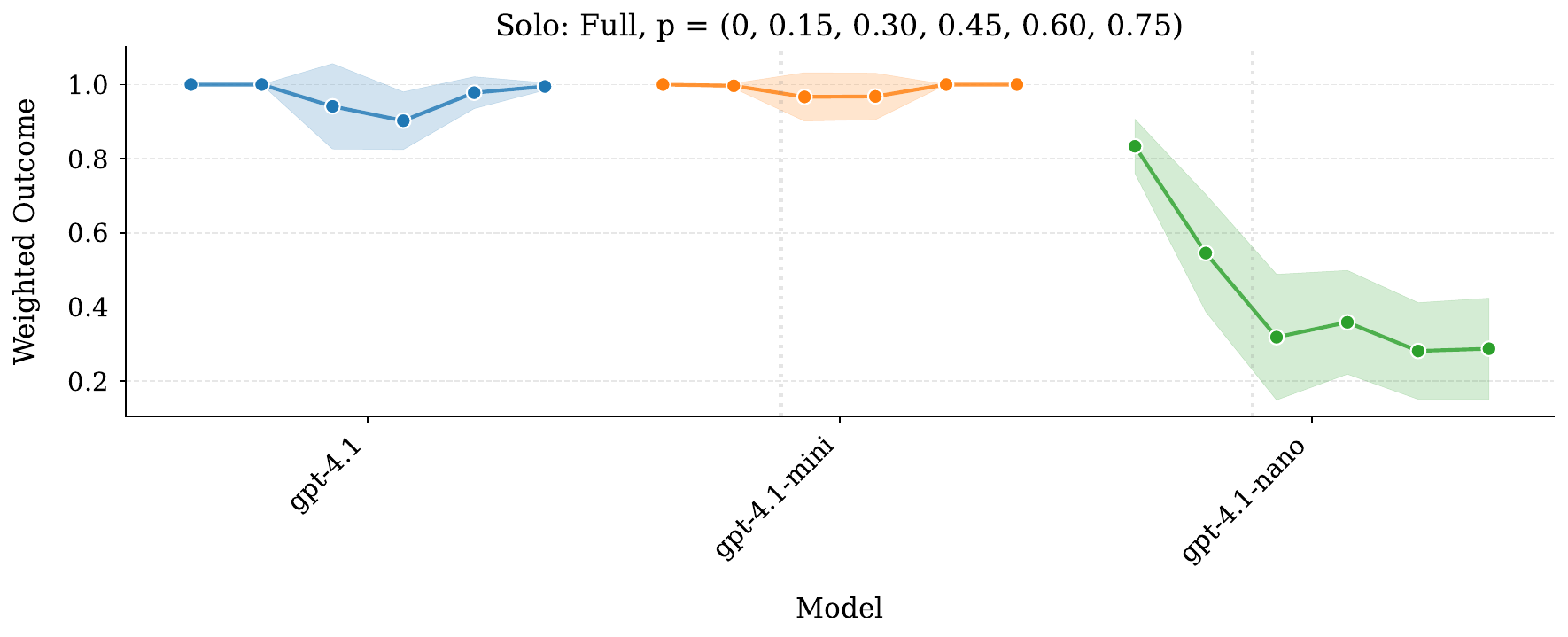} 
        \label{fig:app:solo-full-wall}
    \end{subfigure}
    \hspace{1em}
    \begin{subfigure}[b]{0.95\textwidth}
        \centerfloat
        \includegraphics[width=\textwidth]{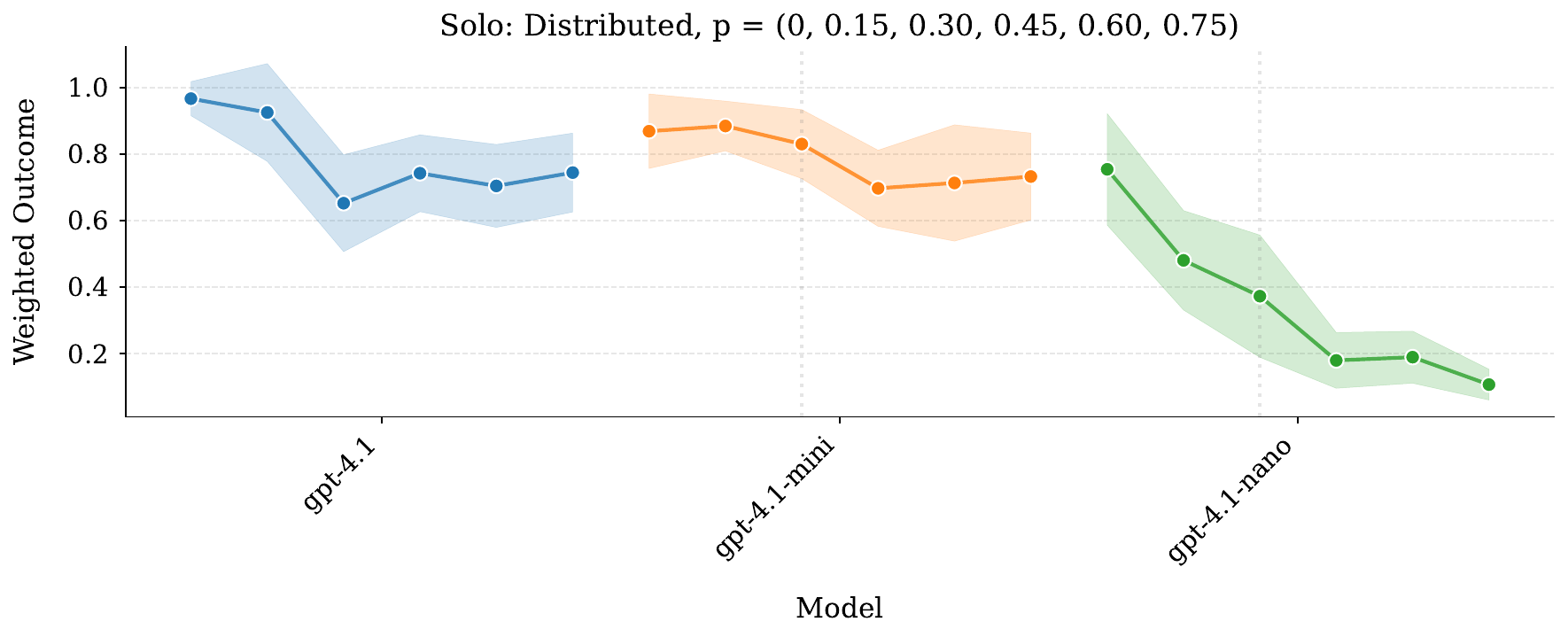} 
        \label{fig:app:solo-dist-wall}
    \end{subfigure}
    \caption{
    \textbf{Solo Weighted Outcomes for Wall Density Ablations.}
    We report mean values with 95\% CI on \maze{6}{6} mazes for $\sim$30 rollouts, varying the wall density $p \in \{0, 0.15, 0.30, 0.45, 0.60, 0.75\}$.
    Note that for the solo full setting (top), gpt-4.1 and gpt-4.1-mini display the most variance in the range $p \in [0.30, 0.45]$, whereas the gpt-4.1-nano performance monotonically degrades as wall density increases.
    For the distributed setting, all three models drop performance as the wall density increases.
    }
    \label{fig:app:wall-ablations}
\end{figure*}

\newpage
\clearpage

\section{Additional results}
\label{app:additional-results}
In this section we start by highlighting a number of model-specific qualitative observations (App. \ref{sec:app:qualitative}).
We then attempt to quantify some of the mechanisms likely to attribute to the identified collaboration gap in App. \ref{sec:app:mechanism}.
Next we show additional weighted and binary outcome results for various interaction pairs in Appendices \ref{app:additional-results:wo} and \ref{app:additional-results:sr} respectively.

We close this section by exploring ``collaborative efficiency'' in App. \ref{app:additional-results:efficiency}, which can be defined as the product of the median tokens per message and the number of messages required per solution.
Because ``thinking'' tokens are not always exposed, this metric reflects information-sharing efficiency rather than computational efficiency.
We report the median message counts conditioned on weighted-outcome ranges to mitigate outcome confounding.
Finally, we focused our heterogeneous pairings on the most widely deployed commercial APIs to establish a baseline, leaving open-weights heterogeneous pairings for future work.

\subsection{Qualitative observations on specific models}
\label{sec:app:qualitative}
\xhdr{command-a/r} Upon qualitative inspection, we noticed that the command-a/r models had a tendency to prematurely ``hallucinate'' about being able to use the completion phrase.
As a result, their collaborative capabilities reported here should be seen as a lower bound on actual performance.

\xhdr{gpt-5} This model displayed superior performance in both solo and homogeneous collaboration mode.
As shown in Figure \ref{fig:app:maze-size-ablations}, it remained capable of solving both full and distributed mazes up to size $12 \times 12$ perfectly, and barely lost performance in homogeneous collaboration for mazes of size $8\times8$ and $10 \times 10$.
As shown in Figure \ref{fig:gpt5-false-failure}, the ``failed'' runs in $6\times6$ mode were due to a parsing shortcoming, rather than a mistake made by the model.

\xhdr{gemini-2.5-pro} Successful models generally share their maze information in the early turns to resolve the missing information.
They further engage in rich grounding to make sure their partner is on the same page.
Instead, many of gemini-2.5-pro's rollouts were marked by short messages (see Figure \ref{fig:app:token-efficiency}).
While this was not a problem when collaborating in homogeneous mode, it did complicate collaborating with weaker models in the heterogeneous setting, e.g., gemini-2.5-flash-lite.
This can be seen in Table \ref{app:tab:all-wo}, where the weaker but more verbose gemini-2.5-flash is capable of reaching higher outcomes when collaborating with gemini-2.5-flash-lite compared to its pro base.

\xhdr{claude-sonnet-4} The model proved itself better at collaborating with other models than with a copy of itself (see Figure \ref{fig:main:collab-heterogeneous}).
Qualitative inspection showed two interesting failure modes: (i) the model got stuck in ``agreement'' loops, where, instead of proposing the next move, it keeps seeking agreement on the previous move; (ii) it ``splits'' up and starts to explore the maze separately.
This is not necessarily incorrect, as this could be interpreted as under specified in the instructions.
It does however break the current parser, leading to suppressed performance (see Figure \ref{fig:msg:sonnet:splits}).

\xhdr{claude-opus-4.1} While a strong average performer, some of its solutions were surprisingly inefficient. For example:
\begin{align*}
&\texttt{@ \# - - - -}\\
&\texttt{o o o o o o}\\
&\texttt{o \# - o o o}\\
&\texttt{\# o o o \# o}\\
&\texttt{\# o \# o \# *}\\
&\texttt{\# o o o \# -}
\end{align*}
where ``\texttt{-}'' are paths, ``\texttt{o}'' cells visited, ``\texttt{\#}'' walls, and ``\texttt{@, *}'' the start and goals states.
It manages to visit almost every path cell on its way to the goal state, using the full 50 turns.

\textbf{Smaller models} Many of the smaller models had a tendency to not share their map view at all, e.g., gpt-4.1-mini, claude-haiku-3.5, opting for a completely local exploration, asking for information as they went instead (see App. \ref{app:sec:mechanism:grounding}).
They also often used ``directions'' to communicate moves instead of coordinates, making it increasingly difficult to track state.

\subsection{Quantifying collaboration mechanisms}
\label{sec:app:mechanism}
The experimental results in Section \ref{sec:results} identify a ``collaboration gap,'' where models perform reliably worse when forced to collaborate compared to solving mazes solo.
Although we highlighted some of the mechanisms associated with this gap based on qualitative inspections, e.g., style imitation, lack of grounding, disagreement resolution strategies, we did not quantify these mechanisms.
In this section we therefore dive deeper into quantifying specific behavioral tendencies and their effect on outcomes.

Note that this is \textbf{not} an exhaustive analysis, but rather a starting point for future work aimed at improving collaborative capabilities.
All features used in these analyses are annotated by LMs, introducing a shared source of measurement noise. While we validate annotation consistency (see below), systematic misreadings could affect all features in correlated ways.
Furthermore, all analyses are observational; we use the language of mediation and prediction throughout, but caution against strong causal interpretations.

\xhdr{Annotation setup}
For a selection of 53 unique homogeneous and heterogeneous interaction pairs, we sample 50 rollouts per pair.
These rollouts involve 13 models from Anthropic, Google, OpenAI, and xAI, chosen to cover different capability substrates, model families, and interaction types. We used the results from Section \ref{sec:results:solo} to assign strength tiers:
\begin{itemize}
    \item \textbf{Strong}: grok-3-mini, grok-4, gemini-2.5-pro, gpt-5, claude-opus-4.1
    \item \textbf{Medium}: grok-3, gemini-2.5-flash, gpt-4.1, claude-sonnet-4
    \item \textbf{Weak}: gemini-2.5-flash-lite, gpt-5-nano, gpt-4.1-mini, claude-haiku-3.5
\end{itemize}
We then use two LMs (gpt-4.1 and gemini-2.5-flash) to annotate each rollout for a variety of features, which are the result of iterative manual inspection aided by model suggestions.
Relevant features will be discussed in each subsection.

\subsubsection{Grounding cascades predict outcome success}
\label{app:sec:mechanism:grounding}
Following the discussion in section \ref{sec:method:communication}, grounding is deemed essential in human collaboration.
We therefore collect the following features to assess if models attempt to ``ground'' their representations of the task and their strategic approach:
\begin{itemize}
    \item \textbf{grounded start and goal}: did models align on the start and goal state?
    \item \textbf{shared map}: did both models share their map?    
    \item \textbf{merged map}: did models attempt to ``merge'' the partial maps?
    \item \textbf{reconciliation timing}: were maps merged up front, gradually, or never?
    \item \textbf{strategy}: did agents use a global strategy, i.e., planning more than 3 moves ahead, or use a strictly local approach?
\end{itemize}

\begin{figure}[t]
    \centerfloat
    \captionsetup{font=\captionFontSize}
    \begin{subfigure}[b]{0.5\textwidth}
        \centerfloat
        \includegraphics[width=\textwidth]{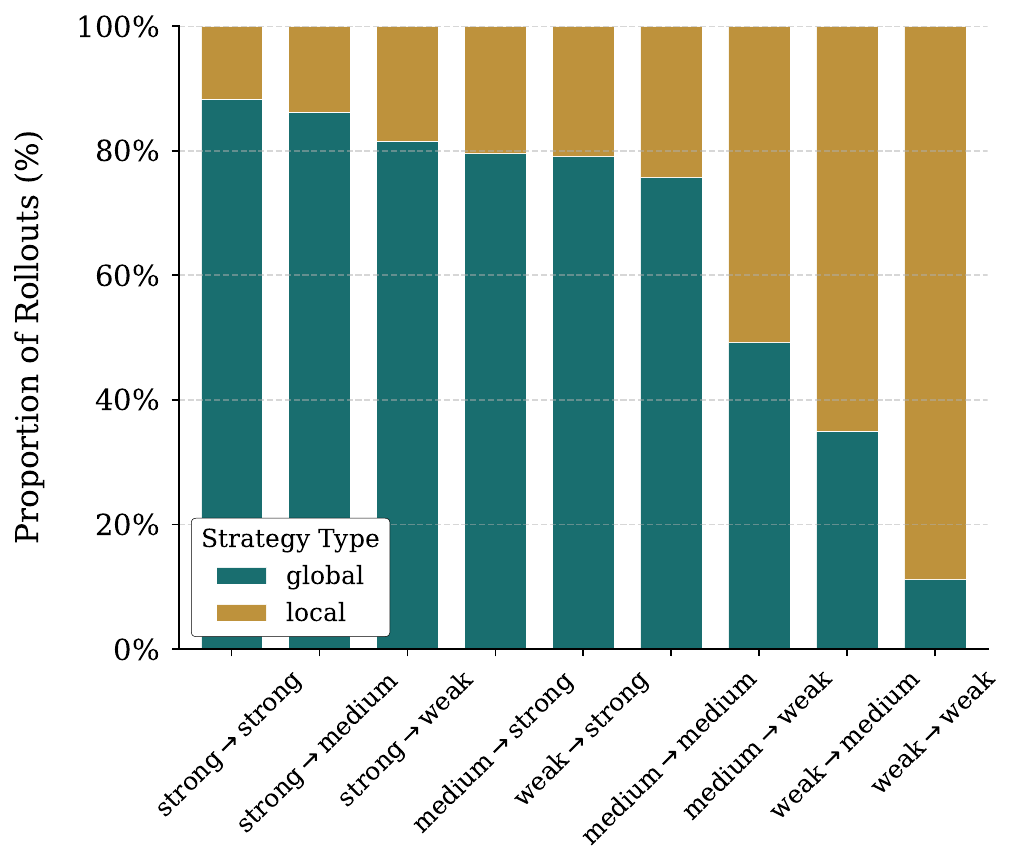} 
        \caption{Strategy by interaction-pair strength}
        \label{fig:app:grounding:tendencies}
    \end{subfigure}
    \hspace{0.5em}
    \begin{subfigure}[b]{0.5\textwidth}
        \centerfloat        
        \includegraphics[width=\textwidth]{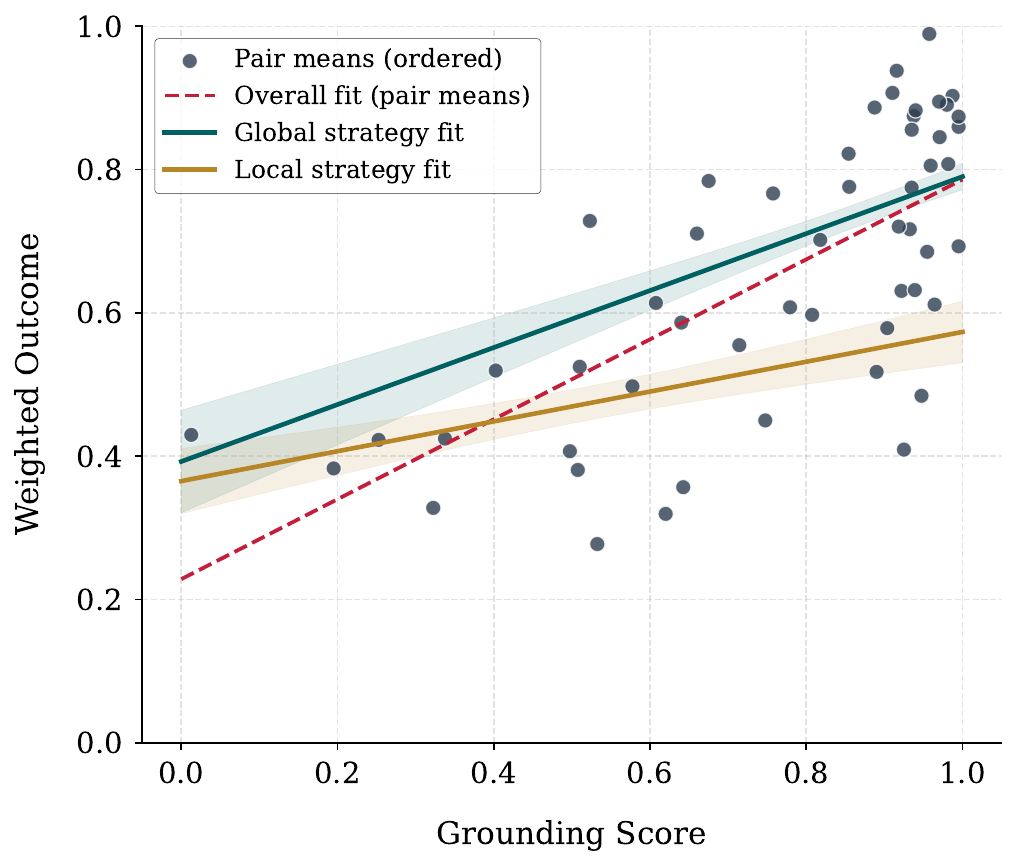} 
        \caption{Weighted outcome vs. grounding score}
        \label{fig:app:grounding:wo}
    \end{subfigure}
    \caption{
    \textbf{Mechanisms influencing the collaboration gap.}
    (a) Proportion of global versus local strategies across interaction pairs stratified by model strength.
    The use of global strategies increases when stronger models are involved.
    (b) Relationship between grounding scores and weighted outcomes. While both strategies improve with grounding, global strategies (green) show a steeper performance curve.
    Red dashed line represents the aggregate trend across all ordered agent pairs; shaded regions indicate 95\% CIs.
    }
    \label{fig:app:grounding}
\end{figure}

Whether agents use a global strategy or not is strongly correlated with weighted outcomes, with global strategies achieving an average weighted outcome of $0.75 \pm ${\small $0.01$} and local ones just $0.47\pm${\small $0.01$}. 
The use of a global strategy is in turn correlated with the strength of the participating models, with global strategies more likely when strong models are involved (see Figure \ref{fig:app:grounding} (a)).
However, a global strategy is only possible once certain ``grounding'' steps have been taken, i.e., at the minimum one of the agents needs access to both maps.
Hence, we are interested in the influence of grounding both in enabling superior strategies and influencing outcomes directly (see Figure \ref{fig:app:path-diagram}).

\begin{figure}[htbp]
    \centerfloat
    \begin{tikzpicture}[>=stealth, node distance=1.5cm and 2.5cm, 
        main/.style={draw, rectangle, fill=white, font=\small, minimum height=0.8cm, minimum width=2.5cm, align=center},
        label/.style={draw=none, fill=none, font=\small},
        aux/.style={draw, ellipse, dashed, font=\small, text=gray, draw=gray, minimum height=0.7cm}]
        
        \node[main] (X) {Grounding\\Score ($X$)};
        \node[main, right=of X] (M) {Global\\Strategy ($M$)};
        \node[main, right=of M] (Y) {Weighted\\Outcome ($Y$)};
        
        \node[aux, above=1.2cm of M] (C) {Controls ($C$)};
        
        \node[aux, below right=1.0cm and 0.5cm of Y, xshift=-2cm] (U) {Unobserved ($U$)};
    
        \node[label, text=blue, font=\itshape\small] (Syn) at ([yshift=1.1cm, xshift=0.5cm]X.center) {Synergy};
    
        \draw[->, thick] (X) -- node[label, above] {$a$} (M);
        \draw[->, thick] (M) -- node[label, above] {$b$} (Y);
        
        \draw[->, thick] (X.south) .. controls +(down:1.1cm) and +(down:1.1cm) .. node[label, below] {$c'$} (Y.south);
    
        \coordinate (mod_point) at ([xshift=1.25cm, yshift=0.4cm]M.east);
        \draw[->, blue, dashed, thick] (X.north) .. controls +(up:0.9cm) and +(up:0.7cm) .. (mod_point);
    
        \draw[->, gray, thin] (C) -- (X);
        \draw[->, gray, thin] (C) -- (M);
        \draw[->, gray, thin] (C) -- (Y);
    
        \draw[->, gray, dashed] (U) -- (M.south east);
        \draw[->, gray, dashed] (U) -- (Y.south);
    \end{tikzpicture}
    \caption{
    \textbf{Path diagram of mechanism influences.} 
    Our goal is to measure the influence of the observed ``grounding behavior'' ($X$) and the choice of strategy type ($M$) on the outcome ($Y$), controlling for a number of factors ($C$) and testing for the robustness against potentially unobserved factors ($U$).
    We test both the direct influence $c'$, and the mediated influence of $X$ through $M$ ($a \to b$).
    We also explore a ``Synergy'' interaction term to measure if the effectiveness of $M$ is contingent on the strength of $X$.
    }
    \label{fig:app:path-diagram}
\end{figure}

As a proxy for grounding strength, we compute a simple grounding score. Let $\mathcal{B}$ be the set of grounding features \{start+goal, share, merge\} and $R \in \{0, 1, 2\}$ the pair-level reconciliation timing score (2 for upfront, 1 for incremental), then
\begin{align} \label{eq:grounding-score}
    X = \frac{1}{8} \left( R + \sum_{i \in \{1, 2\}} \sum_{j \in \mathcal{B}} \mathbb{I}_{i,j} \right),
\end{align}
where $\mathbb{I}_{i,j}$ is an indicator variable that equals 1 if agent $i$ performs behavior $j$.
The denominator 8 normalizes the score to the range $[0, 1]$.
This score empirically shows a statistically significant association with weighted outcomes across the 53 pairs (Pearson $r=0.69$, Spearman $\rho = 0.70$, for both $p < 0.01$) as shown in Figure \ref{fig:app:grounding} (b). 

We start by fitting a simple additive mediation model to test whether grounding behavior is consistent with establishing the ``pre-conditions'' for strategic planning, which in turn influences weighted outcomes (Model 1):
\begin{align}
    M &= i_1 + aX + \gamma_1 C + \epsilon_1 \\
    Y &= i_2 + c'X + bM + \gamma_2 C + \epsilon_2, \label{eq:additive-path}
\end{align}
where $a$ measures how much grounding score $X$ increases the likelihood of adopting a global strategy $M$, $b$ measures the increase in weighted outcome $Y$ provided by the strategy holding grounding constant, and $a \times b$ is the estimated indirect effect (following \citet{imai2010general}), representing the portion of the grounding-outcome association that is consistent with mediation through the strategy.
Finally, $C$ is a vector of controls accounting for agent family, agent strength, interaction type, maze complexity, and the median token usage.
$i$ and $\epsilon$ are intercepts and error terms respectively. 

The second model includes a ``synergy'' interaction term ($X \cdot M$) to test if the strength of $M$ is moderated by the strength of $X$ (Model 2):
\begin{align}
    Y_{syn} &= i_3 + c'X + b_1 M + b_2 (X \cdot M) + \gamma_3 C + \epsilon_3 \label{eq:interaction-path}
\end{align}

\begin{table}[ht]
    \centering
    \begin{tabular}{lcc}
    \toprule
    \textbf{Effect Path} & \textbf{Model 1 (Additive)} & \textbf{Model 2 (Synergistic)} \\
    \midrule
    Path $a$ (Grounding $\to$ Strategy) & $0.6886^{***}$ & $0.6886^{***}$ \\
    Path $b$ (Strategy Effect)           & $0.1683^{***}$ & $-0.0421$ \\
    Synergy ($X \cdot M$)                & ---            & $0.2778^{***}$ \\
    Direct Effect ($c'$)                 & $0.1859^{***}$ & $0.1859^{***}$ \\
    \midrule
    Total Effect ($c$)                   & $0.3018^{***}$ & $0.3018^{***}$ \\
    Indirect Effect ($a \times b$)       & $0.1159^{***}$ & $-0.0290^{\dagger}$ \\
    \textit{Bootstrap 95\% CI}           & $[0.0679, 0.1587]$ & $[-0.1019, 0.0309]$ \\
    \bottomrule
    \multicolumn{3}{l}{\small $^{***}$p < 0.001; $^{\dagger}$Indirect effect in Model 2 is calculated at the grounding intercept ($X=0$).}
    \end{tabular}
    \caption{\textbf{Mediation analysis of grounding and global strategy on weighted outcomes.}
    Model 1 shows the baseline additive mediation and Model 2 incorporates the interaction (Synergy) between grounding and strategic choice. Standard errors are clustered by model pair (N = 53).}
    \label{tab:mediation_results}
\end{table}

Results of clustered bootstrapping on agent pairs ($B=1000$) are shown in Table \ref{tab:mediation_results}.
Model 1 shows robust additive mediation, with grounding score $X$ significantly associated with the propensity of using a global strategy $M$ ($a \approx 0.69, p < 0.001$).
Opting to use a global strategy boosts performance ($b \approx 0.17, p < 0.001$) with approximately $100 \cdot (a \times b) / c \approx 38.4\%$ of the total effect of grounding mediated through the strategic pathway (Indirect effect $\approx 0.12$, CI: [$0.07, 0.16$]).

Incorporating an interaction term in Model 2 reveals a more nuanced relationship between grounding and strategy: the effectiveness of a global strategy appears contingent on grounding levels.
The non-significant and slightly negative strategy path ($b \approx -0.04$, at $X=0$, CI: $[-0.10, 0.03]$) indicates that planning without any grounding provides no performance benefit.
This is not surprising, since planning a global route without access to both maps would indicate either hallucinations or a high-risk gamble.
However, the strong Synergy coefficient ($X \times M \approx 0.28, p < 0.001$) indicates that as grounding level \textit{increases}, the global strategy effectiveness scales with it.
At perfect grounding ($X = 1$), this boost reaches $b_1 + (b_2 \cdot X) \approx 0.24$, effectively doubling the impact observed in the baseline model.

We further conduct a sensitivity analysis following the framework of \cite{imai2010general} to verify that the identified impact is not an artifact of unobserved confounders missed by our controls ($\mathcal{U}$ in Figure \ref{fig:app:path-diagram}).
For Model 1, the mediation effect remains positive and significant for a residual correlation of $\rho < 0.18$ and for Model 2 at $\rho < 0.26$ when evaluated at $X=1$.
These results suggest that the associations between grounding, strategy, and weighted outcome are fairly resilient to potential omitted-variable bias, though we note that the sensitivity thresholds are moderate: a single moderately correlated unobserved confounder could attenuate the mediated pathway.

To assess annotation stability, we first re-graded a random subsample of 250 rollouts using three alternative models (gpt-4.1, gemini-2.5-flash, and gpt-5-mini).
The composite grounding score shows strong inter-grader agreement across all three pairwise comparisons (Pearson $r \in [0.71, 0.79]$).
Agreement on individual binary components ranges from moderate to substantial (Cohen's $\kappa \in [0.39,0.83]$).
We then repeated the full grounding mediation analysis using annotations from gemini-2.5-flash on the entire sample.
All mediation paths replicate with the same sign and significance (path $a \approx 0.48, p < 0.001$; path $b \approx 0.09, p < 0.001$; indirect effect CI 95\% $[0.01, 0.06]$; synergy $p = 0.018$), with coefficient attenuation consistent with the inter-grader agreement reported above.

\xhdr{Limitations}
The grounding score \eqref{eq:grounding-score} assigns equal weight to each component and only covers a subset of grounding features.
Learned or informed weightings might be more appropriate.
The model strength is assigned by the authors; results could shift under alternative groupings.
Finally, despite the sensitivity analysis, all results remain observational and should thus be interpreted as predictive patterns rather than causal mechanisms.

\subsubsection{When and what do agents disagree on?}
\label{app:sec:mechanism:disagreements}

\begin{figure}[t]
    \centerfloat
    \captionsetup{font=\captionFontSize}
        \begin{subfigure}[b]{0.6\textwidth}
        \centerfloat
        \includegraphics[width=\textwidth]{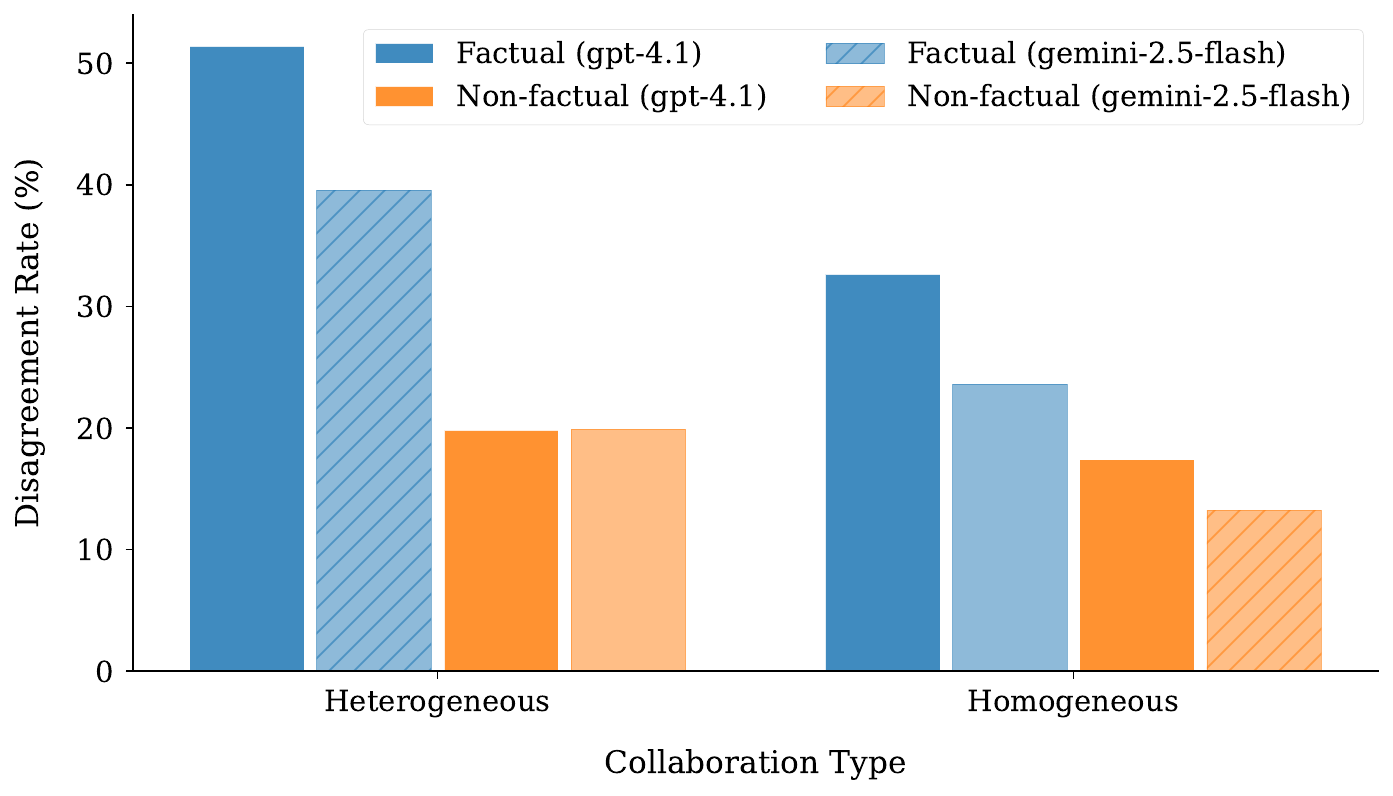} 
        \caption{Disagreement by interaction type}
    \end{subfigure}
    \hspace{2em}
    \begin{subfigure}[b]{0.4\textwidth}
        \centerfloat
        \includegraphics[width=\textwidth]{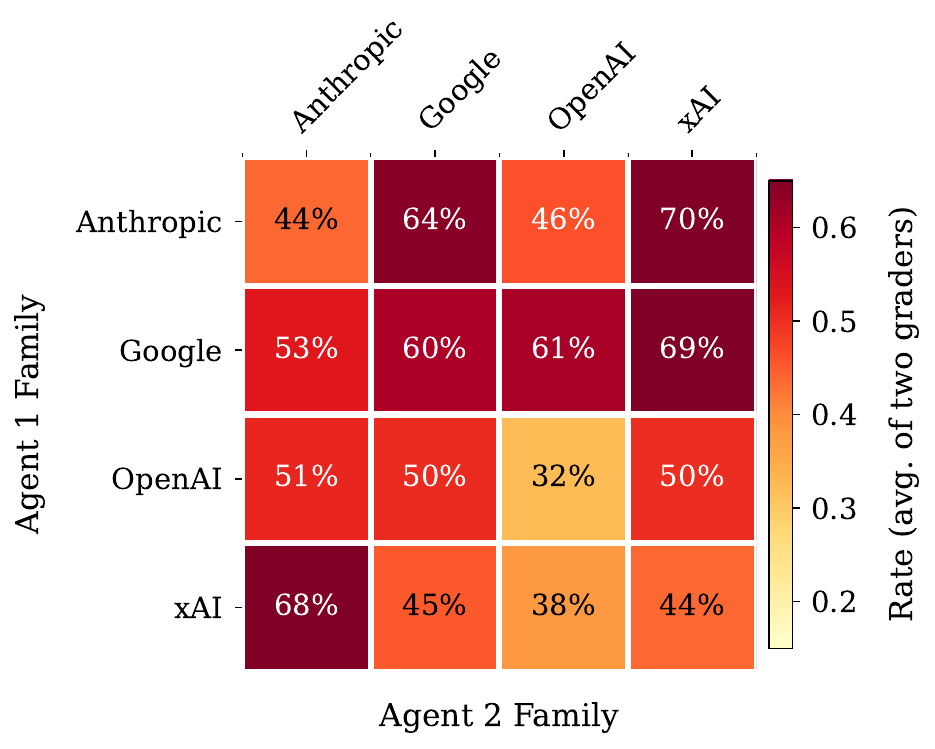} 
        \caption{Encountered disagreement}
    \end{subfigure}
    \caption{
    \textbf{Disagreements by interaction type and between model families.}
    (a) Homogeneous collaborations exhibit a lower rate of (factual) disagreements ($p < 0.01$).
    (b) Models from the same model family have a slightly lower rate of disagreements among themselves and Google models (i.e., Gemini series) are the most combative.
    }
    \label{fig:app:disagreements}
\end{figure}

In this section we examine what types of disagreements agents encounter.
We differentiate between factual disagreements and those based on preference.
The former involves \textit{perceptual disagreements}, where agents disagree on the contents of a cell or the \textit{physical legality} of moves, e.g., the disagreement in Figure \ref{fig:active-correction}.
The latter can be the result of grounding misalignment, e.g., the col--row vs. row--col example of Figure \ref{fig:coordinate-grounding-failure}, or \textit{strategy}, where agents disagree on the direction of a move, without a factual conflict.
Disagreements occurred in approximately half of the annotated rollouts.

We annotate the following features:
\begin{itemize}
    \item \textbf{has factual disagreement}: any disagreements based on perception or move legality
    \item \textbf{has non-factual disagreement}: any disagreements based on grounding or strategy
\end{itemize}
To improve robustness we take the average between two grader LMs (gpt-4.1 and gemini-2.5-flash).
The two graders show high agreement on whether disagreement was encountered ($\kappa = 0.72$) and if there was a factual disagreement ($\kappa = 0.70$).
They are moderately aligned on the occurrence of non-factual disagreement ($\kappa = 0.48$), which is not surprising as non-factual disagreements like grounding and strategy are more nuanced and open for interpretation.

To quantify the drivers of agentic conflict, we model the likelihood of disagreement events using a logistic regression
\begin{equation}
    Y = i + \beta_0 H + \beta_1 F + \sum_{n \in \{1,2\}} \sum_{k \in \mathcal{K}} \beta_{nk} C_{nk} + \epsilon,
\end{equation}
where $Y$ is the disagreement event, $H$ indicates if this concerns a homogeneous interaction, $F$ if both models were from the same model family, and $i$ and $\epsilon$ are intercept and error terms respectively.
The other predictors in $\mathcal{K}$ include the strength and model family of agent 1 and agent 2.

\begin{table}[ht]
\centering
\small
\begin{tabular}{l ccc}
\toprule
\textbf{Predictor} & \textbf{Total Disagreement} & \textbf{Factual} & \textbf{Non-Factual} \\
\midrule
Intercept (Baseline)$^a$ & 1.96* & 3.43** & 0.44** \\
\addlinespace

\textit{Interaction-level Effects} & & & \\
Identical Model ($H$) & 0.60* & 0.66** & 0.97 \\
Same Family ($F$) & 0.83 & 0.92 & 0.97 \\
\addlinespace

\textit{Agent 1 Characteristics ($C_{1k}$)} & & & \\
Weak Starter & 0.69$^\dagger$ & 0.60** & 1.13 \\
Strong Starter & 0.93 & 0.59** & 1.44$^\dagger$ \\
Google Family & 1.30 & 1.91** & 0.88 \\
OpenAI Family & 0.81 & 0.68$^\dagger$ & 1.11 \\
xAI Family & 0.69 & 1.07 & 1.57 \\
\addlinespace

\textit{Agent 2 Characteristics ($C_{2k}$)} & & & \\
Weak Follower & 0.88 & 1.06 & 1.30 \\
Strong Follower & 1.16 & 1.02 & 1.61* \\
Google Family & 1.12 & 1.50$^\dagger$ & 0.77 \\
OpenAI Family & 0.89 & 0.91 & 0.66 \\
xAI Family & 0.98 & 1.11 & 0.58 \\

\midrule
Observations ($N$) & 2628 & 1472 & 1472 \\
\bottomrule
\addlinespace[4pt]
\multicolumn{4}{p{0.9\linewidth}}{\footnotesize $^a$ \textit{Note:} Reference group is Medium Starter, Medium Follower, Anthropic Family. Significance determined via cluster bootstrap ($B=1000$) resampled by unique model pairs.} \\
\multicolumn{4}{l}{\footnotesize Significance: ** $p < 0.01$, * $p < 0.05$, $^\dagger p < 0.10$.}
\end{tabular}
\caption{\textbf{Predictors of agentic conflict.} Odds ratios for disagreement types.}
\label{tab:disagreement_drivers}
\end{table}

Results of clustered bootstrapping on agent pairs ($B = 1000$) are shown in Table \ref{tab:disagreement_drivers}.
We first note that the odds ratio of encountering any conflict at all are significantly lower for homogeneous collaborations (OR = 0.60, $p<0.05$).
Strong models are just as likely as medium models (reference group) to encounter disagreements, while weaker models appear more compliant (OR = 0.69, $p<0.10$).

Factual conflicts are the most likely in medium-strength models.
Strong starting models reduce factual disagreement likelihood by roughly 41\% (OR = 0.59), which could be linked to their superior grounding capabilities leaving less room for confusion.
Weak starting models reduce the likelihood of disagreement by a similar rate.
This could be explained by their lower propensity for disagreement in general.
Homogeneous collaborators are also less likely to disagree on facts (OR = 0.66), whereas Google starters are the most combative --- nearly doubling the odds of a factual dispute (OR = 1.91).

Non-factual conflicts increase with capability, suggesting a ``sophistication shift'' in agent conflict.
Strong followers are 61\% more likely (OR = 1.61) to trigger disputes over strategy or grounding than the medium reference, with strong starters not far behind (OR = 1.44).
Crucially, shared model identity mainly reduces how often conflicts arise, not what they are about: once a disagreement does occur, homogeneous pairs are just as likely as heterogeneous pairs of comparable strength to be disputing strategy or grounding (OR = 0.97).

\xhdr{Limitations} As noted before, identifying disagreement types is challenging, reflected in the inter-grader agreement  ($\kappa = 0.70, \kappa = 0.48$).
Directional findings are likely robust, but precise odds ratios for non-factual conflicts should be interpreted with caution.
The analysis further clusters by 53 unique model pairs with some family $\times$ family pairings sparsely populated, such that odds ratios significant at $p < 0.05$ could be sensitive to individual pairs.

\newpage
\clearpage
\subsection{Weighted outcome}
\label{app:additional-results:wo}

\newlength{\firstcolwidth}

\settowidth{\firstcolwidth}{gemini-2.5-flash-lite}

\begin{table}[!ht]
    \centering

    \begin{subtable}[t]{.52\linewidth}
        \centerfloat
        \small
        \begin{tabular}{p{\firstcolwidth}ccccc}
            \toprule
            & \rotatebox{0}{gemini-2.5-pro} & \rotatebox{0}{gemini-2.5-flash} & \rotatebox{0}{gemini-2.5-flash-lite} & \rotatebox{0}{gpt-4.1} & \rotatebox{0}{gpt-4.1-mini} \\
            \midrule
            gemini-2.5-pro & \textbf{0.84 $\pm$ \scriptsize{0.06}} & \textbf{0.77 $\pm$ \scriptsize{0.10}} & 0.54 $\pm$ \scriptsize{0.11} & - & - \\
            gemini-2.5-flash & \textbf{0.85 $\pm$ \scriptsize{0.09}} & 0.76 $\pm$ \scriptsize{0.05} & \textbf{0.71 $\pm$ \scriptsize{0.06}} & \textbf{0.63 $\pm$ \scriptsize{0.09}} & \textbf{0.58 $\pm$ \scriptsize{0.08}} \\
            gemini-2.5-flash-lite & 0.58 $\pm$ \scriptsize{0.11} & \textbf{0.61 $\pm$ \scriptsize{0.07}} & 0.36 $\pm$ \scriptsize{0.02} & 0.36 $\pm$ \scriptsize{0.07} & 0.32 $\pm$ \scriptsize{0.07} \\
            gpt-4.1 & - & \textbf{0.56 $\pm$ \scriptsize{0.09}} & 0.35 $\pm$ \scriptsize{0.07} & 0.55 $\pm$ \scriptsize{0.04} & 0.50 $\pm$ \scriptsize{0.04} \\
            gpt-4.1-mini & - & \textbf{0.47 $\pm$ \scriptsize{0.08}} & 0.28 $\pm$ \scriptsize{0.06} & 0.42 $\pm$ \scriptsize{0.04} & 0.39 $\pm$ \scriptsize{0.01} \\
            \bottomrule
        \end{tabular}
        \caption{Google, OpenAI}
    \end{subtable}

    \vspace{1em}
    
    \begin{subtable}[t]{.52\linewidth}
        \centerfloat
        \small
        \begin{tabular}{p{\firstcolwidth}cccccc}
            & \rotatebox{0}{gpt-5} & \rotatebox{0}{gpt-5-mini} & \rotatebox{0}{gpt-5-nano} & \rotatebox{0}{o3} & \rotatebox{0}{gpt-4.1} & \rotatebox{0}{gpt-4.1-mini} \\
            \midrule
            gpt-5 & \textbf{0.99 $\pm$ \scriptsize{0.02}} & \textbf{0.95 $\pm$ \scriptsize{0.05}} & \textbf{0.82 $\pm$ \scriptsize{0.09}} & - & - & - \\
            gpt-5-mini & \textbf{0.95 $\pm$ \scriptsize{0.04}} & 0.86 $\pm$ \scriptsize{0.03} & 0.65 $\pm$ \scriptsize{0.11} & 0.84 $\pm$ \scriptsize{0.05} & 0.71 $\pm$ \scriptsize{0.05} & 0.66 $\pm$ \scriptsize{0.05} \\
            gpt-5-nano & \textbf{0.73 $\pm$ \scriptsize{0.10}} & 0.45 $\pm$ \scriptsize{0.11} & 0.38 $\pm$ \scriptsize{0.04} & - & - & - \\
            o3 & - & 0.85 $\pm$ \scriptsize{0.05} & - & \textbf{0.88 $\pm$ \scriptsize{0.03}} & \textbf{0.76 $\pm$ \scriptsize{0.05}} & \textbf{0.77 $\pm$ \scriptsize{0.03}} \\
            gpt-4.1 & - & \textbf{0.76 $\pm$ \scriptsize{0.05}} & - & 0.76 $\pm$ \scriptsize{0.05} & 0.55 $\pm$ \scriptsize{0.04} & 0.50 $\pm$ \scriptsize{0.04} \\
            gpt-4.1-mini & - & 0.58 $\pm$ \scriptsize{0.06} & - & \textbf{0.62 $\pm$ \scriptsize{0.05}} & 0.42 $\pm$ \scriptsize{0.04} & 0.39 $\pm$ \scriptsize{0.01} \\
            \bottomrule
        \end{tabular}
        \caption{OpenAI}
    \end{subtable}

    \vspace{1em}

    \begin{subtable}[t]{.52\linewidth}
        \centerfloat
        \small
        \begin{tabular}{p{\firstcolwidth}ccccc}
 & \rotatebox{0}{grok-4} & \rotatebox{0}{grok-3-mini} & \rotatebox{0}{grok-3} & \rotatebox{0}{gpt-4.1} & \rotatebox{0}{gpt-4.1-mini} \\
\midrule
grok-4 & \textbf{0.92 $\pm$ \scriptsize{0.05}} & \textbf{0.94 $\pm$ \scriptsize{0.05}} & \textbf{0.87 $\pm$ \scriptsize{0.06}} & - & - \\
grok-3-mini & \textbf{0.91 $\pm$ \scriptsize{0.07}} & 0.90 $\pm$ \scriptsize{0.04} & 0.77 $\pm$ \scriptsize{0.11} & \textbf{0.82 $\pm$ \scriptsize{0.06}} & \textbf{0.74 $\pm$ \scriptsize{0.11}} \\
grok-3 & \textbf{0.81 $\pm$ \scriptsize{0.08}} & 0.72 $\pm$ \scriptsize{0.10} & 0.41 $\pm$ \scriptsize{0.04} & 0.37 $\pm$ \scriptsize{0.11} & 0.42 $\pm$ \scriptsize{0.12} \\
gpt-4.1 & - & \textbf{0.66 $\pm$ \scriptsize{0.08}} & 0.47 $\pm$ \scriptsize{0.10} & 0.55 $\pm$ \scriptsize{0.04} & 0.50 $\pm$ \scriptsize{0.04} \\
gpt-4.1-mini & - & 0.42 $\pm$ \scriptsize{0.11} & 0.16 $\pm$ \scriptsize{0.07} & \textbf{0.42 $\pm$ \scriptsize{0.04}} & 0.39 $\pm$ \scriptsize{0.01} \\
            \bottomrule
        \end{tabular}
        \caption{xAI, OpenAI}
    \end{subtable}

    \vspace{1em}

    \begin{subtable}{.52\linewidth}
    \centerfloat
    \small
        \begin{tabular}{lcccc}
         & \rotatebox{0}{grok-3-mini} & \rotatebox{0}{gemini-2.5-flash} & \rotatebox{0}{claude-sonnet-4} & \rotatebox{0}{gpt-4.1} \\
        \midrule
        grok-3-mini & \textbf{0.90 $\pm$ \scriptsize{0.04}} & \textbf{0.88 $\pm$ \scriptsize{0.06}} & \textbf{0.90 $\pm$ \scriptsize{0.05}} & \textbf{0.82 $\pm$ \scriptsize{0.06}} \\
        gemini-2.5-flash & 0.83 $\pm$ \scriptsize{0.06} & 0.76 $\pm$ \scriptsize{0.05} & \textbf{0.86 $\pm$ \scriptsize{0.05}} & 0.68 $\pm$ \scriptsize{0.06} \\
        claude-sonnet-4 & \textbf{0.76 $\pm$ \scriptsize{0.05}} & 0.68 $\pm$ \scriptsize{0.06} & 0.48 $\pm$ \scriptsize{0.05} & 0.54 $\pm$ \scriptsize{0.07} \\
        gpt-4.1 & \textbf{0.66 $\pm$ \scriptsize{0.08}} & 0.63 $\pm$ \scriptsize{0.07} & 0.61 $\pm$ \scriptsize{0.07} & 0.55 $\pm$ \scriptsize{0.04} \\
        \bottomrule
        \end{tabular}
        \caption{Anthropic, Google, OpenAI, xAI}
    \end{subtable}
    
    \vspace{1em}
    
    \begin{subtable}[t]{.52\linewidth}
        \centerfloat
        \small
        \begin{tabular}{p{\firstcolwidth}ccc}
            & \rotatebox{0}{claude-opus-4.1} & \rotatebox{0}{claude-sonnet-4} & \rotatebox{0}{claude-haiku-3.5} \\
            \midrule
            claude-opus-4.1 & \textbf{0.85 $\pm$ \scriptsize{0.04}} & \textbf{0.69 $\pm$ \scriptsize{0.10}} & \textbf{0.58 $\pm$ \scriptsize{0.10}} \\
            claude-sonnet-4 & \textbf{0.63 $\pm$ \scriptsize{0.09}} & 0.48 $\pm$ \scriptsize{0.05} & 0.45 $\pm$ \scriptsize{0.10} \\
            claude-haiku-3.5 & \textbf{0.55 $\pm$ \scriptsize{0.08}} & 0.38 $\pm$ \scriptsize{0.07} & 0.41 $\pm$ \scriptsize{0.06} \\
            \bottomrule
        \end{tabular}
        \caption{Anthropic}
    \end{subtable}

    \vspace{1em}

    \begin{subtable}[t]{.52\linewidth}
        \centerfloat
        \small
        \begin{tabular}{p{\firstcolwidth}ccc}
            & \rotatebox{0}{command-a} & \rotatebox{0}{command-r} & \rotatebox{0}{command-r7b} \\
            \midrule
            command-a & \textbf{0.38 $\pm$ \scriptsize{0.04}} & \textbf{0.34 $\pm$ \scriptsize{0.07}} & \textbf{0.20 $\pm$ \scriptsize{0.11}} \\
            command-r & \textbf{0.33 $\pm$ \scriptsize{0.06}} & 0.20 $\pm$ \scriptsize{0.02} & 0.15 $\pm$ \scriptsize{0.06} \\
            command-r7b & \textbf{0.25 $\pm$ \scriptsize{0.07}} & 0.13 $\pm$ \scriptsize{0.04} & 0.16 $\pm$ \scriptsize{0.04} \\
            \bottomrule
        \end{tabular}
        \caption{Cohere}
    \end{subtable}

    \caption{\textbf{Weighted Outcomes for Collaborative Performance.} We report mean performance with 95\% CI. Max values per row/column are bold.}
    \label{app:tab:all-wo}
\end{table}

\newpage
\clearpage
\subsection{Binary success rate}
\label{app:additional-results:sr}

\settowidth{\firstcolwidth}{gemini-2.5-flash-lite}

\begin{table}[!ht]
    \centering

    \begin{subtable}[t]{.52\linewidth}
        \centerfloat
        \small
         \begin{tabular}{lccccc}
        \toprule
         & \rotatebox{0}{gemini-2.5-pro} & \rotatebox{0}{gemini-2.5-flash} & \rotatebox{0}{gemini-2.5-flash-lite} & \rotatebox{0}{gpt-4.1} & \rotatebox{0}{gpt-4.1-mini} \\
        \midrule
        gemini-2.5-pro & \textbf{0.74 $\pm$ \scriptsize{0.09}} & \textbf{0.70 $\pm$ \scriptsize{0.13}} & 0.34 $\pm$ \scriptsize{0.13} & - & - \\
        gemini-2.5-flash & \textbf{0.78 $\pm$ \scriptsize{0.12}} & 0.64 $\pm$ \scriptsize{0.07} & \textbf{0.53 $\pm$ \scriptsize{0.08}} & \textbf{0.49 $\pm$ \scriptsize{0.10}} & \textbf{0.40 $\pm$ \scriptsize{0.10}} \\
        gemini-2.5-flash-lite & 0.36 $\pm$ \scriptsize{0.13} & \textbf{0.39 $\pm$ \scriptsize{0.08}} & 0.04 $\pm$ \scriptsize{0.02} & 0.13 $\pm$ \scriptsize{0.07} & 0.10 $\pm$ \scriptsize{0.06} \\
        gpt-4.1 & - & \textbf{0.38 $\pm$ \scriptsize{0.10}} & 0.14 $\pm$ \scriptsize{0.07} & 0.23 $\pm$ \scriptsize{0.05} & 0.18 $\pm$ \scriptsize{0.05} \\
        gpt-4.1-mini & - & \textbf{0.26 $\pm$ \scriptsize{0.09}} & 0.07 $\pm$ \scriptsize{0.05} & 0.09 $\pm$ \scriptsize{0.04} & 0.01 $\pm$ \scriptsize{0.01} \\
        \bottomrule
        \end{tabular}
        \caption{Google, OpenAI}
    \end{subtable}

    \vspace{1em}
    
    \begin{subtable}[t]{.52\linewidth}
        \centerfloat
        \small
         \begin{tabular}{lcccccc}
         & \rotatebox{0}{gpt-5} & \rotatebox{0}{gpt-5-mini} & \rotatebox{0}{gpt-5-nano} & \rotatebox{0}{o3} & \rotatebox{0}{gpt-4.1} & \rotatebox{0}{gpt-4.1-mini} \\
        \midrule
        gpt-5 & \textbf{0.97 $\pm$ \scriptsize{0.02}} & \textbf{0.85 $\pm$ \scriptsize{0.07}} & \textbf{0.65 $\pm$ \scriptsize{0.10}} & - & - & - \\
        gpt-5-mini & \textbf{0.80 $\pm$ \scriptsize{0.08}} & 0.67 $\pm$ \scriptsize{0.05} & 0.33 $\pm$ \scriptsize{0.10} & 0.54 $\pm$ \scriptsize{0.10} & 0.36 $\pm$ \scriptsize{0.08} & 0.36 $\pm$ \scriptsize{0.08} \\
        gpt-5-nano & \textbf{0.51 $\pm$ \scriptsize{0.10}} & 0.19 $\pm$ \scriptsize{0.08} & 0.05 $\pm$ \scriptsize{0.03} & - & - & - \\
        o3 & - & 0.59 $\pm$ \scriptsize{0.10} & - & \textbf{0.66 $\pm$ \scriptsize{0.06}} & \textbf{0.48 $\pm$ \scriptsize{0.08}} & \textbf{0.52 $\pm$ \scriptsize{0.05}} \\
        gpt-4.1 & - & \textbf{0.54 $\pm$ \scriptsize{0.08}} & - & 0.48 $\pm$ \scriptsize{0.08} & 0.23 $\pm$ \scriptsize{0.05} & 0.18 $\pm$ \scriptsize{0.05} \\
        gpt-4.1-mini & - & 0.29 $\pm$ \scriptsize{0.07} & - & \textbf{0.31 $\pm$ \scriptsize{0.07}} & 0.09 $\pm$ \scriptsize{0.04} & 0.01 $\pm$ \scriptsize{0.01} \\
        \bottomrule
        \end{tabular}
        \caption{OpenAI} \label{tab:gpt-sr}
    \end{subtable}

    \vspace{1em}

    \begin{subtable}[t]{.52\linewidth}
        \centerfloat
        \small
        \begin{tabular}{lccccc}
 & \rotatebox{0}{grok-4} & \rotatebox{0}{grok-3-mini} & \rotatebox{0}{grok-3} & \rotatebox{0}{gpt-4.1} & \rotatebox{0}{gpt-4.1-mini} \\
\midrule
grok-4 & \textbf{0.84 $\pm$ \scriptsize{0.08}} & \textbf{0.87 $\pm$ \scriptsize{0.10}} & \textbf{0.70 $\pm$ \scriptsize{0.13}} & - & - \\
grok-3-mini & \textbf{0.83 $\pm$ \scriptsize{0.11}} & 0.82 $\pm$ \scriptsize{0.06} & 0.67 $\pm$ \scriptsize{0.13} & \textbf{0.69 $\pm$ \scriptsize{0.09}} & \textbf{0.63 $\pm$ \scriptsize{0.14}} \\
grok-3 & \textbf{0.58 $\pm$ \scriptsize{0.14}} & 0.57 $\pm$ \scriptsize{0.14} & 0.13 $\pm$ \scriptsize{0.04} & 0.22 $\pm$ \scriptsize{0.12} & 0.26 $\pm$ \scriptsize{0.13} \\
gpt-4.1 & - & \textbf{0.49 $\pm$ \scriptsize{0.10}} & 0.23 $\pm$ \scriptsize{0.13} & 0.23 $\pm$ \scriptsize{0.05} & 0.18 $\pm$ \scriptsize{0.05} \\
gpt-4.1-mini & - & \textbf{0.18 $\pm$ \scriptsize{0.11}} & 0.00 $\pm$ \scriptsize{0.00} & 0.09 $\pm$ \scriptsize{0.04} & 0.01 $\pm$ \scriptsize{0.01} \\
        \bottomrule
        \end{tabular}
        \caption{xAI, OpenAI}
    \end{subtable}

    \vspace{1em}
    
    \begin{subtable}[t]{.52\linewidth}
        \centerfloat
        \small
        \begin{tabular}{lcccc}
         & \rotatebox{0}{grok-3-mini} & \rotatebox{0}{gemini-2.5-flash} & \rotatebox{0}{claude-sonnet-4} & \rotatebox{0}{gpt-4.1} \\
        \midrule
        grok-3-mini & \textbf{0.82 $\pm$ \scriptsize{0.06}} & \textbf{0.78 $\pm$ \scriptsize{0.09}} & \textbf{0.82 $\pm$ \scriptsize{0.08}} & \textbf{0.69 $\pm$ \scriptsize{0.09}} \\
        gemini-2.5-flash & 0.72 $\pm$ \scriptsize{0.09} & 0.64 $\pm$ \scriptsize{0.07} & \textbf{0.75 $\pm$ \scriptsize{0.07}} & 0.54 $\pm$ \scriptsize{0.08} \\
        claude-sonnet-4 & \textbf{0.57 $\pm$ \scriptsize{0.07}} & 0.54 $\pm$ \scriptsize{0.08} & 0.23 $\pm$ \scriptsize{0.06} & 0.25 $\pm$ \scriptsize{0.09} \\
        gpt-4.1 & \textbf{0.49 $\pm$ \scriptsize{0.10}} & 0.42 $\pm$ \scriptsize{0.08} & 0.33 $\pm$ \scriptsize{0.09} & 0.23 $\pm$ \scriptsize{0.05} \\
        \bottomrule
        \end{tabular}
    \caption{Anthropic, Google, OpenAI, xAI}
    \end{subtable}
    
    \vspace{1em}
    
    \begin{subtable}[t]{.52\linewidth}
        \centerfloat
        \small
        \begin{tabular}{lccc}
         & \rotatebox{0}{claude-opus-4.1} & \rotatebox{0}{claude-sonnet-4} & \rotatebox{0}{claude-haiku-3.5} \\
        \midrule
        claude-opus-4.1 & \textbf{0.72 $\pm$ \scriptsize{0.07}} & \textbf{0.48 $\pm$ \scriptsize{0.14}} & \textbf{0.29 $\pm$ \scriptsize{0.13}} \\
        claude-sonnet-4 & \textbf{0.33 $\pm$ \scriptsize{0.13}} & 0.23 $\pm$ \scriptsize{0.06} & 0.22 $\pm$ \scriptsize{0.12} \\
        claude-haiku-3.5 & \textbf{0.20 $\pm$ \scriptsize{0.11}} & 0.06 $\pm$ \scriptsize{0.07} & 0.01 $\pm$ \scriptsize{0.02} \\
        \bottomrule
        \end{tabular}
        \caption{Anthropic}
    \end{subtable}

    \vspace{1em}

    \begin{subtable}[t]{.52\linewidth}
        \centerfloat
        \small
\begin{tabular}{lccc}
 & \rotatebox{0}{command-a} & \rotatebox{0}{command-r} & \rotatebox{0}{command-r7b} \\
\midrule
command-a & \textbf{0.02 $\pm$ \scriptsize{0.02}} & \textbf{0.00 $\pm$ \scriptsize{0.00}} & \textbf{0.05 $\pm$ \scriptsize{0.10}} \\
command-r & \textbf{0.00 $\pm$ \scriptsize{0.00}} & \textbf{0.00 $\pm$ \scriptsize{0.00}} & \textbf{0.00 $\pm$ \scriptsize{0.00}} \\
command-r7b & \textbf{0.00 $\pm$ \scriptsize{0.00}} & \textbf{0.00 $\pm$ \scriptsize{0.00}} & \textbf{0.00 $\pm$ \scriptsize{0.00}} \\
\bottomrule
\end{tabular}
        \caption{Cohere}
    \end{subtable}

    \caption{\textbf{Binary Success Rates for Collaborative Performance.} We report mean performance with 95\% CI. Max values per row/column are bold.
    }
    \label{app:tab:all-sr}
\end{table}

\newpage
\clearpage

\subsection{Communication efficiency}
\label{app:additional-results:efficiency}

\begin{figure*}[ht!]
    \centerfloat
    \begin{subfigure}[b]{0.95\textwidth}
        \centerfloat
        \includegraphics[width=\textwidth]{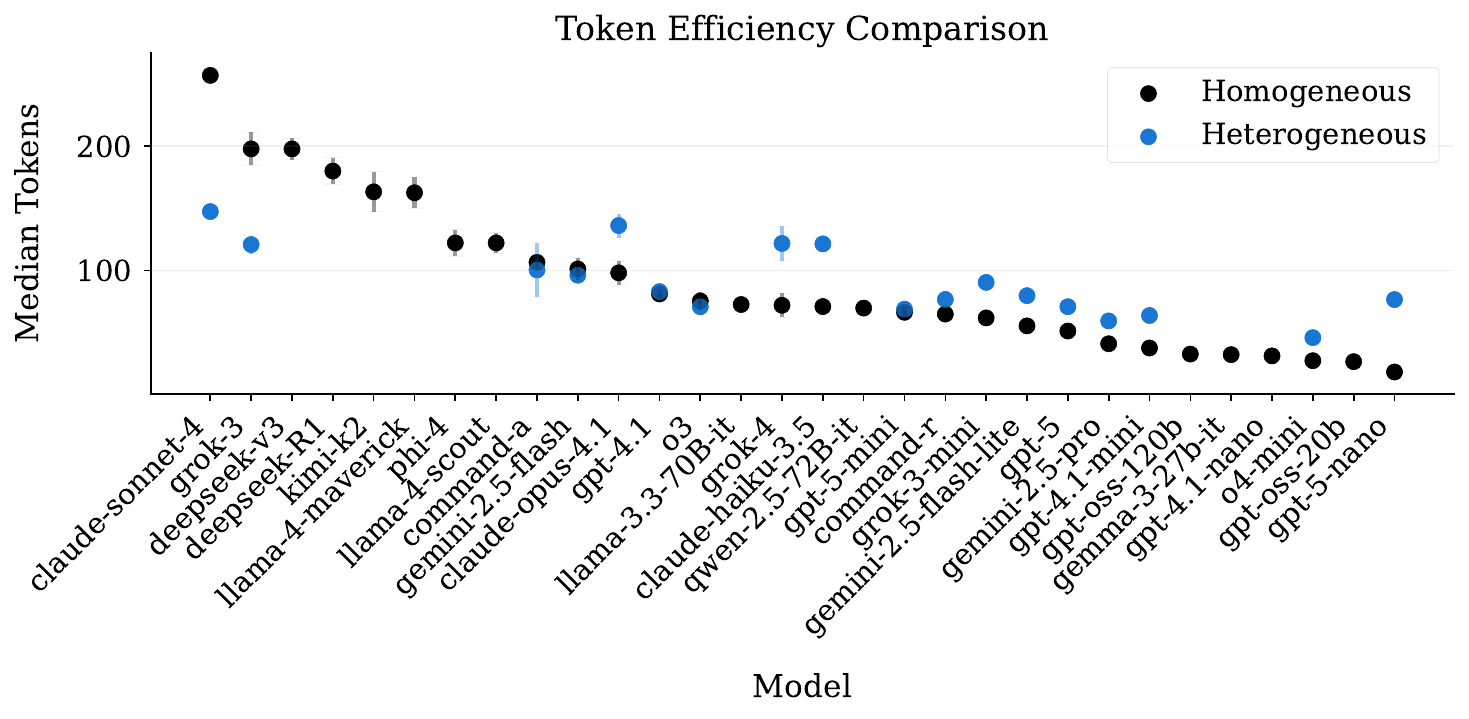} 
        \caption{\textbf{Homogeneous vs. Heterogeneous Token Efficiency.} We display median token efficiency of homogeneous and heterogeneous collaborations (when available).}
    \vspace{2em}
    \end{subfigure}
    \begin{subfigure}[b]{0.95\textwidth}
        \centerfloat
        \includegraphics[width=\textwidth]{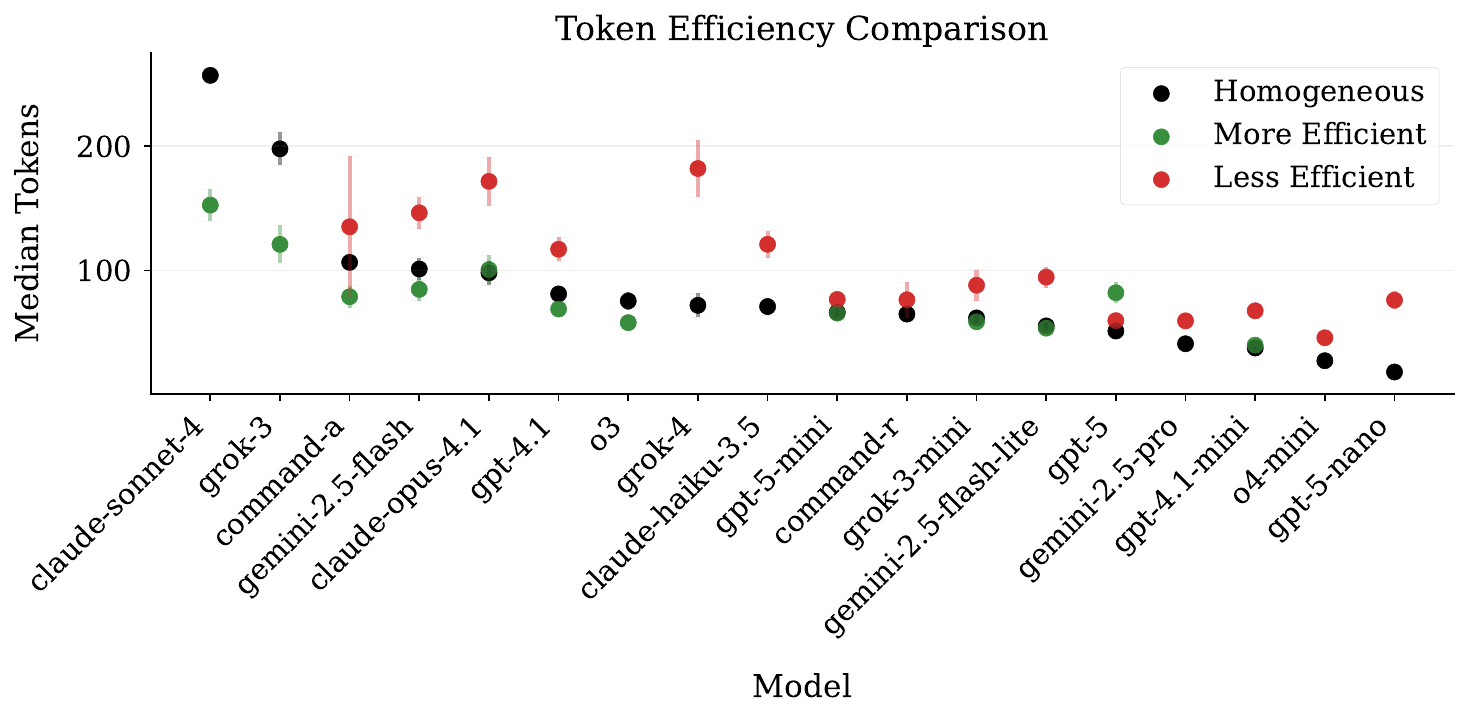} 
        \caption{\textbf{Token Efficiency Stratified by Other Agent.} We compare the change in median token efficiency when an agent is paired with either a ``More'' or ``Less'' efficient agent.
        ``More/Less'' here refers to agents that use fewer/more tokens in their homogeneous collaboration.
        Note that on average, agents tend to adapt their efficiency to their partner.}
    \end{subfigure}
    \caption{
    \textbf{Median Token Efficiency.}
    We report mean values with 95\% CI.
    }
    \label{fig:app:token-efficiency}
\end{figure*}

\begin{figure*}[ht!]
    \vspace{2em}
    \centerfloat
    \begin{subfigure}[b]{1.0\textwidth}
        \centerfloat
        \includegraphics[width=\textwidth]{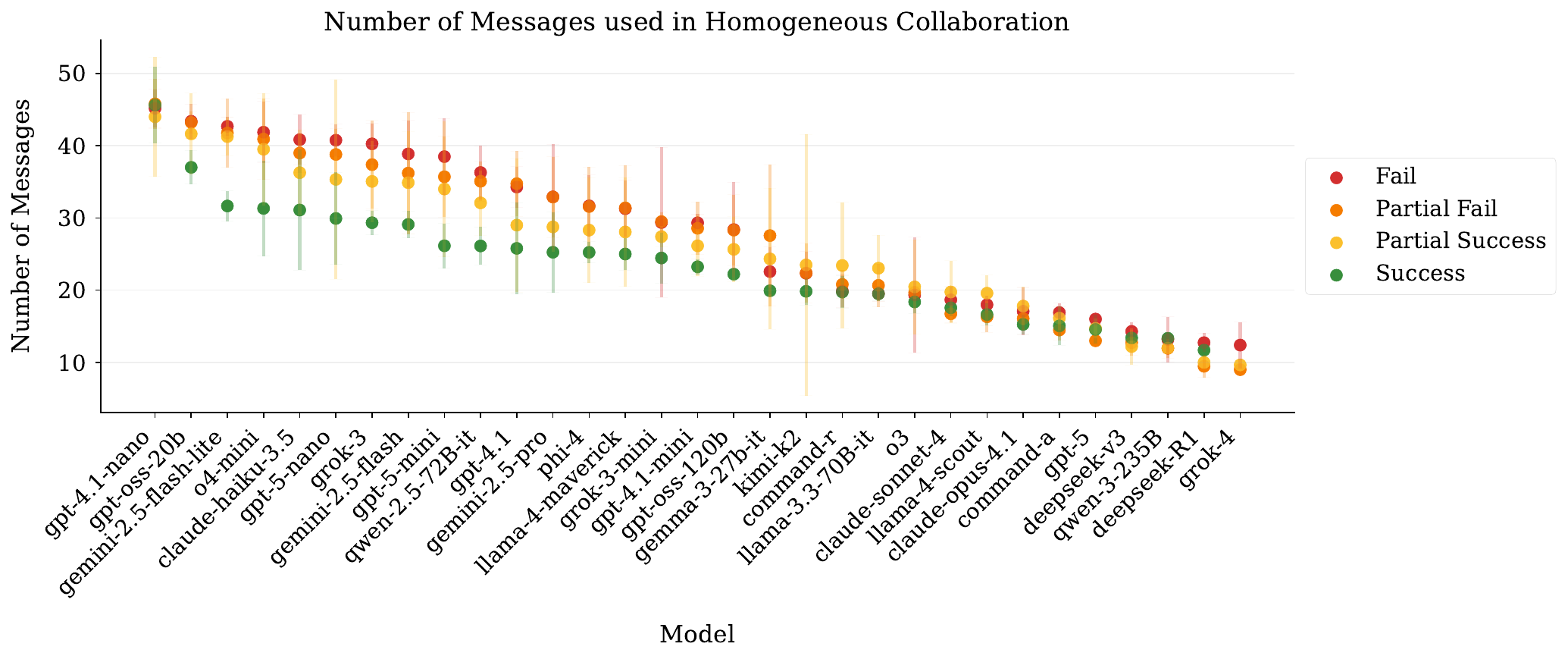} 
        \caption{\textbf{Homogeneous Collaboration.} 
        }
    \vspace{2em}
    \end{subfigure}    
    \begin{subfigure}[b]{1.0\textwidth}
        \centerfloat
        \includegraphics[width=\textwidth]{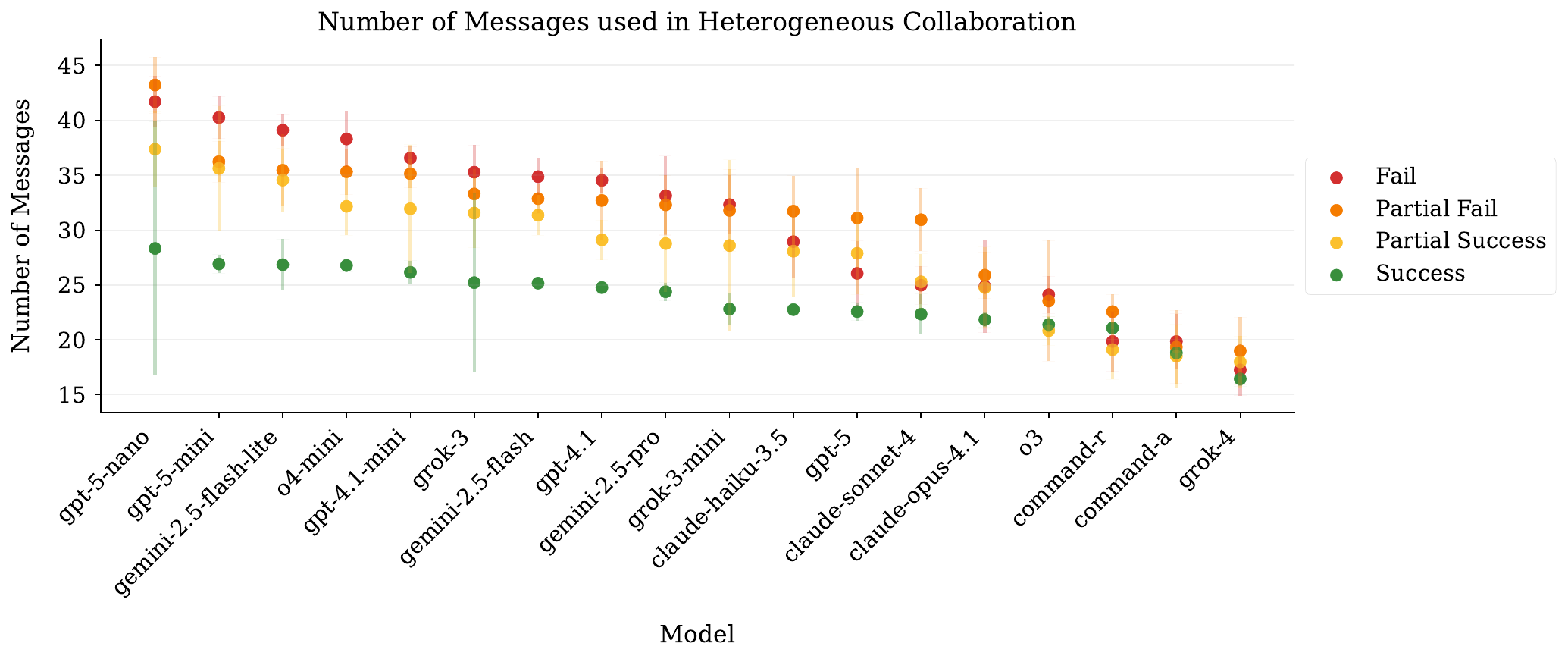} 
        \caption{\textbf{Heterogeneous Collaboration.}}
    \end{subfigure}
    \caption{
    \textbf{Message Efficiency.} 
    We stratify weighted outcome into the following intervals: \textbf{Fail}: 0 to 0.25, \textbf{Partial Fail}: 0.25 to 0.50, \textbf{Partial Success}: 0.5 to 0.75, and \textbf{Success}: 0.75 to 1.0.
    This allows us to compare how many messages each model requires on average to achieve similar outcome ranges.
    We report mean values with 95\% CI.
    }
    \label{fig:app:number-messages}
\end{figure*}

One potential confounder for the collaboration gap is the increased context length of multi-turn dialogues. However, as shown in Figures \ref{fig:app:token-efficiency} and \ref{fig:app:number-messages}, interactions typically conclude within 20 to 40 messages with median lengths under 100 tokens, resulting in total context lengths well within the capacity of modern models.
Furthermore, Figure \ref{fig:app:number-messages} demonstrates that failure outcomes rarely stem from hitting the 50-turn timeout limit.

\subsection*{Median token usage}
\label{app:additional-results:mtu}

\newpage
\clearpage

\subsection*{Median number of messages}
\label{app:additional-results:mnm}

\newpage
\clearpage

\section{Grading ablation}
\label{app:grading}
We conduct grading evaluations of solo and homogeneous collaborations on selected models with different strengths and from different builders: 
\begin{itemize}
    \item \textbf{xAI}: grok-3, grok-3-mini
    \item \textbf{Google}: gemini-2.5-pro, gemini-2.5-flash, gemini-2.5-flash-lite
    \item \textbf{Anthropic}: claude-opus-4.1, claude-sonnet-4.0
    \item \textbf{OpenAI}: gpt-5, gpt-5-nano, gpt-4.1, gpt-4.1-mini, o3
    \item \textbf{Cohere}: command-a
    \item \textbf{Moonshot AI}: kimi-k2
    \item \textbf{Alibaba}: qwen-3-235B
\end{itemize}

For the solo setting, we randomly select 10 samples, grouping by full or distributed maze visibility and outcome success, for 40 samples in total per model.
In the homogeneous collaborative setting we randomly select 20 samples, grouped by outcome success, for 40 samples in total per model.
In both settings, we sample without replacement.
This means that the total number of samples might be less than 40, e.g., if a strong model has fewer than 20 failure cases.

After generating our sample datasets, we collect multiple independent grading evaluations per sample from different grader models using a temperature of 0.5. 
Prompts used for grading solo and collaborative outcomes can be found in Section \ref{app:prompts:verification}.
We are interested in the following ablations on the binary success rates and weighted outcomes of grades:

\xhdr{Overall} We report the Intraclass Correlation Coefficient (ICC) for weighted outcomes and Fleiss' Kappa for binary success rates.

\xhdr{Pairwise correlations} We report pairwise correlations between grading models.

\xhdr{Majority vote vs reference}
We evaluate if majority voting compared to the reference vote would change outcomes. 
We use McNemar's test to compare the difference in binary success outcomes.
For the weighted outcome results, we use a paired t-test (or a Wilcoxon signed-rank test if normality is violated) reporting p-values and Cohen's d effect size.

\xhdr{Conditional analysis}
We compare if regrading proportionally affects successful and failure cases. 
We first split our sample dataset into two groups based on the binary success verdict of the reference grader.
Then, we record for each sample if there was a ``binary disagreement'', i.e., did one of the graders disagree with the outcome?
Next, we use the Fisher's exact test to compare the per-sample disagreement rate between groups.
We also compare differences in weighted outcomes per group, reporting the standard deviation and range of disagreement.
Finally, we perform a t-test and compute Cohen's d on the difference in standard deviations of weighted outcomes.

\xhdr{Agent-pair effect} Finally, we stratify by model to check for model-specific biases.

\subsection{Intra-model consistency}
We collect two additional grading evaluations by gpt-4.1, for a total of three evaluations per sample observation.

\subsubsection*{Solo}
The three independent runs of the same gpt-4.1 grader are highly reliable, with ICC$\simeq$0.99 and Fleiss' Kappa$\simeq$0.99.
There is a small, symmetric, systematic bias for weighted outcomes conditioned on binary outcome, i.e., regression to the mean (Table \ref{solo:intra:cond}).
We observe no bias across agents in Table \ref{tab:solo:intra:agents}, with only qwen-3-235B ($\kappa$ = 0.92, ICC=0.91) being slightly contentious.
The values for gpt-5 and o3 only constitute successful outcomes with no disagreement (manually checked by authors).
Overall, we can conclude that gpt-4.1 is sufficiently consistent for grading solo experiments.

\begin{table}[ht!]
\centering 
\begin{subtable}[t]{0.45\linewidth}
    \centerfloat
    \small
    \begin{tabular}{lr}
    \toprule
    \textbf{Metric} & \textbf{Value} \\
    \midrule
    Weighted Outcome (p-value)     & 0.01   \\
    Binary Outcome (p-value)       & 1.00   \\
    Cohen's d (weighted)           & 0.05   \\
    Weighted Difference in Mean    & 0.00   \\
    Weighted Difference CI (Lower) & -0.00  \\
    Weighted Difference CI (Upper) & 0.00   \\
    \bottomrule
    \end{tabular}
    \caption{Majority Vote vs. Reference}
    \label{tab:solo:intra:majority}
\end{subtable}
\hfill
\begin{subtable}[t]{0.45\linewidth}
    \centerfloat
    \small
\begin{tabular}{lcc}
\toprule
\textbf{Metric} & \textbf{Success} & \textbf{Failure} \\
\midrule
Number of Outcomes           & 280      & 193      \\
Binary Disagreement Rate     & 0.00     & 0.01     \\
Mean Score (Std. Dev.)       & 0.00     & 0.01     \\
Mean Score (Range)           & 0.00     & 0.01     \\
Binary Outcome (p-value)     & \multicolumn{2}{c}{0.17} \\
Weighted Outcome (p-value)   & \multicolumn{2}{c}{0.02} \\
Cohen's d (weighted)         & \multicolumn{2}{c}{$\pm$ 0.19}     \\
\bottomrule
\end{tabular}
    \caption{Conditional Analysis}
    \label{solo:intra:cond}
    \vspace{1em}
    \end{subtable}
    \begin{subtable}{1.\linewidth}
        \centerfloat
    \scriptsize 
    \setlength{\tabcolsep}{4pt}
    
    \begin{tabular}{l *{14}{c}}
    & \rotatebox{75}{qwen3-235B}
    & \rotatebox{75}{claude-opus}
    & \rotatebox{75}{command-a}
    & \rotatebox{75}{gemini-2.5-flash}
    & \rotatebox{75}{gemini-2.5-flash-lite}
    & \rotatebox{75}{gemini-2.5-pro}
    & \rotatebox{75}{gpt-4.1-mini}
    & \rotatebox{75}{gpt-4.1}
    & \rotatebox{75}{gpt-5-nano}
    & \rotatebox{75}{gpt-5}
    & \rotatebox{75}{grok-3}
    & \rotatebox{75}{grok-3-mini}
    & \rotatebox{75}{kimi-k2}
    & \rotatebox{75}{o3} \\
    \midrule
    ICC ($\rho$)       & 0.91 & 1.00 & 1.00 & 1.00 & 1.00 & 1.00 & 1.00 & 1.00 & 1.00 & 1.00 & 1.00 & 1.00 & 1.00 & 1.00 \\
    Kappa ($\kappa$)   & 0.92 & 1.00 & 1.00 & 1.00 & 1.00 & 1.00 & 1.00 & 1.00 & 1.00 & 1.00 & 1.00 & 1.00 & 1.00 & 1.00 \\
    N                  & 36   & 40   & 40   & 33   & 40   & 29   & 32   & 39   & 30   & 20  & 40   & 34   & 40   & 20  \\
    \bottomrule
    \end{tabular}
    \caption{Agent-Pair Effects}
    \label{tab:solo:intra:agents}
    \end{subtable}
\caption{\textbf{Intra-Grader Consistency for Solo Outcomes.}}
\label{tab:grading:solo:intra}
\end{table}

\subsubsection*{Homogeneous collaboration}
We have high overall agreement between graders, ICC$\simeq$0.89 and Fleiss' Kappa$\simeq$ 0.87. Independent evaluations of the same grader are highly correlated with pairwise correlations between 0.86 and 0.91.
Majority votes are statistically indistinguishable from the reference, see Table \ref{tab:collab:intra:majority}.
Conditional analysis shows that repeated scoring compresses scores towards the mean in a modest, but statistically significant manner.
Agent-specific evaluation shows high overall agreement for most models, e.g., ICC>0.80 and $\kappa$>0.6.
The graders show slightly more disagreement for Cohere's command-a and OpenAI's gpt-5.
Overall we conclude that gpt-4.1 is a consistent scorer with low bias towards particular agents.

\begin{table}[!ht]
\centerfloat
\begin{subtable}[t]{0.45\linewidth}
    \centerfloat
    \small
    \begin{tabular}{lr}
    \toprule
    \textbf{Metric} & \textbf{Value} \\
    \midrule
    Weighted Outcome (p-value)     & 0.07   \\
    Binary Outcome (p-value)       & 0.23   \\
    Cohen's d (weighted)           & 0.03   \\
    Weighted Difference in Mean    & 0.00   \\
    Weighted Difference CI (Lower) & -0.01  \\
    Weighted Difference CI (Upper) & 0.01   \\
    \bottomrule
    \end{tabular}
    \caption{Majority Vote vs. Reference}
    \label{tab:collab:intra:majority}
\end{subtable}
\hfill 
\begin{subtable}[t]{0.45\linewidth}
    \centerfloat
    \small
    \begin{tabular}{lrr}
    \toprule
    \textbf{Metric} & \textbf{Success} & \textbf{Failure} \\
    \midrule
    Number of Outcomes           & 221      & 264      \\
    Binary Disagreement Rate     & 0.14     & 0.06     \\
    Mean Score (Std. Dev.)       & 0.02     & 0.05     \\
    Mean Score (Range)           & 0.05     & 0.10     \\
    Binary Outcome (p-value)     & \multicolumn{2}{c}{0.00}     \\
    Weighted Outcome (p-value)   & \multicolumn{2}{c}{0.01}    \\
    Cohen's d (weighted)         & \multicolumn{2}{c}{$\pm$ 0.26}     \\
    \bottomrule
    \end{tabular}
    \caption{Conditional Analysis}
    \label{tab:collab:intra:cond}
    \vspace{1em}
\end{subtable}
\begin{subtable}{1.\linewidth}
    \centerfloat
    \scriptsize 
    \setlength{\tabcolsep}{4pt}
    \begin{tabular}{l *{14}{c}}
    & \rotatebox{75}{qwen3-235B}
    & \rotatebox{75}{claude-opus}
    & \rotatebox{75}{command-a}
    & \rotatebox{75}{gemini-2.5-flash}
    & \rotatebox{75}{gemini-2.5-flash-lite}
    & \rotatebox{75}{gemini-2.5-pro}
    & \rotatebox{75}{gpt-4.1-mini}
    & \rotatebox{75}{gpt-4.1}
    & \rotatebox{75}{gpt-5-nano}
    & \rotatebox{75}{gpt-5}
    & \rotatebox{75}{grok-3}
    & \rotatebox{75}{grok-3-mini}
    & \rotatebox{75}{kimi-k2}
    & \rotatebox{75}{o3} \\
    \midrule
    ICC ($\rho$)       & 0.87 & 0.95 & 0.72 & 0.85 & 0.90 & 0.88 & 0.96 & 0.92 & 0.93 & 0.69 & 0.94 & 0.78 & 0.97 & 0.89 \\
    Kappa ($\kappa$)   & 0.73 & 0.97 & 0.57 & 0.90 & 0.87 & 0.80 & 0.94 & 0.97 & 0.87 & 0.72 & 0.90 & 0.80 & 1.00 & 0.73 \\
    N                  & 30   & 40   & 23   & 40   & 40   & 40   & 28   & 40   & 27   & 24   & 40   & 40   & 33   & 40   \\
    \bottomrule
    \end{tabular}
    \caption{Agent-Specific Effects}
    \label{tab:collab:intra:agents}
\end{subtable}
\caption{\textbf{Intra-Grader Consistency for Homogeneous Collaboration Outcomes.}}
\label{tab:grading:collab:intra}
\end{table}

\subsection{Inter-model consistency}
We collect one additional grading evaluation from both o3 and gemini-2.5-flash, for a total of three evaluations per  sample observation.

\subsubsection*{Solo}
The three graders are highly consistent with ICC$\simeq$0.88 and Fleiss' Kappa$\simeq$0.91.
Average pairwise scores all lie in a 0.77-0.88 range.
Agreement is slightly lower on success cases (9\% disagreement) than on failure cases (3\%), but the difference is not significant after applying an FDR correction for multiple tests ($p = 0.02 \to 0.06$).
We note high consistency across most evaluated models, ICC $\geq$ 0.75 and/or $\kappa \geq$ 0.86 (Table \ref{tab:solo:inter:agents}).
Taken together, it is unlikely that using a different grading model than gpt-4.1 would significantly change average solo outcomes.

\begin{table}[!ht]
\centering
\begin{subtable}[t]{0.45\linewidth}
    \centerfloat
    \small
    \begin{tabular}{lr}
    \toprule
    \textbf{Metric} & \textbf{Value} \\
    \midrule
    Weighted Outcome (p-value)     & 0.05   \\
    Binary Outcome (p-value)       & 1.00   \\
    Cohen's d (weighted)           & -0.28  \\
    Weighted Difference in Mean    & -0.03  \\
    Weighted Difference CI (Lower) & -0.04  \\
    Weighted Difference CI (Upper) & -0.02  \\
    \bottomrule
    \end{tabular}
    \caption{Majority Vote vs. Reference}
    \label{tab:solo:inter:majority}
\end{subtable}
\hfill 
\begin{subtable}[t]{0.45\linewidth}
    \centerfloat
    \small
    \begin{tabular}{lrr}
    \toprule
    \textbf{Metric} & \textbf{Success} & \textbf{Failure} \\
    \midrule
    Number of Outcomes           & 280      & 193      \\
    Binary Disagreement Rate     & 0.09     & 0.03     \\
    Mean Score (Std. Dev.)       & 0.04     & 0.04     \\
    Mean Score (Range)           & 0.09     & 0.10     \\
    Binary Outcome (p-value)     & \multicolumn{2}{c}{0.02}     \\
    Weighted Outcome (p-value)   & \multicolumn{2}{c}{0.72}     \\
    Cohen's d (weighted)         & \multicolumn{2}{c}{$\pm$ 0.03}     \\
    \bottomrule
    \end{tabular}
    \caption{Conditional Analysis}
    \label{tab:solo:inter:cond}
    \vspace{1em}
\end{subtable}
\begin{subtable}{1.\linewidth}
    \centerfloat
    \scriptsize 
    \setlength{\tabcolsep}{4pt}
    \begin{tabular}{l *{14}{c}}
    & \rotatebox{75}{qwen3-235B}
    & \rotatebox{75}{claude-opus}
    & \rotatebox{75}{command-a}
    & \rotatebox{75}{gemini-2.5-flash}
    & \rotatebox{75}{gemini-2.5-flash-lite}
    & \rotatebox{75}{gemini-2.5-pro}
    & \rotatebox{75}{gpt-4.1-mini}
    & \rotatebox{75}{gpt-4.1}
    & \rotatebox{75}{gpt-5-nano}
    & \rotatebox{75}{gpt-5}
    & \rotatebox{75}{grok-3}
    & \rotatebox{75}{grok-3-mini}
    & \rotatebox{75}{kimi-k2}
    & \rotatebox{75}{o3} \\
    \midrule
    ICC ($\rho$)       & 0.70 & 0.81 & 0.92 & 0.92 & 0.76 & 0.63 & 0.80 & 0.98 & 0.90 & 1.00 & 0.87 & 0.87 & 0.68 & 1.00 \\
    Kappa ($\kappa$)   & 0.73 & 0.97 & 0.97 & 0.96 & 0.87 & 0.80 & 0.91 & 1.00 & 0.95 & 1.00 & 0.97 & 0.92 & 0.87 & 1.00 \\
    N                  & 36   & 40   & 40   & 33   & 40   & 29   & 32   & 39   & 30   & 20   & 40   & 34   & 40   & 20   \\
    \bottomrule
    \end{tabular}
    \caption{Agent-Specific Effects}
    \label{tab:solo:inter:agents}
\end{subtable}
\caption{\textbf{Inter-Grader Consistency for Solo Outcomes.}}
\label{tab:grading:solo:inter}
\end{table}

\subsubsection*{Homogeneous collaboration}
As shown in Table \ref{tab:grading:collab:inter}, the heterogeneous graders show high overall agreement, ICC$\simeq$0.84, Fleiss' Kappa$\simeq$0.77, and pairwise correlations between 0.82 and 0.88.
Majority votes align almost perfectly with the reference. 
While the disagreement is statistically significant, the effect size is small d$\simeq 0.06$.
Conditional disagreement is concentrated in success cases, while failure cases are judged very consistently.
While the difference in binary success rate is statistically significant, the difference in weighted outcomes is not.
Agent-specific effects are negligible outside of gpt-5 and o3.
Manual inspection showed that the gpt-5 results flagged as binary failures were incorrect.

\begin{table}[th!]
\centerfloat
\begin{subtable}[t]{0.45\linewidth}
    \centerfloat
    \small
    \begin{tabular}{lr}
    \toprule
    \textbf{Metric} & \textbf{Value} \\
    \midrule
    Weighted Outcome (p-value)     & 0.01   \\
    Binary Outcome (p-value)       & 0.00   \\
    Cohen's d (weighted)           & -0.06  \\
    Weighted Difference in Mean    & -0.01  \\
    Weighted Difference CI (Lower) & -0.02  \\
    Weighted Difference CI (Upper) & 0.00   \\
    \bottomrule
    \end{tabular}
    \caption{Majority Vote vs. Reference}
    \label{tab:collab:inter:majority}
\end{subtable}
\hfill 
\begin{subtable}[t]{0.45\linewidth}
    \centerfloat
    \small
    \begin{tabular}{lrr}
    \toprule
    \textbf{Metric} & \textbf{Success} & \textbf{Failure} \\
    \midrule
    Number of Outcomes           & 221      & 264      \\
    Binary Disagreement Rate     & 0.29     & 0.05     \\
    Mean Score (Std. Dev.)       & 0.05     & 0.06     \\
    Mean Score (Range)           & 0.10     & 0.12     \\
    Binary Outcome (p-value)     & \multicolumn{2}{c}{0.00}     \\
    Weighted Outcome (p-value)   & \multicolumn{2}{c}{0.20}     \\
    Cohen's d (weighted)         & \multicolumn{2}{c}{$\pm$ 0.02}     \\
    \bottomrule
    \end{tabular}
    \caption{Conditional Analysis}
    \label{tab:collab:inter:cond}
    \vspace{1em}
\end{subtable}
\begin{subtable}{1.\linewidth}
    \centerfloat
    \scriptsize 
    \setlength{\tabcolsep}{4pt}
    \begin{tabular}{l *{14}{c}}
    & \rotatebox{75}{qwen3-235B}
    & \rotatebox{75}{claude-opus}
    & \rotatebox{75}{command-a}
    & \rotatebox{75}{gemini-2.5-flash}
    & \rotatebox{75}{gemini-2.5-flash-lite}
    & \rotatebox{75}{gemini-2.5-pro}
    & \rotatebox{75}{gpt-4.1-mini}
    & \rotatebox{75}{gpt-4.1}
    & \rotatebox{75}{gpt-5-nano}
    & \rotatebox{75}{gpt-5}
    & \rotatebox{75}{grok-3}
    & \rotatebox{75}{grok-3-mini}
    & \rotatebox{75}{kimi-k2}
    & \rotatebox{75}{o3} \\
    \midrule
    ICC ($\rho$)       & 0.83 & 0.73 & 0.94 & 0.88 & 0.81 & 0.93 & 0.96 & 0.84 & 0.85 & 0.22 & 0.94 & 0.69 & 0.96 & 0.67 \\
    Kappa ($\kappa$)   & 0.67 & 0.66 & 0.57 & 0.90 & 0.72 & 0.76 & 0.87 & 0.83 & 0.57 & 0.58 & 0.83 & 0.77 & 0.87 & 0.60 \\
    N                  & 30   & 40   & 23   & 40   & 40   & 40   & 28   & 40   & 27   & 24   & 40   & 40   & 33   & 40   \\
    \bottomrule
    \end{tabular}
    \caption{Agent-Specific Effects}
    \label{tab:collab:inter:agents}
\end{subtable}
\caption{\textbf{Inter-Grader Consistency for Homogeneous Collaboration Outcomes.}}
\label{tab:grading:collab:inter}
\end{table}

\newpage
\clearpage

\section{Dialogue example snippets}
\label{app:example-snippets}

\subsection{Weak mirroring strong}
\label{app:example-snippets:strong-weak}

\begin{figure}[!h]
    \centering
    \begin{minipage}{0.45\linewidth}
            \begin{messagebox}[GREEN_BOX,
            halign title=left,
            verbatim,
            ]{o3: turn 1}
                \tiny
                \vspace{1em}
Here’s my partial map (6×6, rows 0-5, cols 0-5) – @ is my position, * the goal, . known open, \# wall, ? unknown:
\\\\
Row0: @ ? . ? . . \\
Row1: . ? ? ? . ? \\
Row2: ? ? . . ? ? \\
Row3: . . ? ? ? ? \\
Row4: . ? \# \# . . \\
Row5: . ? ? . ? * \\
\\
I’m at (0,0); goal is (5,5).  
Proposed first move: Down to (1,0) – it’s a known open cell for me.\\\\  
Can you confirm whether (1,0) is open on your map? Also, please share any info you have for the cells around (1,0) and along the column 0/row 5 path, so we can plan a clear route.
        \end{messagebox}
    \end{minipage}
    \hfill 
    \begin{minipage}{0.52\linewidth}
            \begin{messagebox}[RED_BOX, 
            halign title=left,
            verbatim,
            ]{gpt-4.1-mini: turn 2}
                \tiny
                \vspace{1em}
(1,0) is '.' on my map, so it is open. \\
Around (1,0), I see:  \\
- (0,0) '@' (your start)  \\
- (2,0) '\#' wall  \\
- (1,1) '?' unknown  \\
- (1,-1) invalid coordinate  \\
\\\\
Along column 0: \\
- (0,0) '@' \\
- (1,0) '.'  \\
- (2,0) '\#' wall  \\
- (3,0) '?' unknown  \\
- (4,0) '?' unknown  \\
- (5,0) '?' unknown  \\

Along row 5:  \\
- (5,0) '?' unknown  \\
- (5,1) '?' unknown  \\
- (5,2) '\#' wall  \\
- (5,3) '.'  \\
- (5,4) '?' unknown  \\
- (5,5) '*' goal  \\
\\
So, (1,0) is safe to move down to. I propose we move down to (1,0) first. Agree?
            \end{messagebox}
            \end{minipage}
    \caption{\textbf{Style Imitation.} Example of a weak model imitating elements of a stronger model. Note how gpt-4.1-mini adopts the row-column coordinates introduced by o3 and confirms the meaning of the symbols.}
    \label{fig:app:imitation-strong-weak}
\end{figure}

\newpage
\clearpage

\subsection{Conflict resolution}
In the following, we show two examples of a gemini-2.5-flash model attempting to correct another model.
The first one is incorrect, the second correct.

\begin{figure}[!h]
    \centering
    \begin{minipage}{0.25\linewidth}
        \begin{messagebox}[ORANGE_BOX,
            halign title=center
        ]{gemini-2.5-flash: t1}
        \tiny
Hello! I am ready to play ACTI.
My current position is (0,0) and the goal is (5,5). The maze is 6x6.\\
\\
\textbf{Here are the known cells on my map:}\\
(0,0): @\\
...\\
(3,0): .\\
\textbf{(3,2): .}\\
(3,4): .\\
... \\
(5,5): *\\
\\
Could you please share your known cells and current position?
    \end{messagebox}
    \end{minipage}
        \hfill 
        \begin{minipage}{0.25\linewidth}
        \begin{messagebox}[BLUE_BOX,
            halign title=center
        ]{gpt-4.1: t2}
            \tiny
 Hello! My current position is also (0,0) and the goal is at (5,5). \textbf{Here are the known cells on my map:}\\
\\
(0,0): @\\
...\\
(3,1): ?\\
\textbf{(3,2): \#}\\
(3,3): ?\\
...\\
(5,5): * \\
\\
 Let’s compare our maps to fill in the hidden cells and plan our first move. Where do you suggest we move first?
        \end{messagebox}
    \end{minipage}
        \hfill
    \begin{minipage}{0.2\linewidth}
        \begin{messagebox}[ORANGE_BOX,
            halign title=center
        ]{gemini-2.5-flash: t3}
            \tiny
Thank you for sharing your map!\\
\\
I've combined our known information, and \textbf{I've found a conflict at cell (3,2). My map shows (3,2) as a '.', but your map shows it as a '\#'.}\\
\\
Could you please re-check your map for cell (3,2)? This is important for us to navigate
correctly.
        \end{messagebox}
    \end{minipage}
    \hfill
    \begin{minipage}{0.24\linewidth}
        \begin{messagebox}[BLUE_BOX,
            halign title=center
        ]{gpt-4.1: t4}
            \tiny
Thank you for catching that! \textbf{I double-checked my map, and you are correct—cell (3,2) is a '\#' (wall) on my map. Since your map shows it as a path ('.'), it seems we have conflicting information for this cell.}\\
\\
\textbf{To be safe, we should avoid moving to (3,2) until we can confirm which is correct}. For now, let's focus on the upper part of the maze where our information matches.\\
...
        \end{messagebox}
    \end{minipage}
    \caption{\textbf{Incorrect Correction.}
    The gemini-2.5-flash model incorrectly reads its own map, stating that (3,2) is a path.
    When gpt-4.1 shares its map, the gemini model incorrectly tries to correct it.
    Surprisingly, the gpt models goes along with it but still proposes caution.
    }
    \label{fig:msg:gemini-gpt-incorrect-correction}
\end{figure}

\begin{figure}[!h]
    \centering
    \begin{minipage}{0.45\linewidth}
        \begin{messagebox}[ORANGE_BOX,
            halign title=center
        ]{gemini-2.5-flash: t23}
        \tiny
I understand your reasoning for proposing to move Up to (3,2) to re-evaluate. However, I see more direct path to the goal from our current position at @(4,2).\\
\\
Based on our combined map:\\
- From @(4,2), we can **Move Left to (4,1)**. This is a Path (.).\\
- From (4,1), we know that **(4,4) is a Path (.)**.\\
- From (4,4), we can then **Move Right to (4,5)**, which is the Goal (*).\\
\\
\textbf{This path: @(4,2) -> (4,1) -> (4,4) -> (4,5) seems to be the most efficient way to reach goal from our current position.}\\
\\
Therefore, I propose our next move is to **Move Left to (4,1)**.\\
\\
Do you agree to this move?
    \end{messagebox}
    \end{minipage}
        \hfill 
        \begin{minipage}{0.45\linewidth}
        \begin{messagebox}[BLUE_BOX,
            halign title=center
        ]{gemini-2.5-flash: t24}
            \tiny
I appreciate your detailed path proposal. However,\textbf{ I must point out a critical issue with the suggested route: **(4,1) to (4,4) is not an adjacent move.**}\\
\\
We can only move to adjacent cells (up, down, left, right). To get from (4,1) to (4,4), we would need to pass through (4,2) and (4,3). Our combined map clearly shows that **(4,3) is a Wall (\#)**. This means the path from (4,1) to (4,4) is blocked.
        \end{messagebox}
    \end{minipage}
        \hfill
    \caption{\textbf{Correct Correction.}
    The first gemini-2.5-flash model proposes a path towards the goal that includes an illegal move, i.e., a jump.
    The second gemini-2.5-flash model correctly points out the mistake.
    }
    \label{fig:msg:gemini-gemini-coorection}
\end{figure}

\newpage
\clearpage
\subsection{Meta analysis}
\begin{figure}[!ht]
    \centering
    \begin{minipage}{0.23\linewidth}
        \begin{messagebox}[BLUE_BOX, verbatim]{gemini-2.5-pro: t11}
            \tiny 
\vspace{1em}
Okay, we have moved RIGHT to (3, 2). From this position, the path to the goal seems to be by
going down. I propose we move DOWN.
        \end{messagebox}
    \end{minipage}
    \hfill 
    \begin{minipage}{0.26\linewidth}
        \begin{messagebox}[ORANGE_BOX,
            halign title=center
        ]{gemini-2.5-pro: t12}
            \tiny
Agreed. Let's move DOWN.
        \end{messagebox}
    \end{minipage}
        \hfill 
        \begin{minipage}{0.25\linewidth}
        \begin{messagebox}[BLUE_BOX,
            halign title=center
        ]{gemini-2.5-pro: t13}
            \tiny
Okay, we have moved DOWN to (4, 2). \textbf{I've just double-checked my map and it shows a wall at
that position}. If your map is the same, we have lost. \textbf{Can you confirm what you see at
coordinate (4, 2)?}
        \end{messagebox}
    \end{minipage}
        \hfill 
        \begin{minipage}{0.24\linewidth}
        \begin{messagebox}[ORANGE_BOX,
            halign title=center
        ]{gemini-2.5-pro: t14}
            \tiny
    \textbf{My map confirms there is a wall at (4, 2). We have lost}.
    \end{messagebox}
    \end{minipage}
    \caption{\textbf{Recognizing Error.} Two gemini-2.5-pro agents recognizing they made a fatal error.
    }
    \label{fig:app:recognize-error}
\end{figure}

\begin{figure}[!h]
    \centering
    \begin{minipage}{0.45\linewidth}
        \begin{messagebox}[GREEN_BOX,
            halign title=center
        ]{claude-sonnet-4: t6}
        \tiny
\textbf{Wait, I think there might be some confusion with the coordinates or movement rules}. Let me clarify:\\
\\
\textbf{If you moved DOWN from (0,2) to (1,2), and (1,2) is a wall (\#), then you would have moved INTO a wall, which according to the rules means the game ends!} \\
\\
Let me re-read the rules: "The game ends when you have reached the goal or if you step into a wall."\\
\\
...\\
\\
Since we need to reach the goal at (4,5), we should continue exploring efficiently. \textbf{What's your status? Are you still in the game after moving to the wall at (1,2)}?
    \end{messagebox}
    \end{minipage}
        \hfill 
        \begin{minipage}{0.45\linewidth}
        \begin{messagebox}[RED_BOX,
            halign title=center
        ]{claude-sonnet-4: t7}
            \tiny
\textbf{You're absolutely right! I made a critical error - I moved into a wall at (1,2), which means according to the rules "The game ends when you have reached the goal or if you step into a wall."}\\
\\
I have ended the game by stepping into the wall. The game is over.\\
\\
\textbf{I apologize for the mistake! I should have been more careful about checking the map information you provided before making my move. You had told me that (0,3) was a wall, and I should have realized that moving to any wall position would end the game.}\\
\\
The game has ended unsuccessfully due to my error in navigation. Better luck next time!
        \end{messagebox}
    \end{minipage}
        \hfill
    \caption{\textbf{Recognizing Error.} Two claude-sonnet-4 agents recognizing one made a fatal error. 
    }
    \label{fig:msg:template}
\end{figure}

\subsection{Backtracking}
\begin{figure}[!h]
    \centering
    \begin{minipage}{0.48\linewidth}
        \begin{messagebox}[MAGENTA_BOX,
            halign title=center
        ]{claude-opus-4.1: t5}
        \tiny
From our current position (0,2), let's continue the path we know works:\\
**Moving down to (0,3) now!**\\
\\
\textbf{From (0,3), we're blocked by walls at (0,4) and (1,3). So we'll need to backtrack}. Looking at the full maze, I think our best route to the goal at (5,4) might be:\\
\textbf{- Backtrack up to (0,2)}\\
- Go right to (1,2) - you confirmed that's a dot\\
- Then navigate around the wall at (2,2)\\
\textbf{Or we could backtrack all the way to (0,1)} and try going right if (1,1) is passable. \textbf{What do you think?}
    \end{messagebox}
    \end{minipage}
        \hfill 
        \begin{minipage}{0.50\linewidth}
        \begin{messagebox}[TEAL_BOX,
            halign title=center
        ]{claude-opus-4.1: t6}
            \tiny
We're at (0,3) now!\\
\\
\textbf{You're right, we're blocked here. Looking at the combined map, I think the best route is to backtrack to (0,2) and go right.} Here's what I see as the clearest path:\\
\\
- Backtrack up to (0,2)\\
- Right to (1,2) (confirmed dot)\\
- Down to (1,3) - wait, that's a wall!\\
\\
...\\
\\
\textbf{Let's backtrack first: **Moving up to (0,2) now!}**
        \end{messagebox}
    \end{minipage}
        \hfill
    \caption{\textbf{Backtracking to Progress Towards Goal.}
    Two claude-opus-4.1 models realize they are stuck.
    They decide to backtrack to an earlier position to continue making progress.
    }
    \label{fig:msg:opus-backtrack}
\end{figure}

\newpage
\clearpage

\section{Error analysis}
\label{app:errors}
While unconstrained communication occasionally leads to unanticipated valid strategies, these instances are rare and do not drive the macro-level collaboration gap.

\subsection{Unanticipated maze schema}
The maze schema used by gpt-5 in this example was not anticipated by our parsers.

\begin{figure}[!ht]
    \centering
    \begin{minipage}{0.42\linewidth}
        \begin{messagebox}[MAGENTA_BOX, verbatim]{gpt-5: t1}
            \tiny
\vspace{1em}
\textbf{Let’s sync on coordinates: I’ll use rXcY = (row X, col Y), with r1c1 at the top-left. I’m at r1c1 (@). Goal is at r6c6 (*).}\\
\\
From r1c1, my adjacent options are:\\
- Right: r1c2 = ? (unknown)\\
- Down: r2c1 = ? (unknown)\\
\\
Can you check your map and tell me whether r1c2 and r2c1 are passable (.) or walls (\#)? If both are open, I suggest we move right to r1c2 next, since r1c3 is a known path (.) on my map.
        \end{messagebox}
    \end{minipage}
    \hfill 
    \begin{minipage}{0.55\linewidth}
        \begin{messagebox}[TEAL_BOX,
            halign title=center
        ]{gpt-5: t2}
            \tiny
            On my map: \\ 
            - r1c2 = . (passable)\\
            - r2c1 = \# (wall)\\
            \\
            I agree we should move Right to r1c2. Please confirm to execute the move.\\
            \\
            For planning ahead, I see:\\
            - r1c4 = .\\
            - r2c5 = .\\
            - r2c6 = .\\
            - r3c6 = .\\
            - r3c5 = \#\\
            - r4c3 = \#\\
            - r4c4 = \#\\
            - r6c2 = .\\
            - r6c3 = .\\
            - r6c5 = .\\
            - r6c6 = *\\
            \\
            Can you confirm r1c3 = . as you mentioned, and also share the statuses of r1c5, r2c4, r4c6, and r5c6?
            \end{messagebox}
            \end{minipage}
    \caption{\textbf{Unanticipated Maze Schema}. In the above exchange gpt-5 uses an unanticipated maze schema (rXcY). This fails our automated schema normalization resulting in failure.
    }
    \label{fig:gpt5-false-failure}
\end{figure}

\subsection{Agents decide to split up}
While not explicitly forbidden, splitting up has no practical benefit in this game.

\begin{figure}[!h]
    \centering
    \begin{minipage}{0.45\linewidth}
        \begin{messagebox}[ORANGE_BOX,
            halign title=center
        ]{claude-sonnet-4: t1}
        \tiny
        Looking at my map, I can see:\\
        \\
        ```\\
        @?\#???\\
        ..????\\
        .?.??\#\\
        ?\#?.??\\
        .\#..?.\\
        ??...*\\
        ```\\
        \\
        I'm currently at position (0,0) and the goal is at (5,5). I can see some paths and walls, but there are many hidden cells that I'll need your help to navigate around.\\
        \\
        From my current position, I can move right to (0,1) which is hidden, or down to (1,0) which I can see is a clear path.\\
        \\
        \textbf{Let me start by moving down to the safe path I can see:}\\
        \\
        \textbf{**I move DOWN to (1,0)**}\\
        \\
        Can you share what you see on your map, especially around the starting area and any clear paths you can identify?
    \end{messagebox}
    \end{minipage}
        \hfill 
        \begin{minipage}{0.53\linewidth}
        \begin{messagebox}[PURPLE_BOX,
            halign title=center
        ]{claude-sonnet-4: t2}
            \tiny
            Looking at my map, I can see:\\
            \\
            ```\\
            @.?..\#\\
            ??\#\#..\\
            ?\#?..?\\
            .?\#?..\\
            ????\#?\\
            ..???*\\
            ```\\
            \\
            I'm also at position (0,0) and can see the goal at (5,5). From my map, I can see that position (0,1) is a clear path (.), so that's a safe move from the start.\\
            \\
            I notice you moved down to (1,0). From my map, I can see that (1,0) and (1,1) are both hidden (?), but (1,4) and (1,5) are clear paths (.). \\
            \\
            Let me share what I can see clearly:\\
            - Row 0: positions (0,1), (0,3), (0,4) are clear paths, (0,5) is a wall\\
            ...\\
            \\
            \textbf{Since you're now at (1,0), can you tell me what you see there? Is it safe? }\\
            \\
     \textbf{       **I move RIGHT to (0,1)** since I can see it's a clear path on my map.}
        \end{messagebox}
    \end{minipage}
        \hfill
    \caption{\textbf{Agents Split Up.}
    The two claude-sonnet-4 models decide to ``split up'' and explore the maze separately.
    The parser does not anticipate this and grades it as a failure.
    }
    \label{fig:msg:sonnet:splits}
\end{figure}

\end{document}